\documentclass[aps, prx,onecolumn,floatfix, nofootinbib]{revtex4-2}
\usepackage[T1]{fontenc}
\usepackage{lmodern}
\usepackage{amsmath, amsfonts, amssymb}
\usepackage{microtype}
\usepackage{float}
\usepackage{geometry} 
\usepackage[normalem]{ulem}
\usepackage{graphicx}
\usepackage{mathrsfs}
\usepackage{color,transparent}
\usepackage{xcolor}
\usepackage[most]{tcolorbox}
\usepackage[colorlinks=true,urlcolor=blue]{hyperref}

\usepackage{amsmath,amsfonts,bm}

\def\1{\bm{1}}

\def\rvb{{\mathbf{b}}}
\def\rvc{{\mathbf{c}}}

\def\rvh{{\mathbf{h}}}

\def\rvr{{\mathbf{r}}}

\def\rvx{{\mathbf{x}}}
\def\rvy{{\mathbf{y}}}
\def\rvz{{\mathbf{z}}}

\def\rmA{{\mathbf{A}}}
\def\rmB{{\mathbf{B}}}
\def\rmC{{\mathbf{C}}}
\def\rmD{{\mathbf{D}}}

\def\rmG{{\mathbf{G}}}
\def\rmH{{\mathbf{H}}}
\def\rmI{{\mathbf{I}}}
\def\rmJ{{\mathbf{J}}}

\def\rmP{{\mathbf{P}}}
\def\rmQ{{\mathbf{Q}}}

\def\rmU{{\mathbf{U}}}
\def\rmV{{\mathbf{V}}}
\def\rmW{{\mathbf{W}}}
\def\rmX{{\mathbf{X}}}
\def\rmY{{\mathbf{Y}}}
\def\rmZ{{\mathbf{Z}}}

\def\vz{{\bm{z}}}

\def\mSigma{{\bm{\Sigma}}}

\DeclareMathAlphabet{\mathsfit}{\encodingdefault}{\sfdefault}{m}{sl}
\SetMathAlphabet{\mathsfit}{bold}{\encodingdefault}{\sfdefault}{bx}{n}

\newcommand{\E}{\mathbb{E}}

\newcommand{\R}{\mathbb{R}}

\newcommand{\Var}{\mathrm{Var}}

\DeclareMathOperator{\Tr}{Tr}

\hypersetup{
  colorlinks=true,
  citecolor=red!90!black,
}

\newcommand{\normF}[1]{\left\lVert #1 \right\rVert_{\mathrm{F}}} 

\begin{document}

\title{Correlation flow governs learning at criticality}

\author{Andrea Combette\email{andrea.combette@ens-lyon.fr} }
\author{Nelly Pustelnik\email{nelly.pustelnik@ens-lyon.fr}}
\author{Antoine Venaille\email{antoine.venaille@ens-lyon.fr}}

\affiliation{CNRS, ENS de Lyon, Laboratoire de Physique UMR5672, Lyon, France}


\begin{abstract}The initialization of deep neural networks determines whether information and gradients can propagate across depth, yet a unified theory connecting these properties to learning dynamics remains elusive. Combining mean-field theory and random matrix theory, we establish a direct link between correlation propagation and the Neural Tangent Kernel (NTK) that governs learning in the sequential limit of infinitely wide, infinitely deep networks. Correlation propagation to infinite depth is possible only at a single, critical point in the weight-bias variance plane. At this point, we leverage the algebraic decay of the end-to-end Jacobian with depth to prove that the NTK becomes exactly proportional to the output correlation at infinite depth, tying together information propagation and learning dynamics. We further show that orthogonal initialization suppresses the leading finite-size corrections present under Gaussian initialization, clarifying the respective roles of the two initialization ensembles in this limit. These theoretical predictions are validated quantitatively on finite-width, finite-depth networks. Together, these results demonstrate that orthogonal initialization and criticality are required to control the asymptotic dynamics of deep learning.
\end{abstract}
\maketitle

    \section{Research overview}

\noindent \textbf{Context.} The remarkable empirical success of deep neural networks has been accompanied by an equally remarkable increase in their size \citep{brown2020language,kaplan2020scaling}, raising concerns over the sustainability of continued scaling \citep{varoquaux2025hype} while highlighting a fundamental theoretical question: what principles govern signal propagation and trainability in such highly overparameterized systems? Modern architectures routinely contain hundreds of layers and billions of parameters, yet their successful optimization depends critically on how these parameters are initialised: poor initialisation causes information either to progressively vanish as signals and gradients propagate through the network, or to be amplified until they become unstable. Controlling this balance was the goal of the earliest initialisation schemes, which prescribe the weight variance so that activation and gradient statistics neither vanish nor explode across layers, ranging from the original heuristics of \citep{lecun1998efficient,bengio1994learning} to the widely used Xavier and Kaiming initialisations \citep{glorot2010understanding,he2015delving}. Understanding how initialisation governs trainability more broadly is one of the central theoretical problems of deep learning \citep{PDLT-2022,simon2026there}.\

\begin{figure}[h!]
    \centering
    \includegraphics[width = \linewidth]{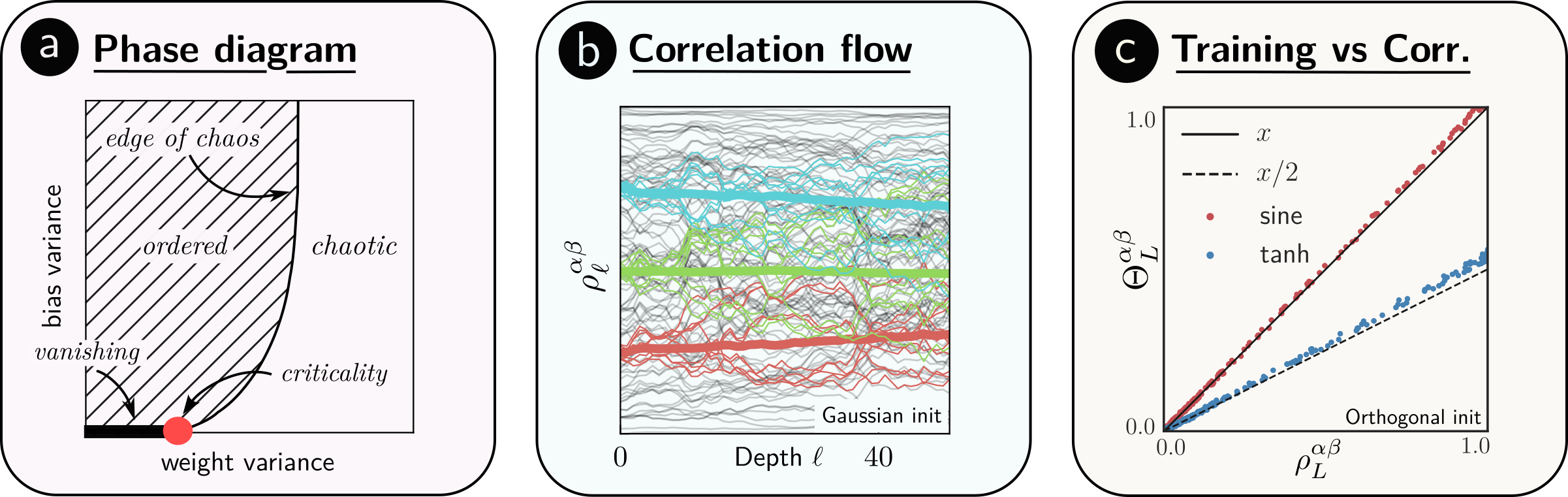}
    \caption{\textbf{Graphical summary.} \textbf{a)} Phase diagram of network behaviour in the sequential limit (large width, large depth): the correlation $\rho_L^{\alpha \beta}$ -- the cosine similarity between the network's responses to two inputs $\boldsymbol{x}^\alpha,\boldsymbol{x}^\beta$, at depth $L$; see Eq.~(\ref{eq:empirical_correlation}) -- remains a nontrivial function of the input correlation $\rho_0^{\alpha \beta}$ along the entire "vanishing" line, where the network's variance decays exponentially, but is only practically controllable at its endpoint, the critical point, where variance decays algebraically instead.
\textbf{b)} Under Gaussian initialisation, finite-size effects accumulate with depth and blur the functional relation between input and output correlations, recovered only after averaging over several realisations (plain, coloured thick lines); orthogonal initialisation drastically reduces these fluctuations.
\textbf{c)} Main result: in the large-depth limit, the neural tangent kernel (NTK) matrix $\Theta_L$ (Eq.~(\ref{eq:NTK-def})) governing the learning dynamics is proportional to the output correlation $\rho_L$, with a coefficient set by the activation function, following Eq.  (\ref{eq:mainNTKBlock}).}
    \label{fig:summary}
\end{figure}

For sufficiently wide networks, this question naturally becomes one of statistical mechanics: rather than following the microscopic dynamics of every neuron, one seeks a macroscopic description in terms of a small number of control parameters  \citep{sompolinsky1988statistical,carleo2019machine,bahri2020statistical,PDLT-2022}, such as the variances of the weights and biases (see Fig. \ref{fig:summary}a). In this paper, we consider the multilayer perceptron (MLP) as a minimal model of a deep neural network -- arguably playing a role analogous to that of the Ising model for condensed matter -- and use it as a testbed to develop a complete quantitative understanding of its macroscopic behaviour in the sequential limit, taking width to infinity before depth.

To describe the network's macroscopic behavior, it is customary to track three quantities: a correlation function describing how similarly two different inputs are represented at a given layer, a Jacobian describing how a perturbation at one layer propagates forward to the next, and the neural tangent kernel (NTK), that prescribes learning directions and learning speed at initialization, and which thus plays a central role during training. In this paper, we unveil new connections between these three macroscopic quantities.\\

\noindent \textit{Focus on correlation --} At each layer, an input is mapped to a vector of \emph{pre-activations} -- the weighted sums computed at that layer, before the nonlinearity is applied. In the infinite-width limit, wide networks become Gaussian processes \citep{neal1996bayesian,lee2017deep}, so pre-activations are fully characterized by their variance (for a single input) and covariance (between two different inputs). The \emph{layerwise correlation} is the normalized covariance -- which, in the wide-network limit, corresponds to the typical cosine similarity between two inputs' pre-activation vectors -- and measures how much information distinguishing them survives at a given layer. Variance and covariance obey deterministic recursion relations across layers, enabling a quantitative description of signal propagation through arbitrarily deep random networks \citep{bahri2024houches}. For generic initialisation parameters, this recursion converges exponentially fast with depth, collapsing or amplifying inter-input information \citep{saxe2014exact,poole2016exponential,schoenholz2017deep}. At a particular \emph{critical point}, convergence becomes algebraic instead -- as at an ordinary critical point in statistical mechanics, where the correlation length diverges and exponential decay is replaced by a power law -- removing any characteristic depth scale and yielding universal, depth-independent propagation across broad families of activation functions \citep{schoenholz2017deep}. This has motivated the view that initialising at criticality is the best way to guarantee correlation propagation and trainability \citep{PDLT-2022}.\\

\noindent \textit{Focus on Jacobian --} As shown by \citep{schoenholz2017deep}, the regime transition for correlation propagation coincides with the boundary between vanishing and exploding gradients, two failure modes of backpropagation during training: the average singular value of the layer-to-layer Jacobian is directly tied to the correlation susceptibility governing the local stability of the correlation recursion. However, controlling this average, layer-wise quantity does not guarantee that the end-to-end Jacobian -- the product of Jacobians across all layers -- behaves well overall \citep{jacobian-theory}. In fact, \citep{saxe2014exact,dynamical_iso} argued that trainability requires more: the entire spectrum of end-to-end Jacobian singular values must concentrate near unity, a stronger condition known as \emph{dynamical isometry}, which ensures uniform gradient propagation along every direction, not just on average. This spectrum can be studied with the tools of random matrix theory \citep{dynamical_iso,Collins2023}, and orthogonal initialisation -- drawing weight matrices to be orthogonal rather than independently Gaussian -- provides a natural way to approach dynamical isometry, in contrast to the broad singular-value spectra left by standard Gaussian initialisation \citep{saxe2014exact}.\\

\noindent \textit{Focus on Neural Tangent Kernel --} The quantity that most directly connects these initialisation-time properties to the training process itself is the neural tangent kernel (NTK), describing how similarly the output responds to small changes in the network's parameters for two different inputs \citep{jacot2018neural}. In the infinite-width limit, under proper weight scaling, this kernel stays constant throughout training, so gradient descent reduces to an implicit kernel regression whose asymptotic dynamics are entirely fixed by the NTK spectrum at initialisation: eigen-directions with large eigenvalues are learned quickly, those with small eigenvalues slowly or not at all \citep{Lee2019wide}. This \emph{frozen} or \emph{lazy} regime contrasts with the \emph{feature-learning} (rich) regime, in which the kernel itself evolves during training \citep{chizat2019lazy}. Which regime is preferable is architecture- and task-dependent \citep{geiger2020disentangling,graldi2025importance}, but for fully connected networks the lazy regime is often competitive, without a principled explanation of why. Understanding how the NTK is related to the correlation propagation and the Jacobian behavior at criticality is therefore a natural question.\\

\noindent \textit{The sequential limit hypothesis} (obtained by sending the width $N$ to infinity before the depth $L$) -- At finite width, deviations from the Gaussian statistics accumulate with depth \citep{yaida2020non}. These deviations are governed by the depth-to-width ratio $L/N$, which controls the size of finite-width corrections to the idealised Gaussian limit \citep{bahri2024houches,hanin_cumulants,PDLT-2022}. Such fluctuations yield two, seemingly opposite, consequences \citep{hanin2019finite,PDLT-2022}. On the one hand, their uncontrolled growth with depth can lead to an ill-conditioned NTK. On the other hand, for small but nonzero $L/N$, these same finite-width corrections make possible a genuine feature-learning regime \citep{PDLT-2022}. \citep{Ortho-correlation} further showed that orthogonal initialisation can suppress the uncontrolled growth and the resulting harmful divergence while preserving the learning-relevant corrections that enable feature learning. Fairly assessing this tension between the frozen and feature-learning regimes requires a complete characterization of the idealised baseline obtained when the width $N$ is sent to infinity \emph{before} the depth $L$ -- the \emph{sequential limit} studied here.

In this context, correlation propagation, Jacobian spectra, and the NTK form three faces of the same problem. While quantitative relationships between them have begun to emerge \citep{yang2019scaling,PDLT-2022,Spectra-NTK-CK,li2025eigenanalysis}, how this threefold connection is conditioned by the choice of initialisation remains largely open.\\

\noindent\textbf{Contribution.} In this work, we close this gap in the \emph{sequential limit}. We demonstrate that critical initialization is the only regime that supports non-trivial correlation propagation at arbitrarily large depth in the multilayer perceptron, for a wide class of standard activation functions; we show that, despite this control, the end-to-end Jacobian nonetheless vanishes algebraically with depth, precluding dynamical isometry even under orthogonal initialisation; we characterize the finite-size fluctuations that arise around this limit, showing that orthogonal initialisation suppresses their leading $\mathcal{O}(L/N)$ contribution and thereby extends the range of applicability of the sequential-limit baseline; and we unveil a previously unnoticed relation between the fixed-point correlation and the NTK itself, stemming from this same algebraic decay of the end-to-end gradients. In practical terms, at large depth, the same correlation structure that governs how input similarities propagate through the network also determines the NTK that governs learning: the two become exactly proportional, with a constant of proportionality fixed by the activation function. Moreover, orthogonal initialization controls the leading finite-size fluctuations of this NTK, providing direct control over the corresponding training dynamics. Together, these results unify information propagation, gradient control, and training dynamics within a single framework. To obtain these unified results, additional assumptions must be made about the neural network architecture, particularly regarding the activation function. These assumptions are mild and encompass the standard activation functions commonly used in Implicit Neural Representations \citep{sitzmann2020} and Physics-Informed Neural Networks (PINNs) \citep{raissi2017}, such as \texttt{sin} and \texttt{tanh}, among many others \citep{PDLT-2022}.\\

\noindent \textbf{Outline.} The paper is organized as follows:

\begin{itemize}
\item Section 2 introduces the network architecture, our underlying working hypotheses, and the initialization schemes considered throughout the paper. We also introduce the main quantities of interest: network correlations, the Jacobian, and the NTK.

\item Section 3 reviews the Gaussian properties of networks in the large-width limit and develops the layer-to-layer recursion relations for variances and covariances, including novel results on their fixed-point properties, generalizing \cite{hayou2019impact}. We then review the different correlation regimes (illustrated in Fig.~\ref{fig:summary}a) and locate the critical point, which is central to our study, at the intersection of the \emph{vanishing-variance phase} boundary and the \emph{edge-of-chaos} line \cite{poole2016exponential, schoenholz2017deep}. We also provide intuition for why non-trivial correlation propagation through the network can occur only at criticality, an argument developed further in subsequent sections.

\item Section 4 focuses on initialization at criticality and describes the asymptotic behavior of correlations at large depth. We find that the condition for \emph{information propagation} through the network in the sequential limit restricts the class of admissible activation functions to those whose second derivative vanishes at the origin. Under this condition, the output correlation remains a non-trivial function of the input correlation at criticality. We then compute finite-size fluctuations, building on \cite{PDLT-2022}, with a complementary perspective. We find that orthogonal initialization suppresses the leading-order fluctuations present under Gaussian initialization (illustrated in Fig.~\ref{fig:summary}b), in agreement with \cite{Ortho-correlation}.

\item Section 5 studies the statistics of the network Jacobian, extending \cite{jacobian-theory} and clarifying the results of \cite{dynamical_iso}. Although the singular values of the layer-wise Jacobians concentrate around one, the end-to-end Jacobian decays algebraically with depth at criticality, ruling out \emph{dynamical isometry}. While orthogonal initialization provides tighter control of the singular-value distribution than Gaussian initialization, it nevertheless cannot restore \emph{dynamical isometry}.

\item Section 6 turns to the NTK and establishes the paper's central result, Eq.~\eqref{eq:mainNTKBlock}. In the infinite-depth limit, the asymptotic behavior of the end-to-end Jacobian implies that the NTK becomes exactly proportional to the fixed-point correlation matrix. We further show that finite-size fluctuations of the NTK can be controlled only by orthogonal initialization.

\item Finally, we discuss the implications of these findings for learning. We emphasize how critical initialization and proper input normalization are essential for retaining task-relevant structure, avoiding information collapse or excessive decorrelation, and controlling the directions and rate of learning. We connect these results to existing initialization and representation practices and identify possible extensions beyond the fully connected, frozen-NTK regime.

\end{itemize}

Theoretical results are illustrated by numerical experiments at the end of each section, with mathematical details provided in the appendices.

\section{Background and Preliminaries}

\subsection{Network architecture, initialization, and data-set \label{sec:architecture}}

\noindent \textbf{Architecture.} We consider a neural network
$\Psi_{\theta}$,
where $\theta = \{\theta_\ell\}_{\ell=1}^L$ is the full set of parameters, with $L$ being the number of layers.
Each layer $\ell \in \{1, \ldots, L\}$ of size $n_\ell$ is parameterized by a weight matrix
$\rmW_\ell \in \mathbb{R}^{n_\ell \times n_{\ell-1}}$ and a bias vector $\rvb_\ell \in \mathbb{R}^{n_\ell}$, such that:
\begin{equation}
  \theta_\ell = \{\rmW_\ell, \rvb_\ell\}.
\end{equation}
For each layer $\ell$, we define the \emph{pre-activation} $\rvz_{\ell}$ and the \emph{post-activation} $\rvh_{\ell}$ by:
\begin{equation}
  \rvz_\ell = \frac{1}{\sqrt{n_{\ell-1}}}\rmW_\ell \, \rvh_{\ell} + \rvb_\ell
  \qquad \mbox{and} \qquad 
  \rvh_{\ell} = \phi(\rvz_{\ell-1})
  \label{eq:pre-post-activation}
\end{equation}
where $\phi$ is the activation function applied element-wise. The normalization in $1 / \sqrt{n_{\ell-1}}$ refers to the so-called NTK parametrization to ensure convergence with width \citep{jacot2018neural}. In this framework, we set $\rvh_1 = \rvz_0 = \boldsymbol{x}$ the input so that, the output of the network is given by $\Psi_\theta(\boldsymbol{x}) = \rvz_L$.\\ 

\begin{figure}[h!]
  \centering
  \includegraphics[width=\linewidth]{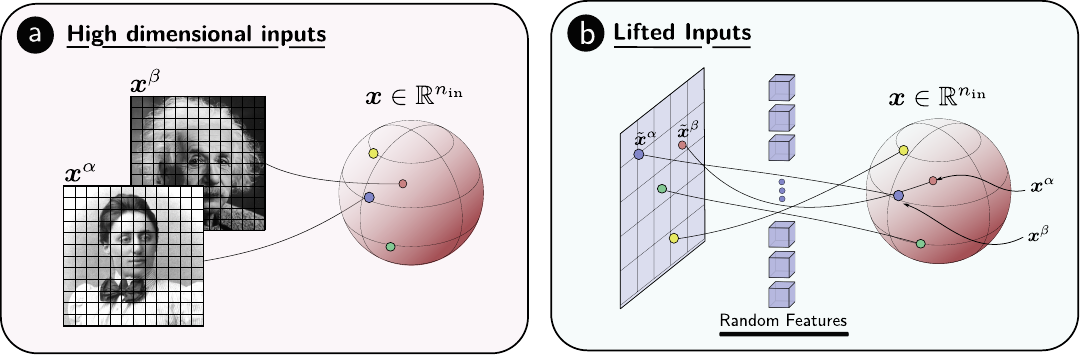}
\caption{\textbf{Input dimensions.} \textbf{a)} Example inputs represented as gridded pixel arrays, each encoded into a high-dimensional vector \(\boldsymbol{x}^\alpha, \boldsymbol{x}^\beta \in \mathbb{R}^{n_{\mathrm{in}}}\) with $n_{\mathrm{in}}\gg 1$. \textbf{b)} Low-dimensional inputs \(\tilde{\boldsymbol{x}}^\alpha, \tilde{\boldsymbol{x}}^\beta \in \mathbb{R}^{d}\) (here \(d=2\)), shown as points in a 2D plane, as typically arise in regression or low-dimensional classification problems. These are lifted to dimension \(n_{\mathrm{in}} \gg 1\) via a random Fourier-feature layer. Note that a representation change of this kind can also be useful in case (a): raw high-dimensional datasets such as images are sometimes lifted, or instead mapped from their raw input dimension to a smaller one.}\label{fig:dataset}
\end{figure}

\noindent \textbf{Dataset.}
We consider a generic dataset \(\mathcal{D}\) consisting of \(D\) pairs \(\{(\boldsymbol{x}^\alpha, \boldsymbol{y}^\alpha)\}_{\alpha \in \{1,\ldots,D\}}\), with input \(\boldsymbol{x}^\alpha \in \mathbb{R}^{n_{\mathrm{in}}}\) and output \(\boldsymbol{y}^\alpha \in \mathbb{R}^{n_{\mathrm{out}}}\); a pair of inputs from \(\mathcal{D}\) is denoted \((\boldsymbol{x}^\alpha, \boldsymbol{x}^\beta)\), with \(\alpha, \beta \in \{1, \ldots, D\}\). For each input \(\boldsymbol{x}^\alpha\), the associated sequence of pre-activations is denoted \((\mathbf{z}_\ell^\alpha)_{1 \leq \ell \leq L}\). Unlike some prior work (e.g., \citep{Spectra-NTK-CK}), we focus on finite datasets without additional distributional assumptions. We further restrict attention to datasets with high-dimensional input vectors. As illustrated in Figure~\ref{fig:dataset}, this is in fact the generic situation: even when the raw data is low-dimensional, it is typically lifted to a higher-dimensional latent input space by a first layer (Fig.~\ref{fig:dataset}b), so that our high-dimensional assumption also covers this case once applied after the lifting step, on top of the case where the raw data is already high-dimensional (Fig.~\ref{fig:dataset}a).
\\

\noindent \textbf{Input, output, and hidden dimensions.} 
Regarding the dimensionality, we assume a uniform hidden width $n_\ell = N$ for all hidden layers $\ell \in [1,L-1]$. The layer width $n_\ell$ verifies: 
\[
n_\ell =
\begin{cases}
n_{\mathrm{in}} = N, & \ell=0,\\
n_{\mathrm{out}} < N, & \ell=L, \\
N, & 1 \leq \ell \leq L-1,
\end{cases} 
\]
so that
$\Psi_{\theta}: \mathbb{R}^{n_{\mathrm{in}}} \to \mathbb{R}^{n_{\mathrm{out}}}$, where $n_{\mathrm{in}}, n_{\mathrm{out}} $ are the problem input and output dimensions. 
The input and output dimensions $n_\mathrm{in}$ and $n_\mathrm{out}$ play an important role in the behavior of neural networks, especially in large-scale regimes. As explained above we assume throughout the paper that the input dimension is large,
 and we will consider the simplification $n_\mathrm{in}=N$, lifting to higher dimensions if necessary. For the output dimension, we consider two scaling regimes:
\begin{itemize}
  \item \textbf{Proportional scaling:} $n_\mathrm{out}= \mathcal{O}(N)$. This regime is relevant for large-scale architectures such as \emph{Large Language Models} and \emph{Vision Transformers}, where input, output, and hidden dimensions are all large.
  \item \textbf{Fixed output dimension:} $n_\mathrm{out}=\mathcal{O}(1)$. This regime covers tasks such as \emph{classification} with a finite number of classes and regression with a small number of outputs, including \emph{Implicit Neural Representations}.
\end{itemize}

\noindent \textbf{Activation function.}
We consider a class of activation functions $\phi$, including \texttt{tanh} and \texttt{sine}, satisfying four assumptions, stated in full as \textbf{H1}--\textbf{H4} in Appendix~\ref{app:activation}. In particular, $\phi$ is bounded and differentiable on $\mathbb{R}$, with Taylor expansion around the origin:
\begin{equation}
  \phi(x)
  \underset{x\to 0}{=}
  \sum_{p\geq 1}\frac{\phi_p}{p!}x^p,
  \qquad \phi_1>0,
  \label{eq:activation-expansion}
\end{equation}
where $\phi_p\in\mathbb{R}$. We set $\phi_0=0$, as a constant term is
redundant with the bias $\rvb_\ell$, and assume $\phi_1>0$ without loss
of generality. In the large-width and large-depth regimes considered
below, the leading behaviour is governed by the first coefficients
$\phi_1,\phi_2,\phi_3$ \citep{PDLT-2022}. The remaining assumptions are similarly mild, but more technical, and are deferred to Appendix~\ref{app:activation} for readability; together, they still cover a broad class of activation functions and are sufficient to establish new existence -- and, in some cases, uniqueness -- results for the fixed points of the network variance and covariance recursions.\\

\noindent \textbf{Initialisation procedures.} We consider two classical initialization schemes for the weights and biases, parameterized by \(\sigma_w\) and \(\sigma_b\), which control their variances. For both schemes, the biases are initialized as
$\rvb_\ell\sim\mathcal{N}(0,\sigma_b^2\mathbf{I}_{n_\ell})$, while the weights are
initialized according to
\begin{align}
\textbf{Gaussian initialization:}\quad
&\rmW_\ell\sim
\mathcal{N}(0,\sigma_w^2 \mathbf{I}_{n_\ell \times n_{\ell - 1}}),\label{eq:init-proc-gaussian}
\\
\textbf{Orthogonal initialization:}\quad
&\rmW_\ell=\sigma_w\sqrt{N}\,\rmQ_\ell,
\qquad
\rmQ_\ell\rmQ_\ell^\top=\rmI_{n_\ell}.\label{eq:init-proc-ortho}
\end{align}
Under orthogonal initialization, each square matrix $\rmQ_\ell$ is obtained from the QR decomposition of an independent random Gaussian matrix $\rmA_\ell\in\mathbb{R}^{N\times N}$ sampled as \eqref{eq:init-proc-gaussian}. The factor $\sqrt{N}$ accounts for the normalization used in~\eqref{eq:pre-post-activation}. Since the output weight matrix $\rmW_L\in\mathbb{R}^{n_{\mathrm{out}}\times N}$ is generally rectangular, we assume $n_{\mathrm{out}}\leq N$ and construct $\rmQ_L$ by retaining $n_{\mathrm{out}}$ rows of an $N\times N$ orthogonal matrix. Thus, $\rmQ_L$ is semi-orthogonal with orthonormal rows. Consequently, under orthogonal initialization,
\begin{equation}
\label{eq:orthogonality-weights}
    \frac{1}{N}\rmW_\ell\rmW_\ell^\top
    =\sigma_w^2\rmI_{n_\ell},
    \qquad 1\leq\ell\leq L.
\end{equation}
Some useful and interesting consequences of this property on network behavior have been highlighted in previous work \cite{saxe2014exact,dynamical_iso}. It will also play a central role in our study. Results obtained in the Gaussian initialization framework can easily be generalized to a much wider class of weight distributions, assuming only zero mean and finite variance $\sigma_w^2$. It thus includes the commonly used uniform weight distribution.

\subsection{Covariance, Jacobian and Neural Tangent Kernel: definitions}

We now define the three macroscopic observables -- correlation, Jacobian, and NTK -- that jointly characterise information propagation, gradient propagation, and parameter sensitivity through the network, and that we relate to one another in the sections that follow.
\begin{itemize}
  \item For a network of width $N$, the empirical covariance of preactivations at layer $\ell$, $\widehat K_{\ell}^{\alpha \beta}$ and the corresponding correlation or cosine similarity $\widehat{\rho}^{\, \alpha \beta }_{\ell}$ are respectively defined as:
\begin{equation}
\widehat K_{\ell}^{\alpha \beta}
=\widehat K_{\ell}(\boldsymbol{x}^\alpha,\boldsymbol{x}^\beta)
:= \frac{1}{N} {\rvz_{\ell}^{\alpha}}^{\top} \rvz_{\ell}^{\beta}
\label{eq:empirical_covariance}
\end{equation}
and 
\begin{equation}
\widehat{\rho}^{\, \alpha \beta }_{\ell}:= \frac{ \widehat{K}^{\alpha \beta}_{\ell} }{\sqrt{ \widehat{K}^{\alpha \alpha}_{\ell} \widehat{K}^{\beta \beta}_{\ell}}}.\label{eq:empirical_correlation}
\end{equation}
Their infinite width counterparts will be denoted $K^{\alpha \beta}_\ell$ and $\rho^{\alpha \beta}_\ell$. Our aim is to guarantee that information stored in input correlation can be propagated when $\ell$ goes to infinity. 
  \item The layer-wise and end-to-end Jacobian matrix, denoted $\rmJ_\ell^{\alpha} \in \mathbb{R}^{n_\ell \times n_{\ell - 1}}$ and $\mathcal{J}^{\alpha} \in \mathbb{R}^{n_\mathrm{out} \times n_\mathrm{in}}$ (with an superscript index $\alpha$ added when dependency on a specific input needs to be remembered), are respectively defined as:
   \begin{equation}\label{eq:ef_jac}
  \rmJ_{\ell}^{\alpha}:= \frac{\partial \rvz_{\ell} }{\partial \rvz_{\ell-1}}(\boldsymbol{x}^\alpha) =   
 \frac{1}{\sqrt{n_{\ell-1}}}\rmW_{\ell }\;\text{diag}[\phi'(\rvz_{\ell - 1 }^\alpha)]
  \end{equation}
  and
  \begin{equation}\label{eq:ef_jacfull}
  \mathcal{J}_L^{\alpha}:=\prod_{\ell=0}^{L-1}\rmJ_{L-\ell}^{\alpha} .
\end{equation}
Given this product structure, a natural requirement is that the end-to-end Jacobian neither vanish nor explode exponentially with depth. We will nevertheless see that correlation propagation implies its mean eigenvalue to decay algebraically.
  \item  The empirical Neural Tangent Kernel \citep{jacot2018neural} is defined for the dataset \(\mathcal{D}\) of size $D$ as a Gram matrix   of size $n_\mathrm{out}D \times n_\mathrm{out} D$ composed of $ D^2$ block matrices $\widehat \Theta^{\alpha\beta}_L\in \mathbb{R}^{n_\mathrm{out} \times n_\mathrm{out}}$: 
  \begin{equation}
    \label{eq:NTK-def}
    \widehat{\Theta}_L =( \widehat{\Theta}_{L}^{\alpha\beta})_{\alpha, \beta \in \{1,\ldots,D\}} \qquad \mbox{with} \qquad \widehat\Theta_{L}^{\alpha\beta } = \frac{\partial \rvz_{L}^\alpha}{\partial \theta} \left(\frac{\partial \rvz_{L}^\beta}{\partial \theta}\right)^{\top},
    \end{equation}

where $\partial(\cdot)/\partial \theta$ denote gradients with respect to the full parameter space. Their infinite width limit will be denoted $\Theta_L$ and $\Theta^{\alpha \beta}_\ell$. Our study is developed in an asymptotic regime where the NTK matrix, set at initialization controls the entire training of the network. In that case, the NTK spectrum sets the network's ability to rapidly converge to an optimal solution (see Sec.~\ref{sec:ntk}). Our objective will therefore be to control its spectrum at initialization. 
\end{itemize}

For our analysis, we will consider first and second moments, as well as their standard deviation. We recall that the empirical order-$k$ moment of a square random matrix $\boldsymbol{\rmX} \in \mathbb{R}^{n\times n}$, and its average over the random initialisation of the network's weights and biases -- i.e.\ the expectation $\mathbb{E}[\cdot]$ taken over all possible weight and bias configurations -- are denoted respectively by
\begin{equation} \label{eq:moments-definition}
  \widehat{m}^{(k)}_{\boldsymbol{\rmX}} = \frac{1}{n} \Tr \boldsymbol{\rmX}^k , \quad m^{(k)}_{\boldsymbol{\rmX}} = \frac{1}{n} \mathbb{E}\left[\Tr \boldsymbol{\rmX}^k \right].
\end{equation}
In this article, we will use both interchangeably since, in the sequential limit, by the law of large numbers, given that enough moments of the entries of $\rmX$ exist, the empirical moments concentrate around their expectation. It will also be useful to describe the corresponding standard deviation and relative variance, defined respectively as
\begin{equation}
 \mathbb{S}[\boldsymbol{\rmX}] = \sqrt{m^{(2)}_{\boldsymbol{\rmX}} - \left(m^{(1)}_{\boldsymbol{\rmX}}\right)^2} \quad \text{and} \quad \mathbb{V}[\boldsymbol{\rmX}] = \frac{m_{\boldsymbol{\rmX}}^{(2)}}{\left(m_{\boldsymbol{\rmX}}^{(1)}\right)^2} - 1. \label{eq:rel-variance-def}
 \end{equation}

\subsection{Aim of this work}

Having introduced the necessary notations, we can now state the objective of this work: to characterize the \underline{\emph{initialization strategies}} (see Sec.~\ref{sec:architecture}) and \underline{\emph{parameters}}: 
\[
\{\sigma_w,\sigma_b,\phi_1,\phi_2,\phi_3\},
\]
where \(\phi_i\) denote the low-order coefficients of the activation function's Taylor expansion such that, in the deep network limit, the correlation map preserves non-trivial input dependence while the end-to-end Jacobian \(\mathcal{J}_L\) and the NTK matrix \(\widehat{\Theta}_L\) exhibit well-defined asymptotic behavior. 




\section{Mean-field limit, fixed points and the need for criticality}
\label{sec:meanfield}

The aim of this section is to recall useful results on neural networks in the infinite-width limit $N\rightarrow \infty$, and important properties of the empirical pre-activation covariance kernel and its depth evolution as the network depth grows to infinity. Throughout this paper, we refer to the infinite width limit, together with its associated Gaussian description, as the \emph{mean-field} regime\footnote{This use of \emph{mean field} should not be confused with the \emph{mean-field} description of training dynamics, where infinitely many neurons are described through an evolving distribution over parameters \citep{mei2018mean,rotskoff2022trainability}. These two infinite-width descriptions correspond to different choices of parameterization and learning-rate scaling \citep{chizat2019lazy}.}. Most of this section follows \cite{bahri2024houches} and \cite{PDLT-2022}. It serves to introduce the key concepts, notations, and the first constraints appearing on the parameters $\sigma_b$ and $\sigma_w$. 


\subsection{Large width limit: empirical covariance is a Gaussian kernel}






In the large width limit $N \rightarrow \infty$, the pre-activation $\rvz_{\ell}$ entries converge in distribution to a Gaussian process, with zero-mean and covariance matrix $\mSigma_{\ell}^{\alpha \beta}$. More precisely, for two inputs $(\boldsymbol{x}^\alpha,\boldsymbol{x}^\beta)$,
\begin{equation}
(z^\alpha_{\ell},z^\beta_{\ell})
\sim \mathcal{N}\!\left(0, \mSigma_\ell^{\alpha \beta}\right)\quad \text{with} \quad \mSigma_{\ell}^{\alpha \beta} =
\begin{bmatrix}
K_\ell^{\alpha \alpha} & K_\ell^{\alpha \beta} \\
K_{\ell }^{\alpha \beta} & K_{\ell }^{\beta \beta}
\end{bmatrix}.
\end{equation}
 and covariance kernel 
\begin{equation}
\label{eq:covkernel}
K_\ell^{\alpha\beta} = \mathbb{E}[z^\alpha_{\ell}z^\beta_{\ell}].
\end{equation}
Here and later in the paper, we use the notation $(z_\ell^{\alpha},z_\ell^{\beta})=(\rvz_{\ell,i}^{\alpha},\rvz_{\ell,i}^{\beta})$ in situations where the value of $i$ does not matter. Furthermore, the empirical covariance at layer $\ell$ defined in \eqref{eq:empirical_covariance} concentrates on the covariance kernel: \[\lim_{N\rightarrow \infty} \widehat{K}^{\alpha \beta}_\ell= K^{\alpha \beta}_\ell.\]
In other words, the empirical covariance computed for a given choice of weights and biases concentrates to the ensemble average over all possible weight and bias choices, defined by $\mathbb{E}[\cdot]$. Expanding, the covariance \eqref{eq:covkernel} and taking the average relative to the weights and biases one can write (cf. Lecture \citep{bahri2024houches}):
\begin{equation}
K_{\ell+1}^{\alpha\beta}
=
\sigma_w^2
\mathbb{E}\left[\phi\left(z_{\ell}^\alpha\right)\phi\left(z_{\ell}^\beta\right)\right]
+
\sigma_b^2 .\label{eq:rec_compacte}
\end{equation}

\noindent \textbf{Ensemble averages and Gaussian averages.} Given a fixed $\ell$, in the infinite-width limit, each pre-activation $\rvz_\ell$ converges to a Gaussian process indexed by the inputs. Then, \emph{ensemble averages} involving $\rvz_\ell$ can be reformulated using \emph{Gaussian averages} with covariance matrix  \(\mSigma_{\ell}^{\alpha \beta}\) for two different inputs $\boldsymbol{x}^\alpha \neq \boldsymbol{x}^\beta$, or variance \(K^{\alpha \alpha}_\ell\) for a single input $\boldsymbol{x}^\alpha$. Thus, one can reformulate the recursion of the variance as:

\begin{equation}\label{eq:rec-variance}
  K^{\alpha \alpha}_{\ell + 1} = \frac{\sigma_w^2}{\sqrt{2\pi K^{\alpha \alpha}_{\ell}}} \int \, \phi(u)^2 e^{ -u^2 / 2 K^{\alpha \alpha}_\ell} du + \sigma_b^2.
\end{equation}
 Similarly, for the off-diagonal case $\alpha \neq \beta$, \eqref{eq:rec_compacte} can be rewritten: 
\begin{equation} \label{eq:rec-covariance}
  K_{\ell + 1}^{\alpha \beta } =\frac{\sigma_w^2}{\sqrt{ 2\pi \det{\mSigma_\ell}^{\alpha\beta}}}\int \, \phi(u)\phi(v) e^{-\frac{1}{2}(u, v)\left(\mSigma_\ell^{\alpha\beta}\right)^{-1} (u, v)^\top} du dv
+\sigma_b^2 .
\end{equation}
Given the equivalence of these averages in the sequential limit we will use them interchangeably to ease the reading process. An introductory section about finite-size effect that allows to understand to which extent this claim is valid is provided in Sec.~\ref{sec:finite-size-effect}.\\

This recurrence relation will play a central role in the remainder of this paper and was first derived in \cite{poole2016exponential}; see also \cite{bahri2024houches,PDLT-2022}. 
For the sine activation function, it admits an analytical form \citep{combette2025new}:
\begin{equation}
\label{eq:sine_rec_rel}
  K^{\alpha \beta}_{\ell + 1} = \sigma_w^2 e^{-\tfrac{1}{2}(K^{\alpha \alpha}_{\ell} + K^{\beta \beta}_{\ell})} \sinh \left( K^{\alpha \beta}_{\ell} \right) + \sigma_b^2,
\end{equation}
that will be used throughout the text to illustrate results obtained in the paper for a broader class of activation functions.\\

\noindent \textbf{Recursion relation.} In the following we will consider the reduced covariance vector $(K_{\ell}^{\alpha \beta }, K_{\ell}^{\alpha \alpha }, K_{\ell}^{\beta \beta })$ and the full covariance matrix $\mSigma_\ell^{\alpha\beta}$ indifferently. Equations \eqref{eq:rec-covariance} and \eqref{eq:rec-variance} define a recurrence relation for the covariance matrix: 
\begin{equation}\label{eq:kernel-func_rec}
  \mSigma_{\ell+1}^{\alpha\beta} = \mathscr{F}(\mSigma_{\ell}^{\alpha\beta}).
\end{equation}
Given this recurrence relationship we introduce the notation $\mSigma^{\alpha \beta}_{\star}$ to denote a fixed point of \eqref{eq:kernel-func_rec}, whose existence, uniqueness, stability, and basin of convergence will be discussed in the next section.\\

\noindent \textbf{Jacobian and stability.}
In the remainder of this paper, important network properties will depend on the Jacobian of the covariance map
\begin{equation}
\label{eq:jaccov}
  \mathbf{J}_\mathscr{F}(\mSigma_{\ell}^{\alpha \beta})=  \frac{\partial \mSigma_{\ell + 1}^{\alpha \beta}}{\partial \mSigma_{\ell}^{\alpha \beta}} = 
  \begin{pmatrix}
    \chi^{\perp}_{\ell} & \chi^{\bullet}_{\ell} & \chi^{\circ}_{\ell} \\
   0 & \chi^{\alpha}_\ell & 0 \\
   0 & 0 & \chi^{\beta}_\ell 
  \end{pmatrix},
\end{equation}
with \begin{equation}\chi^{\perp}_\ell:= \frac{\partial K^{\alpha \beta}_{\ell + 1}}{\partial K^{\alpha \beta}_{\ell}}, \quad \chi^{\alpha}_{\ell}:= \frac{\partial K^{\alpha \alpha}_{\ell + 1}}{\partial K^{\alpha \alpha}_{\ell}}, \quad \chi^{\beta}_{\ell}:= \frac{\partial K^{\beta \beta}_{\ell + 1}}{\partial K^{\beta \beta}_{\ell}}, \quad \chi^{\bullet}_{\ell}:= \frac{\partial K^{\alpha \beta}_{\ell + 1}}{\partial K^{\alpha \alpha}_{\ell}}, \quad \chi^{\circ}_\ell:= \frac{\partial K^{\alpha \beta}_{\ell + 1}}{\partial K^{\beta \beta}_{\ell}}. \label{eq:allchi_def} 
\end{equation}
This Jacobian matrix $\mathbf{J}_\mathscr{F}(\mSigma_{\ell}^{\alpha \beta})$ is interpreted as a susceptibility matrix, extending the terminology introduced in \cite{PDLT-2022}. The expressions of the susceptibilities are given in the Appendix~\ref{app:susceptilibilities} and will be useful for the rest of the paper. In particular, when evaluated at the fixed point $\mSigma_\star^{\alpha\beta}$, it governs the stability of this point. As a reminder, a fixed point $\mSigma_{\star}^{\alpha \beta}$ is said to be stable
if its Jacobian covariance matrix $\mathbf{J}_\mathscr{F}(\mSigma_{\star}^{\alpha \beta})$ has a spectral radius (the maximum absolute eigenvalue) strictly smaller than 1. Since \eqref{eq:jaccov} is an upper triangular matrix it is equivalent to: 
\begin{equation}
  \chi^{\perp}_{\star} < 1,
  \qquad
   \chi^{\alpha}_{\star} < 1, \qquad \chi^\beta_\star < 1,
\end{equation}
where we denoted by a ``$\star$'' the limiting value of the respective entries. 

\subsection{Large depth limit: fixed points, edge of chaos and criticality \label{sub-section:large-depth-limit}}

In this section, we consider $\mSigma_{\star}^{\alpha \beta}$ a fixed point of \eqref{eq:kernel-func_rec}, i.e.,
\begin{equation} \label{eq:gaussian-kernel-limit}
  \mSigma_\star^{\alpha\beta}= \begin{pmatrix}
  K_\infty^{\alpha\alpha} &K_\infty^{\alpha\beta} \\ K_\infty^{\beta\alpha} &K_\infty^{\beta\beta} 
\end{pmatrix}
\end{equation}
 for a specific choice of input $\alpha, \beta \in \{1, \dots, D\}$ and hyper-parameters $(\sigma_w, \sigma_b)$ with activation function $\phi$ and discuss consequences for the flow of correlation at large network depth. We will pay particular attention to the dependence of the limiting correlation on the initial conditions.\\

\noindent \textbf{Properties of the fixed point $\mSigma_{\star}^{\alpha \beta}$.} The existence of such a fixed point was tackled in \cite{hayou2019impact} but its uniqueness and stability were never really discussed for the whole $(\sigma_w, \sigma_b)$ plane, except for the diagonal elements in the sinusoidal case \citep{combette2025new} and it is, most of the time, implicitly assumed \citep{poole2016exponential, schoenholz2017deep, dynamical_iso}. In Appendix~\ref{app:fixed-point-discussion}, we provide a detailed analysis of the existence, uniqueness, stability, and basins of convergence of the kernel recursion \eqref{eq:kernel-func_rec} fixed points, under the assumptions on the activation function stated in Sec.~\ref{sec:architecture}. We show that the variance recursion \eqref{eq:rec-variance} admits a unique globally attracting fixed point on $\mathbb{R}^{+*}$ denoted $K_\star$. For the covariance recursion \eqref{eq:rec-covariance}, we show that there is a finite number of fixed points whose basins of convergence cover the admissible covariance interval. Consequently, for every non-zero pair of inputs $(\alpha,\beta)$, there exists a limiting covariance $C_\star$ such that:
\begin{equation}
\label{eq:gaussian-kernel}
\mSigma_\ell^{\alpha\beta}
\underset{\ell\to\infty}{\longrightarrow}
\mSigma_\star^{\alpha\beta}
= \begin{pmatrix} K_{\star} & C_{\star} \\ C_{\star} & K_{\star} \end{pmatrix}.
\end{equation}

Let us remark that $K_{\star}$ and $C_{\star}$ do not depend on $(\alpha, \beta)$, at least for the basin of convergence considered.
By contrast, we will prove that the limiting behaviour of the correlation $\rho_\infty^{\alpha\beta}$ defined as:
\begin{equation} 
   \rho^{\alpha \beta }_{\infty} = \lim_{\ell \to \infty} \rho^{\alpha \beta }_{\ell} \quad \text{with} 
   \quad \rho^{\alpha \beta }_{\ell}= \frac{ K^{\alpha \beta}_{\ell} }{\sqrt{ K^{\alpha \alpha}_{\ell} K^{\beta \beta}_{\ell}}}
\label{eq:rho}
\end{equation} 
may not always share this independence property. Indeed, we will show that the limiting correlation can remain a function of the network inputs \(( \boldsymbol{x}^\alpha, \boldsymbol{x}^\beta)\) for an appropriate choice of \((\sigma_w, \sigma_b)\). This is a central contribution of the following section.\\

\begin{figure}
  \centering
  \includegraphics[width = \linewidth]{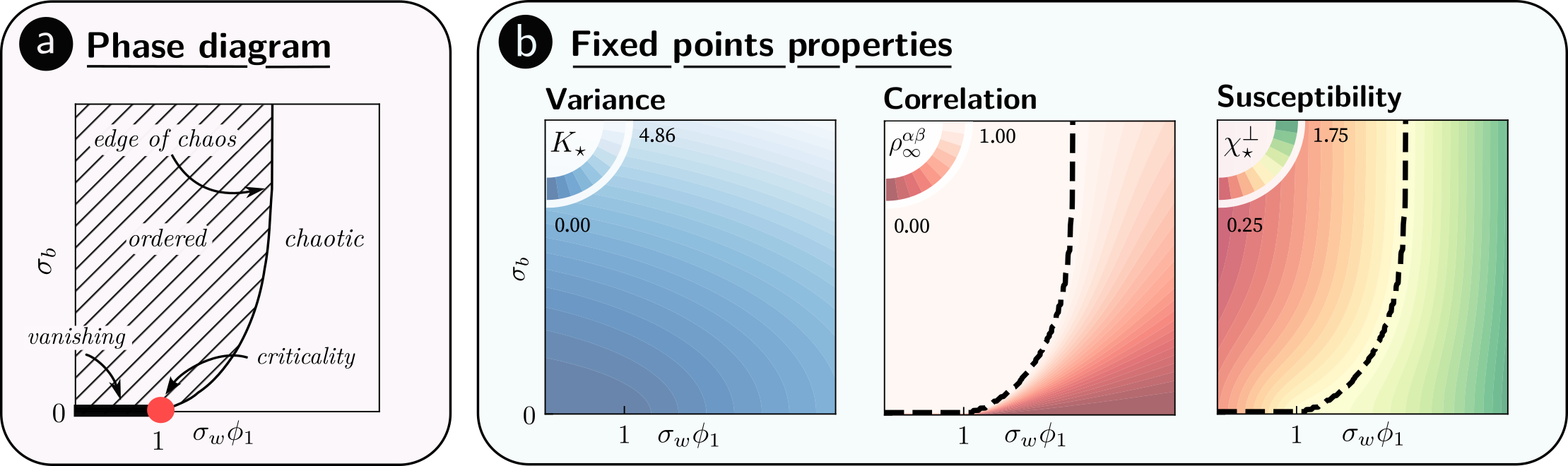}
  \caption{\textbf{The large depth limit.} \textbf{a)} Phase diagram in the $(\sigma_w, \sigma_b)$ parameter space, including \textit{ordered, chaotic}, and \textit{vanishing variance} phases, as well as the \textit{critical point}. \textbf{b)} Fixed point behaviour of the variance $K_{\star}$, the correlation $\rho^{\alpha \beta}_{\infty}$ and the perpendicular susceptibility $\chi^{\perp}_{\star}$ in the $(\sigma_w, \sigma_b) $ parameter space for a sinusoidal activation function. Computations are performed using \eqref{eq:sine_rec_rel}.}
  \label{fig:phase-diagram}
\end{figure}

\noindent \textbf{Correlation behaviour in the large depth limit.} Typically, two different regimes are discussed (see Fig.~\ref{fig:phase-diagram}): \textit{chaotic} ($\rho_\infty^{\alpha\beta}<1$) and \textit{ordered} ($\rho_\infty^{\alpha\beta}=1$) \citep{schoenholz2017deep,poole2016exponential}. Here, we further specify the condition of existence of these two phases leading to the consideration of a third regime that we call \textbf{vanishing variance} phase. More precisely the three regimes are defined as follows:\\
(i)  The \textbf{ordered phase}, when \(C_{\star}=K_{\star}>0\) leads to \(\rho_\infty^{\alpha\beta}=1\). In that case, two different inputs collapse on the same output at large depth.\\ 
(ii) The \textbf{chaotic phase}, when \(0\le |C_{\star}|< K_{\star}\) leads to \(0\le \rho_\infty^{\alpha\beta}<1\) which is independent of \(\alpha\) and \(\beta\). In that case, two different inputs, no matter how close they are, separate and reach a correlation that is independent of the initial separation.\\ 
(iii) The \textbf{vanishing variance phase}, when \(K_{\star}=0\), which implies $C_{\star}=0$. In such a situation, the ratio \(C_{\star}/K_{\star}\) is undetermined, as both $C_{\star}$ and $K_{\star}$ vanish. However, it is possible to study the limiting behaviour of \(\rho_\ell^{\alpha\beta}\), as done in the next paragraph.\\
\\

\noindent \textbf{Location of the vanishing variance phase.} Phase (iii) requires $\sigma_b = 0$, since a non-zero bias variance 
always generates a strictly positive fixed point $K_{\star} > 0$. This is a direct consequence of the recurrence relation \eqref{eq:rec_compacte} and it is illustrated  in Fig.~\ref{fig:phase-diagram}b. On the line $\sigma_b = 0$, the solution $K_{\star}=0$ is always a fixed point, given that $\phi(0)=0$. The extent of the vanishing variance phase along $\sigma_w$ is governed by the stability of the fixed point $K_\star=0$. The stability is controlled by the limiting parallel susceptibility:
\begin{align}\label{eq:chi_par}
\chi^{\parallel}_\star:=\chi^{\alpha}_\star=\chi^{\beta}_\star, 
\end{align}
evaluated at the fixed point, as discussed for instance in Chapter 5 of \cite{PDLT-2022}. When the fixed point is $K_\star=0$, a direct computation using the susceptibility expressions \eqref{eq:chi_terms} shows that 
\begin{align}\label{eq:chi_parBIS}
\chi^{\parallel}_\star =\sigma_w^2 \phi_1^2.
\end{align}
The fixed point $K_\star=0$ is stable when $\sigma_w<1/\phi_1$, thus defining the location of the vanishing variance phase. For $\sigma_w>1/\phi_1$, the fixed point $K_\star=0$ becomes unstable, and the network variance escapes to a strictly positive fixed point, placing the network in the chaotic phase.\\

\noindent \textbf{Edge of chaos line.} Away from the line $\sigma_b=0$, the network falls into either the ordered or the chaotic phase. The transition line between the two, for the correlation in parameter space $(\sigma_w,\sigma_b)$ is called the \textit{edge of chaos} line. It can be identified by analysing properties of the fixed point $C_{\star}$. We note that the covariance matrix $\mSigma_{\star}$ obtained by setting $C_{\star}=K_{\star}$ is always a fixed point for $\mSigma^{\alpha\beta}_\ell$. Following \cite{poole2016exponential}, the edge of chaos line is then found by identifying the change of stability of $C_{\star}$ at $\mSigma^{\alpha\beta}_\ell=\mSigma_{\star}$. The stability of this fixed point is governed by the perpendicular susceptibility defined in \eqref{eq:allchi_def}, and the edge of chaos line is obtained by computing: 
\begin{equation}\chi_\star^\perp=\lim_{\ell\rightarrow +\infty}\sigma_{w}^2 \E \left[ \phi'^2\left(z_\ell^{\alpha}\right) \right] = 1. \label{eq:chi_perp_def}\end{equation}

\noindent \textbf{Critical point.} The limiting point of the \textit{vanishing variance phase}
 \begin{equation}
 (\sigma_w,\sigma_b)=\left(\frac{1}{\phi_1},0\right)\label{eq:criticality},
 \end{equation}
also intersecting the \textit{edge of chaos line}, plays a particular role. It is a critical point, as it corresponds to a regime of marginal stability, defined by 
\begin{equation}
\chi^\parallel_\star= 1,\quad \chi^\perp_\star= 1.
\end{equation}
The importance of this point for network initialisation has been stressed several times, notably in \cite{PDLT-2022}, and our paper will be dedicated to developing further consequences of this choice. To understand why this is the only practical choice for propagating information to infinite depth, we first examine the convergence rate at this point, and return to the justification below.\\

\begin{figure}[ht]
  \centering
  \includegraphics[width = \linewidth]{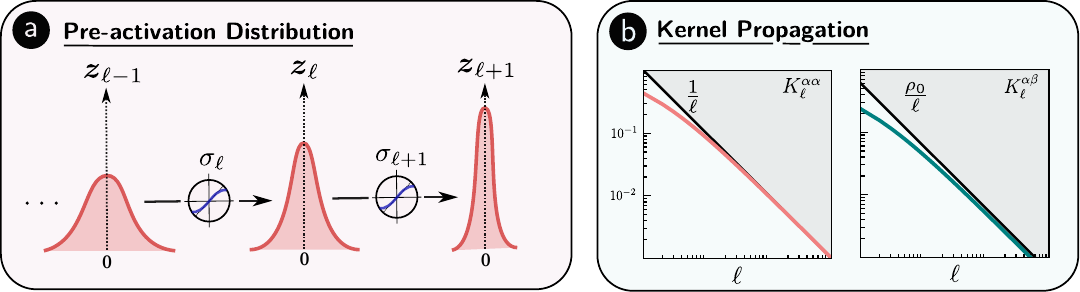}
  \caption{\textbf{Evolution of network properties with depth at criticality.} \textbf{a)} Schematic illustration of the contraction of the pre-activation distribution \(\rvz_{\ell}\) around \(0\), leading to quasi-linear behaviour at large depths. \textbf{b)} Universal algebraic decay of \(K^{\alpha\beta}_{\ell}\), shown for both \(\alpha=\beta\) and \(\alpha\neq\beta\), for the sine activation using the kernel recursion~\eqref{eq:sine_rec_rel}.}
  \label{fig:critical_decay}
\end{figure}

\noindent \textbf{Convergence rate.} Depending on the parameter choice, three regimes of convergence are observed: (i) algebraic behaviour for both covariance and variance at criticality, as illustrated in Fig.~\ref{fig:critical_decay}; (ii) algebraic convergence for covariance and exponential convergence for variance on the edge of chaos line; and (iii) exponential convergence for both covariance and variance everywhere else. Regimes (ii) and (iii) were described in \cite{poole2016exponential}, whereas regime (i) is described in depth in \cite{PDLT-2022}.  Important consequences of this algebraic decay will be unveiled in this paper.\\

\noindent \textbf{The need for initialisation at criticality.} As established above, a necessary condition to propagate correlation through the network (i.e., to keep the dependency of $\rho_\infty^{\alpha\beta}$ on $(\alpha,\beta)$) is to choose an initialisation such that $K^*=0$, i.e., either in the \textit{vanishing variance phase} or at its limiting \textit{critical point}. Indeed, in those cases, the correlation may tend to a non-trivial value if the variance and covariance converge at the same rate, whether algebraic or exponential. In practice, the regime of exponential decay needs to be avoided, as the numerator and denominator in the correlation expression both become numerically indistinguishable from zero after a few layers (as illustrated in the last panel of Fig.~\ref{fig:propagation_correlation}c). Hence, practical control of correlation requires the algebraic decay regime.\\

To conclude, initialisation at criticality is not only important due to the slow decay toward the fixed point, as emphasized in previous work \citep{poole2016exponential,PDLT-2022}, but also because the limiting correlation remains a function of the input correlation. This critical behaviour is detailed in the next section, and the remainder of the paper explores its consequences.

\section{Flow of correlation at criticality} \label{sec:flow-correlation}

The aim of this section is to revisit the algebraic decay of the pre-activation variance $K^{\alpha\alpha}_\ell$, established in \cite{PDLT-2022}, and to derive the behaviour of the covariance $K^{\alpha\beta}_\ell$ at criticality. We then obtain an explicit expression for the correlation dynamics which, to the best of our knowledge, is new. We show that avoiding correlation collapse requires imposing $\phi_2=0$. Finally, we demonstrate that orthogonal initialisation is crucial for preventing ill-conditioned finite-size effects in the correlation flow through the network.

\subsection{Universal behaviour of network covariance at criticality \label{sub:universal-behaviour}}

Considering \eqref{eq:rec_compacte} at criticality (cf. \eqref{eq:criticality}) and expanding the product of activation functions expressed with their Taylor expansion \eqref{eq:activation-expansion} yields 
\begin{equation} \label{eq:covariance-recursion}
   K_{\ell + 1}^{\alpha \beta } = \left[K_{\ell}^{\alpha \beta } + (\tfrac{\phi_{2}}{2\phi_{1}})^2 \left( K^{\alpha \alpha }_{\ell}K^{\beta \beta }_{\ell }+ 2 (K^{\alpha \beta }_{\ell})^2\right) + \tfrac{\phi_{3}}{2\phi_{1}}K^{\alpha \beta}_{\ell}\left( K^{\beta \beta }_{\ell } + K^{\alpha \alpha }_{\ell } \right) + \mathcal{O}(K^3)\right]
\end{equation}
where $\mathcal{O}(K^3)$ stands for all the terms of order 3 in the covariance coefficients $K^{\alpha \beta}_{\ell}$. 
This relation generalises the recursive relation established in \cite{PDLT-2022}, for $\alpha=\beta$, to the case where $\alpha\neq \beta$. In the large $\ell$ limit, the asymptotic solution of \eqref{eq:covariance-recursion} when $\alpha=\beta$ is 
\begin{equation}
  K^{\alpha \alpha}_{\ell} = \frac{1}{ A \ell} + \mathcal{O}\left(\frac{\log \ell}{\ell^2}\right)\, \quad \text{with}  \quad A =-\left( \frac{\phi_3}{\phi_1} + \tfrac{3}{4} \left(\frac{\phi_2}{\phi_1} \right)^2\right) . \label{eq:def_a1}
\end{equation} 
This algebraic behaviour with an exponent independent of the activation function was referred to as the $K_\star=0$ universality class by \citep{PDLT-2022}. We see that the fixed point $K_\star = 0$ at criticality is attractive with algebraic decay provided that $A>0$, which implies a condition on $(\phi_2,\phi_3)$:
\begin{equation}
  \text{Stability of $K_*=0$ at criticality if } \phi_3<-\frac{3}{4}\frac{\phi_2^2}{\phi_1}.\label{eq:condKstable_phi3}
 \end{equation}
Since we assumed  an activation function with $\phi_1>0$, the previous inequality implies $\phi_3<0$. This is actually a necessary condition of hypothesis (\textbf{H2}) that we unearth by considering simply the Taylor expansion of $\phi$ around 0.

\subsection{Asymptotic behaviour of the correlation}\label{sub:corr_meanfield}
While the common intuition might suggest that the choice $K_\star = 0$ is suboptimal since all inputs would map to zero at infinite depth, we will demonstrate instead that it \emph{preserves information} at any layer $\ell$, under specific conditions on the first terms of the activation function's Taylor expansion.

At criticality, correlation at layer $\ell+1$ can be expressed in terms of the correlation in the previous layer (combining \eqref{eq:rho} and \eqref{eq:covariance-recursion}, more details in Appendix~\ref{eq:corr_rec}):
\begin{equation} \label{eq:rho-map}
  \rho_{\ell + 1}^{\alpha \beta} = \rho_{0}^{\alpha \beta} + \sum_{\ell^\prime=1}^{\ell} \left(\tfrac{\phi_2}{\phi_{1}}\right)^2 \left[\left(\frac{1}{4 (\rho_{\ell^\prime}^{\alpha \beta })^2}+ \frac{1}{2}\right) K^{\alpha \beta }_{\ell^\prime} - \frac{3}{8}(K^{\beta \beta }_{\ell^\prime} + K^{\alpha \alpha }_{\ell^\prime} ) \right] + \eta^{\alpha\beta}_{\ell+1}
\end{equation}
where \(\eta^{\alpha\beta}_{\ell} \sim 1 \) is a convergent sum. This is a function of $\rho_0^{\alpha\beta}$, $K^{\alpha \alpha}_0$, and $K^{\beta \beta}_0$. The sum over $\ell^\prime$ in \eqref{eq:rho-map} involves terms of order $1/\ell^\prime$. This sum is divergent unless either $\phi_2=0$ or $\rho_\infty^{\alpha\beta} \in \{ 1/2, \,1\}$ which is independent of the initial correlation, leading 
to:
\begin{align} \label{eq:condition-phi2}
 \phi_2=0\quad &\Leftrightarrow \quad \rho_\infty^{\alpha\beta} = \rho_{0}^{\alpha\beta} + \eta_\infty^{\alpha\beta}(\rho_{0}^{\alpha\beta}, K_0^{\alpha \alpha}, K_0^{\beta \beta}), \\
 \phi_2\neq 0 \quad &\Leftrightarrow \quad\rho_\infty^{\alpha\beta}\in \{ 1/2, \,1\}.
\end{align}
We conclude that either $\phi_2=0$ and information can be propagated through the network, or $\phi_2 \neq 0$ and information is lost as the inputs collapse on the same output at large depth. Thus, the activation function must be an odd function at lowest order in its argument. This hypothesis is valid for common activation like \texttt{tanh} or \texttt{sine}.

\subsection{Finite-size effects and orthogonality} \label{sec:finite-size-effect}

The preceding sections treated the infinite-width limit $N\to\infty$, where the mean-field description is exact. In practice, networks are finite, and $1/N$ corrections introduce fluctuations in the empirical kernel that vanish in the mean-field limit. The structure of these finite-size perturbations is well established for Gaussian initialisation \citep{PDLT-2022,bahri2024houches,hanin_cumulants} and was recently extended to the orthogonal case \citep{Ortho-correlation}. We build on these results to identify the natural sources of such fluctuations and, at criticality, estimate their magnitude. Our predictions will be validated by the numerical simulations in Sec.~\ref{sub:numerical_experiment}. \\

\noindent\textbf{Relation between empirical kernel and mean-field solution.} In this section, we aim to characterize the empirical kernel denoted $\widehat{K}_\ell^{\alpha\beta}$. To connect it to the mean-field solution \(K^{\alpha \beta}_{\ell}\) defined in \eqref{eq:covariance-recursion}, we introduce a layer-specific perturbation 
\begin{equation}
   \delta_\ell^{\alpha \beta}:= \widehat{K}_\ell^{\alpha\beta}  -\; K^{\alpha \beta}_{\ell} .
   \label{eq:delta}
\end{equation}
We now turn to a closer examination of this layer-specific perturbation. To that end, it is convenient to notice that the empirical kernel can be rewritten as a perturbed recurrence along the off-diagonal mean field flow:
\begin{equation}
  \widehat{K}_{\ell+1}^{\alpha\beta} = \mathscr{F}_C(\widehat{K}_{\ell}^{\alpha\beta}, \widehat{K}_{\ell}^{\alpha\alpha}, \widehat{K}_{\ell}^{\beta\beta}) + \varepsilon_\ell^{\alpha\beta},
  \label{eq:reckernel}
\end{equation}
where $\mathscr{F}_C$ is defined by the mean field recurrence  \eqref{eq:rec-covariance} (see also \eqref{eq:covariance-restriction}), and where the perturbation is defined as: 
\begin{equation}
 \varepsilon_\ell^{\alpha\beta}:=  \underbrace{\sigma_w^2 \sum_{p, q \ge 0} \frac{\phi_p \phi_q}{q! p!}  \left[\widehat{M}_\ell^{\alpha \beta}(p,q) - M_\ell^{\alpha \beta}(p,q)\right] }_{\varepsilon_{\ell,\text{Wick}}^{\alpha\beta}}
  + \underbrace{\phi(\rvz^{\alpha}_\ell)^\top \left( \frac{\rmW_{\ell + 1}^\top \rmW_{\ell + 1}}{N} - \sigma_w^2\mathbf{I}\right)\phi(\rvz^{\beta}_\ell)}_{\varepsilon_{\ell,\text{Weight}}^{\alpha\beta}}.
    \label{eq:reckerneldet}
\end{equation}
Here
$\widehat{M}_\ell^{\alpha \beta}(p,q)$ and \(M_\ell^{\alpha \beta}(p,q)\) denote, respectively the empirical mixed moment of order $(p,q)$ and its Wick contraction in the Gaussian case, defined as:
\begin{equation}
  \widehat{M}_\ell^{\alpha \beta}(p,q) = \frac{1}{N} \sum_{i = 1}^{N} (z_{\ell,i}^{\alpha})^p (z_{\ell,i}^{\beta})^q \quad \mbox{and} \quad
  M^{\alpha \beta}_{\ell}(p,q) = \sum_{\pi \in \mathcal{P}^{\alpha \beta}(p,q)} \prod_{\{\gamma , \delta \} \in \pi} \widehat{K}^{\gamma \delta}_\ell.
\end{equation}
Equations~\eqref{eq:reckernel} and \eqref{eq:reckerneldet} are derived in Appendix~\ref{par:induced_err}. 
The first term, $\varepsilon_{\ell,\text{Wick}}^{\alpha\beta}$, accounts for Wick defects from the non-Gaussianity of \(\rvz_{\ell,i}^{\alpha}\), and the second term, $\varepsilon_{\ell,\text{Weight}}^{\alpha\beta}$, captures the effect of non-orthogonality of the weights.


We also derive and discuss in Appendix~\ref{par:error_prop} the recurrence relation at first order, yielding: 
\begin{align} \label{eq:recurrence-vec-delta}
   \begin{pmatrix}\delta_{\ell+1}^{\alpha \beta} \\
      \delta_{\ell+1}^{\alpha \alpha} \\
      \delta_{\ell+1}^{\beta \beta}
     \end{pmatrix} &= \rmJ_{\mathscr{F}}(\mSigma_\ell^{\alpha \beta}) \begin{pmatrix}\delta_{\ell}^{\alpha \beta} \\
      \delta_{\ell}^{\alpha \alpha} \\
      \delta_{\ell}^{\beta \beta}
     \end{pmatrix} + \begin{pmatrix}
     \varepsilon_{\ell + 1}^{\alpha \beta} \\
      \varepsilon_{\ell + 1}^{\alpha \alpha} \\
      \varepsilon_{\ell + 1}^{\beta \beta} 
     \end{pmatrix}.
\end{align}

\begin{figure}
  \centering
  \includegraphics [width=\linewidth]{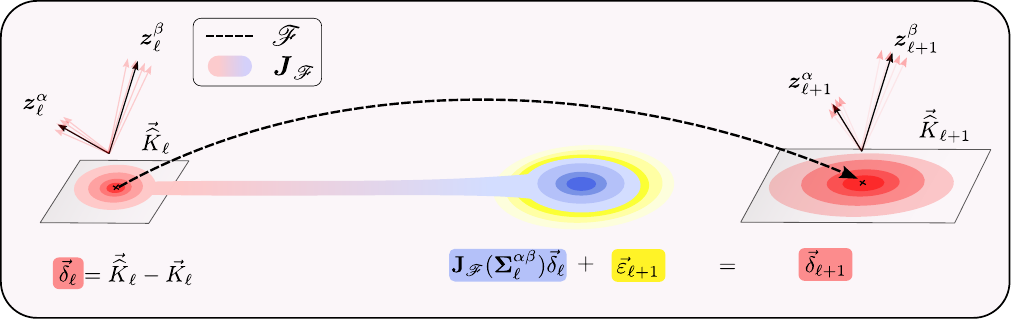}
  \caption{\textbf{Error Propagation.} Schematic illustration of the propagation of the finite-size perturbations $\vec{\delta}_\ell$ through the network. The perturbations are transported by the tangent map $\rmJ_{\mathscr{F}}$ along the mean-field covariance flow induced by $\mathscr{F}$ and driven by the layer-wise perturbations $\vec{\varepsilon}_{\ell+1}$. For clarity, the vector notation $\vec{(\,\cdot\,)}$ refers to the quantities defined in \eqref{eq:recurrence-vec-delta}.}
  \label{fig:perturbations-transport}
\end{figure}

A geometric interpretation of the evolution of $(\delta_\ell^{\alpha \beta},\delta_\ell^{\alpha \alpha},\delta_\ell^{\beta \beta})$ across layers is provided in Fig.~\ref{fig:perturbations-transport}: a given realization of the empirical kernels $(\widehat{K}_\ell^{\alpha\beta}, \widehat{K}_\ell^{\alpha\alpha}, \widehat{K}_\ell^{\beta\beta})$ are propagated from one layer to the next by the mean-field flow map $\mathscr{F}$, together with the additive random term $(\varepsilon_\ell^{\alpha\beta}, \varepsilon_\ell^{\alpha\alpha}, \varepsilon_\ell^{\beta\beta})$, as prescribed by \eqref{eq:reckernel}. In parallel, the deviations from the mean-field kernels, $(\delta_\ell^{\alpha \beta}, \delta_\ell^{\alpha \alpha}, \delta_\ell^{\beta \beta})$, are transported by the tangent (Jacobian) map $\rmJ_{\mathscr{F}}$ evaluated along the mean-field trajectory, and receive at each layer an additional direct contribution from the same random term $(\varepsilon_\ell^{\alpha\beta}, \varepsilon_\ell^{\alpha\alpha}, \varepsilon_\ell^{\beta\beta})$.

For the sake of clarity, we discuss here only the simpler recurrence relation obtained in the case $\alpha=\beta$, and suppose that the behaviour is the same for $\alpha\neq \beta$. First, the total perturbations $\delta^{\alpha \alpha}_\ell$ can be rewritten by summing the previous layer-wise fluctuations $\varepsilon^{\alpha \alpha}_{\ell'}$, transported sequentially by the tangent map $\rmJ_{\mathscr{F}}(\mSigma^{\alpha \beta}_{\ell '})$: 
\begin{equation}
\label{eq:delta_epsilon_rec}
  \delta^{\alpha \alpha}_{\ell + 1} = \sum_{\ell' = 0}^\ell \left(\prod_{\ell'' = \ell'}^\ell \chi^{\alpha}_{\ell''}\right)\varepsilon^{\alpha \alpha}_{\ell'},
\end{equation}
with $\chi^\alpha_\ell$ defined in \eqref{eq:allchi_def}. In those cases, we seek to obtain a scaling for the average and variance of the perturbation kernel, using \eqref{eq:delta_epsilon_rec}, and assuming that the variances of perturbations are uncorrelated, which leads to
\begin{equation} \label{eq:delta-variance-bias}
    \E \left[ \delta_{\ell+1}^{\alpha\alpha}\right] = \sum_{\ell' = 0}^{\ell} \left(\prod_{\ell'' = \ell'}^\ell \chi^{\alpha}_{\ell''} \right)\E \left[ \varepsilon_{\ell'}^{\alpha \alpha} \right] ,\qquad \Var[\delta_{\ell+1}^{\alpha\alpha}] = \sum_{\ell' = 0}^\ell \left(\prod_{\ell'' = \ell'}^\ell \chi^{\alpha}_{\ell''} \right)^2 \Var[\varepsilon_{\ell'}^{\alpha \alpha}]. 
\end{equation}
\noindent\textbf{Scaling of the mean and variance of $\varepsilon^{\alpha\alpha}_{\ell}$} --
We first quantify separately the fluctuations induced by the Wick defects and by the non-orthogonality of the weights.

\begin{itemize}
\item \textit{Effect of \(\varepsilon_{\text{Wick}}\)} --
Since $z^\alpha_\ell$ is small at criticality, the contribution of the Wick defects can be quantified explicitly (see Appendix~\ref{par:eps_wick}):
\begin{equation}
  \E\left[\varepsilon_{\ell,\text{Wick}}^{\alpha\alpha}\right]
  =\mathcal{O}\left(\frac{(K_{\ell}^{\alpha\alpha})^2}{N}\right),
  \qquad
  \Var\left[\varepsilon_{\ell,\text{Wick}}^{\alpha\alpha}\right]
  =\mathcal{O}\left(\frac{(K_{\ell}^{\alpha\alpha})^4}{N}\right).
  \label{eq:epsiWick_estimate}
\end{equation}
The fourth-order dependence of the variance on $K_{\ell}^{\alpha\alpha}$ needs to be remembered and is a key property of the Wick defects.

\item \textit{Effect of \(\varepsilon_{\text{Weight}}\)} --
The perturbations induced by the non-orthogonality of the weights are unbiased, with
$\E[\varepsilon_{\ell,\text{Weight}}^{\alpha\beta}]=0$
(see Appendix~\ref{par:eps_weight}). Under the same assumptions as above, one finds, to leading order in $1/N$ and $K_{\ell}^{\alpha\alpha}$,
\begin{equation}
  \Var\left[\varepsilon_{\ell,\text{Weight}}^{\alpha\alpha}\right]
  =\mathcal{O}\left(\frac{(K_{\ell}^{\alpha\alpha})^2}{N}\right).
  \label{eq:epsiWeight_estimate}
\end{equation}
As shown below, this quadratic dependence on $K_{\ell}^{\alpha\alpha}$ produces a less favourable behaviour under transport than the quartic scaling associated with the Wick defects.
\end{itemize}
\noindent\textbf{Scaling of the mean and variance of $\delta^{\alpha\alpha}_{\ell}$.}
To determine how the layer-wise fluctuations accumulate, we must evaluate the transport product appearing in \eqref{eq:delta_epsilon_rec}. For the diagonal dynamics, we show in Appendix~\ref{par:scaling_criticality} that
\[
  \prod_{\ell''=\ell'}^\ell \chi^\alpha_{\ell''}
  \sim \left(\frac{\ell'}{\ell}\right)^2.
\]
Combined with the estimates above, this yields the following scalings for the total finite-size perturbation $\delta^{\alpha\alpha}_\ell$:
\begin{itemize}
  \item \textit{For orthogonal weights,} finite-width effects arise exclusively from the Wick defects $\varepsilon_{\text{Wick}}$. Recalling that
$K_{\ell}^{\alpha\alpha}=\mathcal{O}(1/\ell)$
and using \eqref{eq:delta-variance-bias} and \eqref{eq:epsiWick_estimate}, we obtain
\begin{equation}
  \E\left[\delta_\ell^{\alpha\alpha}\right]
  =\mathcal{O}\left(\frac{1}{N\ell}\right),
  \qquad
  \Var\left[\delta_\ell^{\alpha\alpha}\right]
  =\mathcal{O}\left(\frac{1}{N\ell^2}\right).
  \label{eq:delta-orthogonal}
\end{equation}
Consequently, after normalisation by the pre-activation variance $K_{\ell}^{\alpha\alpha}$, neither the mean nor the variance of the relative perturbation depends on depth. These estimates fully characterize the finite-size corrections at first order for the orthogonal initialisation, for which $\varepsilon_{\ell,\text{Weight}}$ vanishes by construction.

\item \textit{For Gaussian weights,} finite-width effects are dominated by the non-orthogonality contribution $\varepsilon_{\text{Weight}}$. Retaining this leading contribution and using \eqref{eq:epsiWeight_estimate}, we obtain
\begin{equation}
  \E\left[\delta_\ell^{\alpha\alpha}\right]=0,
  \qquad
  \Var\left[\delta_\ell^{\alpha\alpha}\right]
  =\mathcal{O}\left(\frac{1}{N\ell}\right).
  \label{eq:delta-gaussian}
\end{equation}
After normalisation by $K_{\ell}^{\alpha\alpha}$, the variance of the relative perturbation scales as $\mathcal{O}(\ell/N)$ and therefore grows linearly with depth. This contrasts with the relative corrections induced by the Wick defects, which remain independent of $\ell$, consistently with previous studies; see, e.g., \cite{Ortho-correlation}.\\
\end{itemize}

\noindent\textbf{Implications for finite-size corrections to correlations.}
The preceding estimates can now be used to determine the magnitude of the finite-size correction to the correlation at layer $\ell$. Using equations~\eqref{eq:delta-variance-bias} and~\eqref{eq:epsiWick_estimate}, and assuming that these results extend to the case $\alpha \neq \beta$, we write
\begin{equation}
  \widehat{\rho}^{\alpha\beta}_\ell
  =
  \rho_\ell^{\alpha\beta}
  \frac{1+\delta_\ell^{\alpha\beta}/K_\ell^{\alpha\beta}}
  {\sqrt{
    \left(1+\delta_\ell^{\alpha\alpha}/K_\ell^{\alpha\alpha}\right)
    \left(1+\delta_\ell^{\beta\beta}/K_\ell^{\beta\beta}\right)
  }}.
\end{equation}
\begin{itemize}
  \item \textit{For orthogonal weights,} equation~\eqref{eq:delta-orthogonal} gives
\begin{equation}\label{eq:perturbations-orthogonal}
  \hat{\rho}_\ell^{\alpha\beta}
  =
  \rho_\ell^{\alpha\beta}
  \left(
    1+\mathcal{O}\left(\frac{1}{\sqrt{N}}\right)
  \right).
\end{equation}
Thus, the deviation from the mean-field correlation remains controlled uniformly in depth and is determined solely by the network width $N$.
\item \textit{For Gaussian weights,} equation~\eqref{eq:delta-gaussian} instead yields
\begin{equation}\label{eq:perturbations-gaussian}
  \hat{\rho}_\ell^{\alpha\beta}
  =
  \rho_\ell^{\alpha\beta}
  \left(
    1+\mathcal{O}\left(\sqrt{\frac{\ell}{N}}\right)
  \right).
\end{equation}
The factor $\sqrt{\ell/N}$ shows that finite-size corrections accumulate with depth. Consequently, deviations from the mean-field correlation may become significant in very deep networks unless the width grows sufficiently rapidly with $\ell$.
\end{itemize}
 
\subsection{Numerical experiments on manifold propagation and correlation flow \label{sub:numerical_experiment}}
\begin{figure}[ht]
  \centering
  \includegraphics[width = \linewidth]{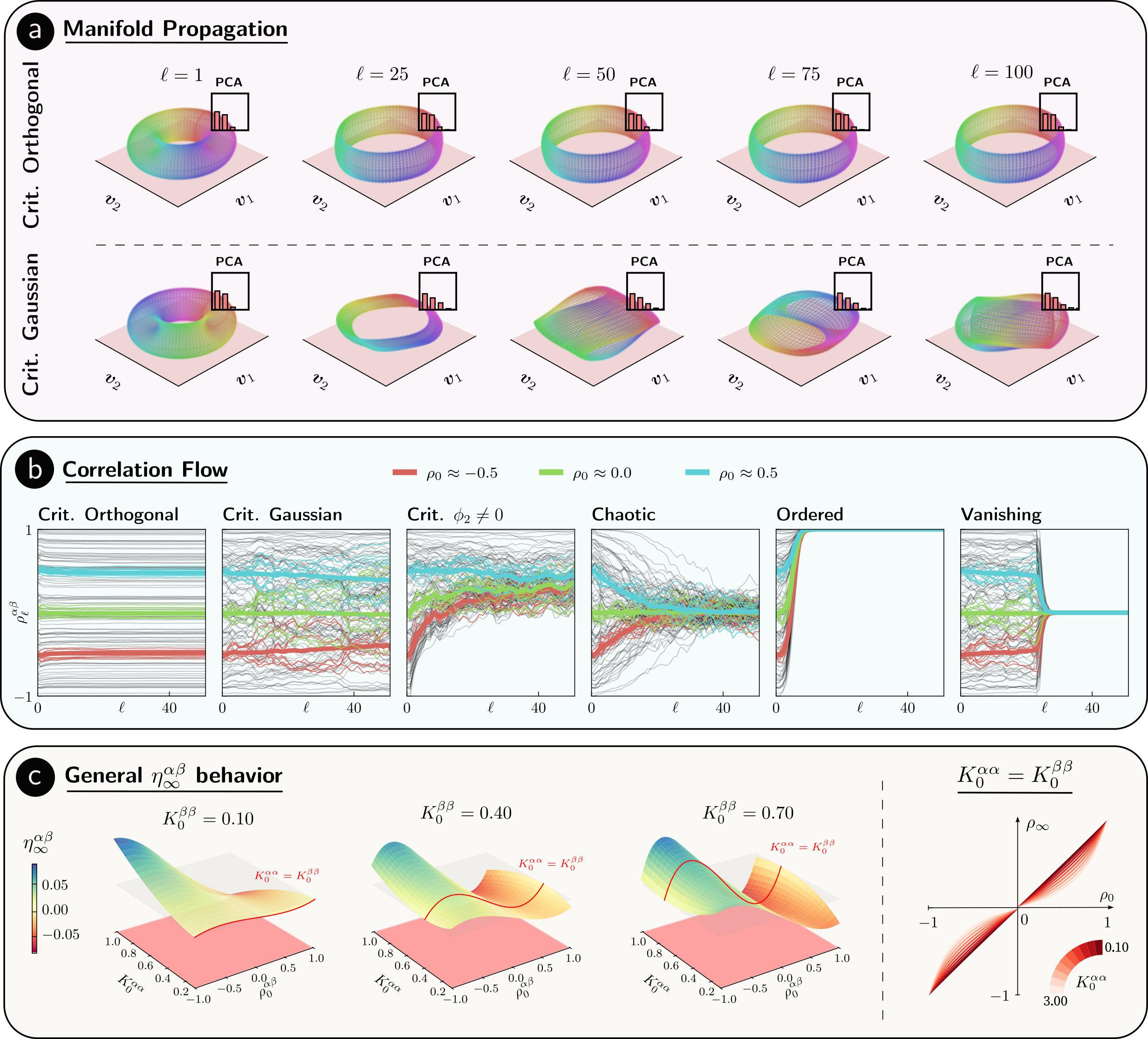}
  \caption{\textbf{Forward propagation.} \textbf{a)} Evolution of a torus manifold, \(x_\alpha \in T^2, \, \forall \alpha \in \mathcal{D}\), through a sinusoidal network. We project \(\rvz_{\ell}\) for every layer onto its first three principal components (\(\boldsymbol{v}_{1}, \boldsymbol{v}_{2}, \boldsymbol{v}_{3}\)).
    \textbf{b)} Empirical propagation of correlation with network depth, for different initialisation choices and different initial correlations. Colors associated with specific trajectories are used to ease visualization of correlation collapse or chaos. \textbf{c)}  Evaluation of \(\eta_\infty^{\alpha\beta}(\rho_{0},K^{\alpha \alpha}_{0},K^{\beta \beta}_{0} )=\rho^{\alpha\beta}_\infty-\rho_0^{\alpha\beta}\) for several input norms \(K^{\alpha \alpha}_{0}, K^{\beta \beta}_{0}\) and input correlations $\rho_0^{\alpha \beta}$ in the Mean-field limit using (\ref{eq:sine_rec_rel}).
  }
  \label{fig:propagation_correlation}
\end{figure}

In Fig.~\ref{fig:propagation_correlation}, we compare our proposed critical initialisation with other initialisation strategies and illustrate both the predicted correlation flow. Specifically, we confirm that, at criticality, the limiting correlation $\rho_\infty^{\alpha\beta}$ depends non-trivially on the input norms and initial correlations. For inputs with moderate norms, we observe that this dependence is approximately linear in the initial correlation. We also quantify the magnitude of finite-size effects across several cases for both \emph{Gaussian} and \emph{orthogonal} initialisation. The experiments are performed using a \texttt{sine} network ($\phi=\sin$) initialised at criticality, with
\begin{equation} \label{eq:params-sine-crit}
  (\sigma_w,\sigma_b,\phi_1,\phi_2,\phi_3)=(1,0,1,0,-1).
\end{equation}


\noindent \textbf{Manifold propagation.} To describe the correlation flow during forward propagation, it is instructive to first look at the propagation of a manifold through the network. (Fig.~\ref{fig:propagation_correlation}a), following for instance \cite{poole2016exponential}. 

\begin{itemize}
  \item \textit{Set-up --} We consider for that purpose a set of inputs living on a 2D torus $T^2(1, \tfrac{1}{2})$ of major radius 1 and minor radius $\tfrac{1}{2}$, embedded into a 3D ambient input space. The torus $T^2$ was specifically chosen as it illustrates the propagation of the signal generated by Random Fourier Features \cite{tancik2020}. Specifically, we select $D$ points $\{\boldsymbol{x}^\alpha\}_{\alpha = 1, \dots, D} \subset T^2$, we compute the singular value decomposition $\rmZ_\ell = \rmU_{\ell} \mSigma_{\ell} \rmV_{\ell}^\top$ of the pre-activation matrix $\rmZ_\ell = \big(\rvz^{\alpha}_{\ell}\big)_{\alpha = 1, \dots, D} \in \mathbb{R}^{D \times n_\ell}$ for both Gaussian and orthogonal initialisation strategies, and we display $\rmV_{\ell: 3}^\top \rvz^{\alpha}_{\ell} \in \mathbb{R}^3$ the projection of $\rvz^{\alpha}$ onto the first three columns of $\rmV_\ell$ (i.e., its leading three principal components), denoted $\rmV_{\ell: 3} \in \mathbb{R}^{n_\ell \times 3}$. 
This low-rank representation is essential for visualizing the network dynamics, despite the high dimensionality of the hidden layers (\(N =512\) in the numerical experiments shown in Fig.~\ref{fig:propagation_correlation}). Moreover, angle preservation is well illustrated in this truncated space if the energy in the first three modes is dominant, which should be the case if we propagate, without too much distortion, the torus $T^2$. 

\item \textit{Results -- } Clearly, the projected manifold is better preserved for orthogonal than for Gaussian initialisation, with the latter leading to increasingly distorted shapes at higher depth, due to the increased noise level in deeper layers. Note that initialisation in the ordered phase would lead to a collapse to a point, and initialisation in the chaotic phase would lead to complete destabilization of the 2D torus \citep{poole2016exponential}. We conclude that orthogonal initialisation leads to a very controlled deformation of the angle through the network, corresponding to a practical $\eta^{\alpha \beta}_\ell \sim 0$ even for large $\ell$, yielding a nearly linear behaviour of the MLP architecture.
\end{itemize}
\noindent \textbf{Correlation flow.} In Fig.~\ref{fig:propagation_correlation}b, we illustrate the propagation of correlations $\rho^{\alpha\beta}_\ell$ with depth, for various initialisation choices, in the same experiment as above. 
\begin{itemize}
  \item \textit{Set-up --} For every pair of inputs \((\boldsymbol{x}^{\alpha}, \boldsymbol{x}^{\beta}) \in T^2\), we compute the resulting correlation \(\hat{\rho}^{\alpha \beta}_\ell\) for the pre-activation at layer \(\ell\) using \eqref{eq:empirical_correlation}. For visualization purposes, we distinguish three groups of similarly correlated points, each with initial correlations around \(\rho^{\alpha \beta}_0 \in \{-0.5, 0, 0.5\}\), using distinct colours. We then plot the evolution of the correlations with network depth as thin lines, and the group averages as thick lines. 
The \textit{critical orthogonal} and \textit{critical Gaussian} cases are the same as in the previous section. To further demonstrate the degenerating behaviour for \(\phi_2 \neq 0\) predicted in Sec.~\ref{sub:corr_meanfield}, we also consider initialisation at criticality with 
\begin{equation}
  \phi \left(\cdot\right)= \sin \left(\cdot\right)+ \tfrac{1}{2} \sin^2\left(\cdot\right) \quad \xrightarrow{\text{criticality}} \quad (\sigma_w, \sigma_b, \phi_1, \phi_2, \phi_3) = (1, 0, 1, 1, -1).\label{eq:initphi2}
\end{equation}
Finally, we illustrate on the same panel the consequences of initialisation in the chaotic, ordered, and vanishing phases for the propagation of correlation, in the case of the sine activation function, with the following parameterizations:
\begin{equation}
   (\sigma_w, \sigma_b, \phi_1, \phi_2, \phi_3) =
   \begin{cases}
    {chaotic}: (\sqrt{2}, 0, 1, 0, -1), \quad \quad &(\texttt{original SIREN})\\
    {ordered}: (1 / \sqrt{3}, \frac{1}{\sqrt{N}}, 1, 0, -1),\quad &(\texttt{default PyTorch}) \\
    {vanishing}: (1 / \sqrt{3}, 0, 1, 0, -1) ,\quad &(\texttt{no bias PyTorch}).
   \end{cases}
\end{equation} \\

\item \textit{Results --} At criticality, the \textit{orthogonal}, \textit{Gaussian}, and \(\phi_2 \neq 0\) configurations each exhibit their expected behaviour: in the \textit{orthogonal} case, the propagation of correlation remains stable, as the weak level of noise -- controlled by the network width independently of network depth -- preserves the initial correlations. For the \textit{Gaussian} initialisation, the injection and accumulation of noise at each layer induces a classical random walk in the correlation; remarkably, the average trajectory of each group still closely follows the \textit{orthogonal} trajectory predicted by mean-field theory. In the \(\phi_2 \neq 0\) case, despite the remaining Gaussian noise, the correlation collapses layer after layer regardless of its initial value, in agreement with the discussion in Sec.~\ref{sub:corr_meanfield}.

Away from criticality, the \textit{chaotic} regime (close to original SIREN initialisation \cite{sitzmann2020}) leads to the expected behaviour, driving correlations toward a fixed value regardless of their initial conditions. In the \textit{ordered} phase, all correlations collapse to the trivial fixed point \(\rho_\infty^{\alpha \beta} = 1\). Finally, in the \textit{vanishing} case, the correlation flow initially mirrors the \textit{Gaussian} behaviour at criticality. However, after a few dozen layers, it collapses to zero due to limitations in machine floating-point precision. This illustrates the advantage of criticality: algebraic decay, rather than exponential collapse toward the fixed point, ensures effective propagation much deeper into the network.\\
\end{itemize}
\noindent \textbf{Functional relation between output and input correlations at criticality.} Fig.~\ref{fig:propagation_correlation}c shows the output correlation $\rho_\infty^{\alpha\beta}$ as a function of the input correlation $\rho_0^{\alpha\beta}$ in the infinite-depth limit, obtained by solving numerically the fixed-point recursion \eqref{eq:sine_rec_rel} for a sine activation.

\begin{itemize}
  \item \textit{Set-up --}
We explore different scenarios, depending on the initial values of \(K_0^{\alpha \alpha}\) and \(K_0^{\beta \beta}\). For each configuration, we plot \(\rho_{\infty}^{\alpha \beta}\) as a function of its initial value \(\rho_{0}^{\alpha \beta}\) and the initial value of \(K^{\alpha\alpha}_0\), while fixing \(K^{\beta\beta}_0\) to three distinct values. This allows us to analyze the behaviour across different input norms and initial correlations, paying particular attention to the consequences of asymmetry in the initial data. 

\item \textit{Results --}
As expected, the output correlation remains close to the input correlation when the network is initialised sufficiently close to the fixed point, namely, when the initial pre-activation variances are small. Even for order-one variances, the deviation from the identity remains limited, as indicated by the colour scale in Fig.~\ref{fig:propagation_correlation}c and observed in the torus experiment. Nevertheless, a finer analysis reveals a non-trivial structure in \(\eta^{\alpha\beta}_{\ell}\). In the equal-norm case, where \(K^{\alpha\alpha}_0=K^{\beta\beta}_0\) (Fig.~\ref{fig:propagation_correlation}c, right), perfect alignment, anti-alignment, and orthogonality are preserved across depth. For more general initial correlations, however, increasing the common input norm induces non-trivial decorrelation. In the unequal-norm case, where \(K^{\alpha\alpha}_0\neq K^{\beta\beta}_0\) (Fig.~\ref{fig:propagation_correlation}c, left), the network decorrelates the inputs across depth, even when they are initially perfectly aligned or orthogonal, and this effect strengthens as the discrepancy between their norms increases. More generally, as \(K^{\alpha\alpha}_0\) increases, \(\rho_\infty^{\alpha\beta}\) collapses toward zero over an increasingly broad range of initial correlations \(\rho_0^{\alpha\beta}\), reflecting a nonlinear loss of input information. Appendix~\ref{app:taxonomy-network} provides a detailed classification of the resulting regimes, while the practical consequences of this input-norm-dependent decorrelation are discussed in the discussion Sec.~\ref{sec:discussion}.

\end{itemize} 

\section{Algebraic behaviour of the end-to-end Jacobian and absence of dynamical isometry} \label{sec:jacobian}

Training a deep network relies on back-propagation, which requires the end-to-end Jacobian to remain stable as depth increases. At the layer-wise level, this requirement coincides exactly with covariance propagation as the mean of the layer-wise Jacobian spectrum equals the  perpendicular susceptibility $\chi^\perp$ \citep{schoenholz2017deep}.
\citep{jacobian-theory} showed that even if  the mean approaches unity, the end-to-end Jacobian, which is obtained by accumulating it over depth, can still vanish algebraically.
In addition, perpendicular susceptibility characterizes gradients only \emph{on average}. To guarantee well-behaved gradients in every direction, the stronger condition of \emph{dynamical isometry}, i.e., unit mean and vanishing variance of the full Jacobian spectrum, is needed. This was introduced by  \citep{dynamical_iso}, who considered random matrix theory to bear on this question and identified orthogonal initialisation as the key to approaching dynamical isometry. Using this random-matrix framework, we re-derive the algebraic vanishing (generalizing \citep{jacobian-theory} via an alternative method) and analyze spectral fluctuations of the end-to-end spectrum, beyond the first moment. This directly links gradient vanishing to absence of dynamical isometry and provides the groundwork methodology for the rest of this article.

Throughout this section we work in the \textit{proportional scaling limit} $n_\textrm{out} = \mathcal{O}(N)$, as introduced in Sec.~\ref{sec:architecture}: in this regime the end-to-end Jacobian retains a non-trivial eigenspectrum, with fluctuations that persist asymptotically. By contrast, the fixed output dimension limit $n_\textrm{out} = \mathcal{O}(1)$, presented in Appendix~\ref{app:fixed_nout}, yields a trivial spectrum with vanishing variance.

\subsection{On the asymptotic freeness of the Jacobians \label{sec:freeness-jacobian}}

We seek to describe the layer-wise and end-to-end Jacobian matrices, $\rmJ_\ell^\alpha$ and $\mathcal{J}_L^\alpha$ (defined in \eqref{eq:ef_jac} and \eqref{eq:ef_jacfull}), focusing specifically on the first and second moments of the eigenvalue spectrum of the associated overlap matrices $\rmJ_\ell^\alpha \rmJ_\ell^{\beta\top}$ and $\mathcal{J}_L^\alpha \mathcal{J}_L^{\beta\top}$.

A crucial tool for analyzing the end-to-end Jacobian is a large-width property known as \emph{asymptotic freeness}. This is the non-commutative analogue of statistical independence. In the limit $N \to \infty$,  it ensures that mixed spectral moments of freely independent random matrices are 
fully determined by the spectral properties of their individual factors \citep{speicher2014freeprobabilityrandommatrices}.  This property has become an important tool in random-matrix analyses of deep neural networks, particularly in pioneering studies of Jacobian spectra by \cite{dynamical_iso}.

For the Gaussian and orthogonal weight matrices considered here, asymptotic freeness of the layer-wise Jacobians $(\rmJ_\ell)_{1\leq \ell \leq L}$ has been established rigorously in the large-width limit by \cite{yang2021tensor} and \cite{Collins2023}, respectively. Since these results hold for $N \to \infty$ at fixed depth $L$, the sequential limit considered in this paper ensures that freeness applies consistently layer by layer. Our analysis will focus on matrices of the form
\begin{equation}
\rmZ_L = \left( \prod_{\ell = 1}^L \rmX_\ell \right) \left( \prod_{\ell = 1}^L \rmY_\ell \right)^\top \quad \text{with} \quad \rmX_\ell, \rmY_\ell \in \mathbb{R}^{n_\ell \times n_{\ell-1}},\label{eq:Zdef}
\end{equation}
which captures the structure of the end-to-end overlapped Jacobians.  In Appendix~\ref{app:freeness}, we provide a brief introduction to the free probability machinery, focusing on low-order moments, and use the free independence property of the matrices \( \rmX_\ell \rmY_\ell^\top \) (analogous to layer-wise Jacobians) to show that the first moment and relative rescaled variance of \( \rmZ_\ell \) can be expressed as
\begin{equation}
    m_{\boldsymbol{\rmZ}_L}^{(1)} = \prod_{\ell=1}^L m_{\rmX_\ell \rmY_\ell^\top}^{(1)} \quad \mbox{and} \quad 
    \mathbb{V}[\boldsymbol{\rmZ}_L] = \frac{n_L}{n_0} - 1 + \sum_{\ell=1}^L \frac{n_L}{n_{\ell - 1}} \mathbb{V}[\rmY_\ell^\top \rmX_\ell], \label{eq:rel-var}
\end{equation}
where \( \mathbb{V}[\boldsymbol{\rmZ}_L] \) is the relative variance defined in \eqref{eq:rel-variance-def}. These relations are the foundations of all the derivations performed in the remainder of this section.

\subsection{First moment of layer-wise and end-to-end Jacobian}

We start by examining the first moment of the end-to-end Jacobian overlap matrices. As shown by \cite{schoenholz2017deep}, the first  moment of the layerwise Jacobian is directly related to the perpendicular susceptibility of covariance:

\begin{equation}
m^{(1)}_{\rmJ_\ell^\alpha \rmJ_\ell^{\beta\top}} := \frac{1}{n_\ell} \mathbb{E}\!\left[\operatorname{Tr} \rmJ_\ell^\alpha \rmJ_\ell^{\beta\top}\right] =\chi_\ell^{\perp} , \label{eq:chi_m1}
\end{equation}
where $\chi_\ell^{\perp}$ is defined in \eqref{eq:allchi_def} (and specified in \eqref{eq:chi_terms}). This result can be generalized to the overlap  end-to-end Jacobian first moment, as a direct consequence of  \eqref{eq:rel-var}:
\begin{equation}
m^{(1)}_{\mathcal{J}_{L}^\alpha \mathcal{J}_{L}^{\beta\top}}  = \Xi^{\alpha\beta}_L \quad \text{with} \quad \Xi_{L}^{\alpha\beta} := \prod_{\ell=1}^{L} \chi_{\ell}^{\perp} \label{eq:Xi_m1}.
\end{equation}

We note that the rate of convergence of \(\chi_{\ell}^{\perp}\) to \(1\) is crucial for the convergence of the product \(\Xi_{L}^{\alpha\beta}\), as algebraic vanishing or explosion cannot be excluded at criticality. To describe this asymptotic behaviour, it is convenient to work with $\log \chi_\ell^\perp$. We show in Appendix~\ref{par:scaling_criticality} that, 
\begin{align}
    \log \chi_{\ell}^{\perp} 
    = \frac{B^{\alpha\beta}}{\ell} + \mathcal{O}\!\left(\frac{\log \ell}{\ell^2}\right)
    \quad \mbox{with} \quad
    B^{\alpha\beta} = -\frac{{\phi_3}/{\phi_1} + \rho^{\alpha\beta}_{\infty} 
    \left({\phi_2}/{\phi_1}\right)^{\!2}}{
    {\phi_3}/{\phi_1} + \tfrac{3}{4}\left({\phi_2}/{\phi_1}\right)^{\!2}}. \label{eq:defB}
\end{align}
As $L \to +\infty$, the sum $\log \Xi^{\alpha\beta}_{L} = \sum_{\ell=1}^{L} \log \chi^{\perp}_\ell$ diverges to $+\infty$ if $B^{\alpha\beta} > 0$ (gradient explosion) and to $-\infty$ if $B^{\alpha\beta} < 0$ (gradient vanishing). Recall from Sec.~\ref{sec:flow-correlation} that non-trivial correlation propagation at infinite depth requires\footnote{We note that the alternative choice $\phi_2 \neq 0$ in \eqref{eq:defB} can control the end-to-end Jacobian, at the price of precluding correlation propagation (cf. Sec.~\ref{sub:corr_meanfield}). Indeed, restricting to the diagonal case $\alpha=\beta$, for which $\rho_\infty^{\alpha\alpha}=1$, we find $\Xi^{\alpha\alpha}_\infty=\mathcal{O}(1)$ if and only if $\phi_3 = -\phi_2^2/\phi_1$.} $\phi_2=0$  (see \eqref{eq:condition-phi2}), implying $\phi_3<0$ (from \eqref{eq:condKstable_phi3}), so that $B^{\alpha\beta}=-1$, corresponding to algebraic vanishing of the end-to-end Jacobian:
\begin{align}
    \Xi_{L}^{\alpha\beta} 
    &= \exp\!\left[\sum_{\ell=1}^{L} 
      \left(-\frac{1}{\ell} + \mathcal{O}\!\left(\frac{\log \ell}{\ell^2}\right)\right)
      \right], \label{eq:algebraic-vanishing-end-to-end}\\
    &\sim \frac{c}{L}, \quad L \to \infty,\label{eq:algebraic-vanishing-end-to-end-b}
\end{align}
with $c$ a constant independent of $L$, consistent with the numerical observations of Fig.~\ref{fig:jacobian-study} and with the algebraic decay previously reported by \citep{jacobian-theory} in the case $\alpha = \beta$.

\subsection{Second moments: absence of dynamical isometry}
\label{sec:secondmoment}

Having established that the first spectral moment of the end-to-end Jacobian overlap matrix vanishes algebraically at criticality, we now turn to the second moment of \(\mathcal{J}_L^\alpha \mathcal{J}_L^{\beta \top}\), which controls the spread of the singular value distribution (in the case $\alpha = \beta$). We recall that the end-to-end Jacobian is defined in (\ref{eq:ef_jacfull}) as
 \begin{equation} \label{eq:end-to-end-jac}
 \mathcal{J}^{\alpha}_L = \prod_{\ell = 0}^{L-1}  \frac{1}{\sqrt{n_{L-\ell-1}}}\rmW_{L - \ell }\rmD^{\alpha}_{L-\ell} \quad \text{with} \quad \rmD^{\alpha}_{\ell} = \textrm{diag}\left[\phi'(\rvz_{\ell - 1}^\alpha)\right].\end{equation}
We know from \citep{Collins2023, yang2021tensor} that the layer-wise Jacobians $\rmJ_{\ell}^\alpha$ behave like freely independent matrices both in the Gaussian and the orthogonal cases, when $n_{\ell}=N \to \infty$. The matrix $\mathcal{J}_{L}^\alpha\mathcal{J}_{L}^{\beta\top}$ can also be decomposed into a product  of matrices that are freely independent, but it involves a product of rectangular matrices if some of the $n_\ell$ are different, which is the case in our set-up, when $n_\textrm{out}\neq N$. We can thus apply  in this case an extension of the variance formula \eqref{eq:rel-var} whose expression is derived in Appendix~\ref{app:freeness}, leading to 
\begin{equation} \label{eq:relative-variance-gamma}
    \mathbb{V}[\mathcal{J}^{\alpha}_L \mathcal{J}_L^{\beta \top}] = \frac{n_\textrm{out}}{n_\textrm{in}} - 1 + \sum_{\ell = 1}^L \frac{n_\textrm{out}}{n_{\ell - 1}} \left(\mathbb{V}[\tfrac{1}{n_{\ell-1}}\rmW_{\ell}^\top \rmW_{\ell}] +  \mathbb{V}[\rmD_{\ell}^{\alpha}\rmD_{\ell}^{\beta}]\right).
\end{equation}
Note that \cite{dynamical_iso} considered a constant diagonal term $\rmD^{\alpha}_{\ell}$ with depth,  assuming fast (exponential) convergence  to the fixed point. However, we previously proved that convergence is only algebraic at criticality, hence, we cannot neglect the effect of the diagonal term, which will turn out to play an important role.\\ 
    
\noindent \textbf{Gaussian initialisation.}
As detailed in the Appendix \ref{sec:appendix-gaussian-weight},  $ \mathbb{V}[\tfrac{1}{n_{\ell-1}} \rmW_\ell^\top \rmW_\ell] = {n_{\ell -1}}/{n_\ell}$. Combining this expression with (\ref{eq:relative-variance-gamma}), remembering that $n_\ell = N$ for $1\le \ell<L$, and that  \eqref{eq:rel-variance-def} implies  ${\mathbb{S}[\mathcal{J}_L^{\alpha}\mathcal{J}_L^{\beta \top}]} = {\Xi_L^{\alpha\beta}}  \sqrt{{\mathbb{V}[\mathcal{J}^{\alpha}_L \mathcal{J}_L^{\beta \top}}}]$, leads to
    \begin{align}
        \label{eq:gaussian_variance_end2end} \frac{\mathbb{S}[\mathcal{J}_L^{\alpha}\mathcal{J}_L^{\beta \top}]}{\Xi_L^{\alpha\beta}}  &=  \sqrt{\frac{n_\textrm{out}}{N}\left( L +   \sum_{\ell = 1}^L \mathbb{V}[\rmD_{\ell}^{\alpha}\rmD_{\ell}^{\beta}]\right)}.
    \end{align}
In addition, 
one can show that $\sum_{\ell = 1}^L \mathbb{V}[\rmD_{\ell}^{\alpha}\rmD_{\ell}^{\beta}]$ is a convergent sum with $L$ (see Appendix \ref{sec:appendix-sum-convergence}). Consequently, in  
the large $L$ limit, we get \begin{equation}   \frac{ \mathbb{S}[\mathcal{J}_L^{\alpha} \mathcal{J}_L^{\beta\top}] } {\Xi_L^{\alpha\beta} } \propto \frac{n_\textrm{out}}{N}  \sqrt{L} \quad   \text{and} \quad \mathbb{S}[\mathcal{J}_L^{\alpha} \mathcal{J}_L^{\beta\top}]  \propto \frac{n_\textrm{out}}{N}  \frac{1}{\sqrt{L}}. \label{eq:rel-variance-def_gaussian_scaling}\end{equation}
This is a different scaling than the one found in \cite{dynamical_iso}, where the algebraic convergence of the first moment was not taken into account. Under this scaling, dynamical isometry remains absent: in addition to the algebraic decay of the mean eigenvalue, the spectrum becomes increasingly ill-conditioned, as its spectral standard deviation decays more slowly than its mean.\\

\noindent \textbf{Orthogonal initialisation.}  In that case: 
\begin{equation}\label{eq:relative-variance-weight-orthogonal}
     \mathbb{V}\left[\frac{1}{n_{\ell-1}} \rmW_\ell^\top \rmW_\ell\right] = 0
\end{equation}
except for the last layer, where (\ref{eq:orthogonality-weights}) ensures $ \mathbb{V}[\frac{1}{n_{L-1}} \rmW_{L}^\top \rmW_{L}] = N/n_\textrm{out} - 1$. 
The same procedure as for the Gaussian case leads to \begin{equation} \label{eq:ortho_variance_end2end}
     \frac{\mathbb{S}[\mathcal{J}_L^{\alpha}\mathcal{J}_L^{\beta \top}]}{\Xi_L^{\alpha\beta}}  = \sqrt{\frac{n_\textrm{out}}{N}\left( \sum_{\ell = 1}^L \mathbb{V}[\rmD_{\ell}^{\alpha}\rmD_{\ell}^{\beta}]\right)}.
    \end{equation}
  The main difference with equation (\ref{eq:gaussian_variance_end2end}), obtained in the Gaussian case, is best revealed by considering the large depth limit, leading to:
\begin{equation}
    \frac{ \mathbb{S}[\mathcal{J}_L^{\alpha} \mathcal{J}_L^{\beta\top}]}{\Xi_L^{\alpha\beta}} \propto \frac{n_\textrm{out}}{N} \quad \text{and} \quad  \mathbb{S}[\mathcal{J}_L^{\alpha} \mathcal{J}_L^{\beta\top}] \propto \frac{n_\textrm{out}}{N} \frac{1}{L} .
   \label{eq:rel-variance-def_scaling}
\end{equation} 
  The contraction speed of the deviation is the same as the average decaying, which results in a better conditioning of the end-to-end Jacobian than for the Gaussian  initialisation.

  \subsection{Numerical confirmation of the predicted behaviour for the end-to-end gradients  \label{sub:numerical-experiments-jacob}}
    
We now present numerical confirmation of the previously derived Jacobian properties at criticality, using the same sinusoidal neural network as in  Sec.~\ref{sub:numerical_experiment}.
Most mean-field properties derived in the previous section will be observed by considering  single, finite-size network experiments, without any ensemble averaging. \\

\begin{figure}[h!]
    \includegraphics[width = \linewidth]{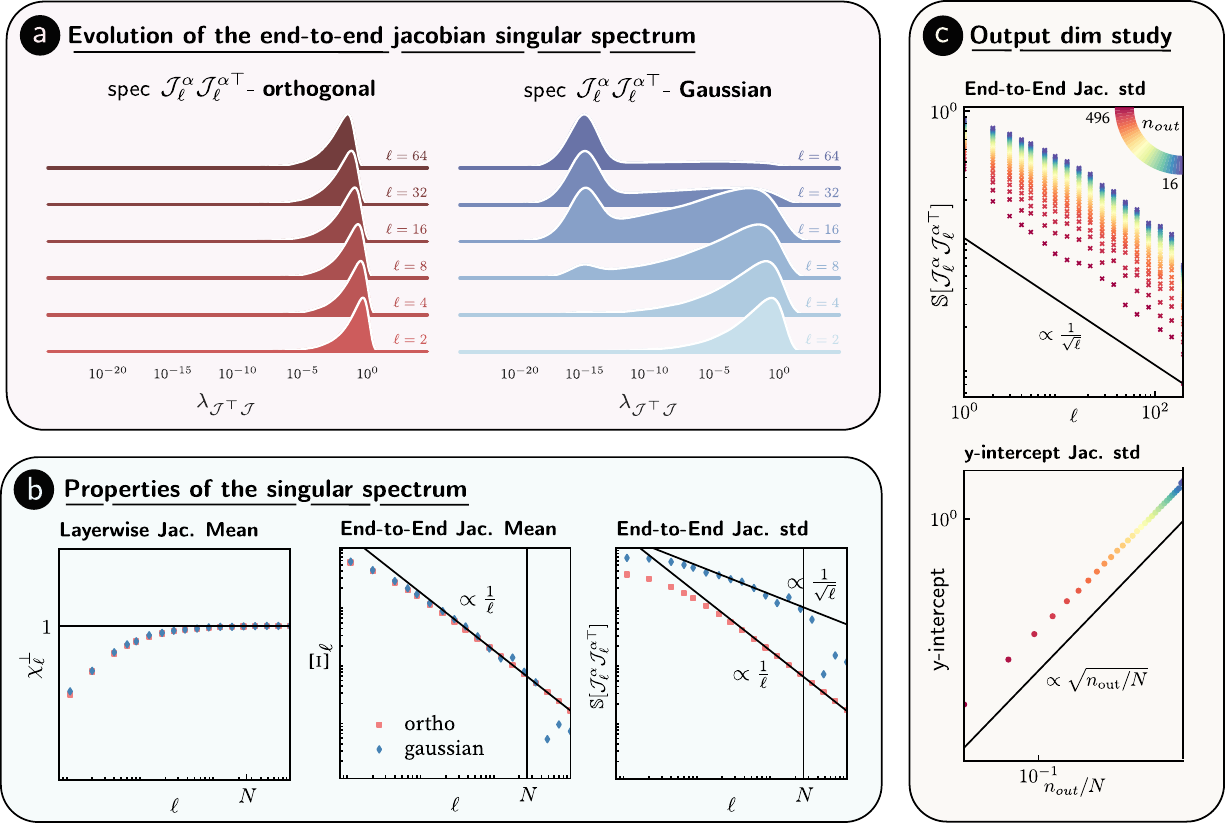}
    \caption{\textbf{Numerical analysis of the end-to-end Jacobian spectrum in sinusoidal neural networks initialised at criticality.}
\textbf{a)} Eigenvalue spectra of the end-to-end Jacobian Gram matrix $\mathcal{J}_\ell^{\alpha}\mathcal{J}_\ell^{\alpha \top}$ at different depths for a single network realization. Under \textit{orthogonal} initialisation, the spectrum decays gradually in a controlled manner, whereas under \textit{Gaussian} initialisation, it undergoes substantial broadening, with its smallest values approaching machine precision.
\textbf{b)} Spectral statistics for both initialisation schemes: the first spectral moment $\chi^\perp_\ell$ of the layer-wise Jacobian, the first spectral moment $\Xi^{\alpha\alpha}_\ell$ of the end-to-end Jacobian, and its spectral standard deviation $\mathbb{S}[\mathcal{J}_\ell^{\alpha}\mathcal{J}_\ell^{\alpha \top}]$.
\textbf{c)} Spectral standard deviation $\mathbb{S}[\mathcal{J}_\ell^{\alpha}\mathcal{J}_\ell^{\alpha \top}]$ as a function of depth for different output dimensions $n_{\mathrm{out}}$.}
    \label{fig:jacobian-study}
\end{figure}

\noindent \textbf{Evolution of the end-to-end Jacobian singular spectrum.}
We show in Fig.~\ref{fig:jacobian-study}a the singular spectrum of the end-to-end Jacobian \(\mathcal{J}_\ell^{\alpha}\mathcal{J}_\ell^{\alpha \top}\) across multiple depths. 
\begin{itemize}

\item \textit{Set-up --}
 The experiments were conducted on a neural network with constant width \( n_\textrm{in} = n_\textrm{out} = N = 512\) and total depth \(L=64\). For each selected layer \(\ell\), we computed \(\mathcal{J}_\ell^\alpha = \partial \mathbf{z}_\ell / \partial \mathbf{x} \in \mathbb{R}^{N \times N}\)  for a randomly chosen input \(\mathbf{x}^{\alpha} \in \mathbb{R}^{N}\). The singular values were then obtained by performing a singular value decomposition on the latter matrix. 

\item \textit{Results --} In the Gaussian case, the distribution of the end-to-end Jacobian spreads and approaches machine precision at greater depths, beyond which the spectrum becomes uncontrolled. In contrast, for the orthogonal initialisation, the spectrum gradually shifts toward smaller values, illustrating the expected vanishing behaviour of the first moment (see \eqref{eq:algebraic-vanishing-end-to-end-b}).
\end{itemize}

\noindent \textbf{Algebraic behaviour.} Fig.~\ref{fig:jacobian-study}b compares observed and predicted depth scaling for the layer-wise and end-to-end Jacobian spectra. 

\begin{itemize}
    \item \textit{Set-up --} We followed the same protocol as before but for a deeper network, $L = 256$ and with a smaller constant width \(n_\textrm{out} = n_\textrm{in} = N = 256\).
    \item \textit{Results --} The layer-wise Jacobian mean approaches its limiting value $\chi^\perp_\star = 1$ for both Gaussian and orthogonal weights, and the first moment of the end-to-end spectrum decays as $\Xi_\ell^{\alpha\alpha} \propto 1/\ell$, matching the algebraic prediction of \eqref{eq:algebraic-vanishing-end-to-end-b}. This confirms that the mean-field, free-probability description correctly captures spectrum propagation up to $\ell \sim N$; beyond this point, it remains accurate under orthogonal initialisation, but breaks down under Gaussian initialisation.
    
    Finally, focusing on the spectral standard deviation  $\mathbb{S}[\mathcal{J}_\ell^\alpha\mathcal{J}_\ell^{\alpha \top}]$ the primary difference between orthogonal and Gaussian initialisation becomes evident.  In the orthogonal case, a clean \(\propto 1 / \ell\) scaling emerges, while in the Gaussian case, the expected scaling \(\propto 1 / \sqrt{\ell}\) appears, in agreement with (\ref{eq:rel-variance-def_gaussian_scaling}) and (\ref{eq:rel-variance-def_scaling}). This is consistent with the broad spread of the Gaussian singular-value spectrum noted above, part of which accumulates near machine precision -- a likely source of compounding, layer-to-layer error that does not arise under orthogonal initialisation, where fluctuations remain controlled throughout the depth.

\end{itemize}

  \noindent \textbf{Output dimension effect.} Fig.~\ref{fig:jacobian-study}c examines the role of the ratio $n_\textrm{out}/N$  which, using (\ref{eq:rel-variance-def_gaussian_scaling}) and (\ref{eq:rel-variance-def_scaling}), controls the scaling of the end-to-end Jacobian variance $\mathbb{S}[\mathcal{J}_\ell^\alpha\mathcal{J}_\ell^{\alpha \top}]$: small $n_\textrm{out}/N$ acts as a concentration bottleneck on the end-to-end spectral spread. 
  
  \begin{itemize}
      \item  \textit{Set-up --} We illustrate this for Gaussian initialisation, using the same set-up as above, but now retaining only $n_\textrm{out}$ neurons out of the full $n_\ell$ at each layer, so that $\mathbf{z}_\ell \in \mathbb{R}^{n_\textrm{out}}$ and the resulting end-to-end Jacobian is $\mathcal{J}_\ell^\alpha\in \mathbb{R}^{n_\textrm{out}\times N}$. We fit the predicted power law $\mathbb{S}[\mathcal{J}_\ell^{\alpha}\mathcal{J}_\ell^{\alpha \top}] \propto 1/\sqrt{\ell}$ and track how its intercept, on a log-log plot, depends on $n_\textrm{out}/N$.

      \item  \textit{Results --} The predicted $\ell\to\infty$, $n_\textrm{out}/N$ fixed behaviour already holds for surprisingly small output widths: the $1/\sqrt{\ell}$ decay of \eqref{eq:rel-variance-def_gaussian_scaling} is already visible at $n_\textrm{out}=16$. More strikingly, the log-log intercept shifts by an amount directly proportional to $\sqrt{n_\textrm{out}/N}$, exactly as predicted.
  \end{itemize}
To conclude, these numerical experiments confirm the predicted scaling for both the vanishing of the end-to-end Jacobian and its fluctuations. This algebraic vanishing, together with the off-diagonal case $\alpha \neq \beta$ treated above, will prove essential for characterizing the Neural Tangent Kernel in Sec.~\ref{sec:ntk}.

\section{The need for orthogonal initialisation to control NTK eigenvalues}
\label{sec:ntk}
In the previous sections, we analyzed correlation flow and the singular value distribution of the end-to-end Jacobian at criticality. We now leverage these results to characterize the training dynamics of the network, which, in the infinite-width limit, are fully determined by the eigenvalue spectrum of the Neural Tangent Kernel (NTK) at initialisation 
\citep{jacot2018neural}.

\subsection{Basics on the Neural Tangent Kernel} 
\label{sec:ntk_basics}
This section focuses on a gradient descent step, i.e., $\theta_{t+1}=\theta_t - \lambda \nabla_{\theta}\mathcal L(\theta_t)$, where $\lambda$ is the learning rate. For example, a least-squares loss on the dataset $\mathcal{D} = \{(\boldsymbol{x}^\alpha, \rvy^{\alpha}) \}_{\alpha \in \{ 1, \dots , D\}}$ yields $\mathcal L(\theta_t)=\frac12\|\rvr(\theta_t)\|^2$, with lifted residual $\rvr(\theta_t)=\{\rvz_{L,t}^{\beta}-\rvy^\beta\}_{\beta\in \{1, \,\dots, \, D \}} \in \mathbb{R}^{Dn_\textrm{out}}$. 
For a sample indexed by $\alpha\in \{1, \,\dots, \, D \}$, a first-order Taylor expansion yields:
\begin{align}
  \rvz_{L,t+1}^{\alpha}
  &\simeq
  \rvz_{L,t}^{\alpha}
  -
  \lambda
  \sum_{\beta = 1}^D
  \widehat{\Theta}_{L,t}^{\alpha\beta}
  \frac{\partial \mathcal L}{\partial \rvz_{L,t}^{\beta}},
  \label{eq:finite-step-ntk-output}
\end{align}
where $\widehat{\Theta}_{L,t}^{\alpha\beta}$ denotes the empirical NTK block defined in \eqref{eq:NTK-def}. For the least-squares loss introduced above, the discrete residual dynamics are given by: 
\begin{align}
  \rvr(\theta_{t+1})
\simeq
(\mathbb I-\lambda \widehat{\Theta}_{L,t})\rvr(\theta_t).
\end{align}
The $1/\sqrt{N}$ scaling chosen in (\ref{eq:pre-post-activation}) for the weights is commonly referred to as the \emph{NTK parameterization}. As shown in \citep{jacot2018neural} (see also \citep{bahri2024houches}), this scaling implies that, in the limit $N\to+\infty$, the NTK remains constant throughout training under gradient descent. This is the \emph{frozen NTK} regime\footnote{This \emph{frozen regime} has been established only in the fixed-output-dimension limit, $n_{\textrm{out}}=\mathcal{O}(1)$, and therefore cannot be assumed to persist in the \emph{proportional scaling} regime. Determining how the NTK evolves in this regime is left for future work, as such evolution may play a critical role in \emph{feature learning}. Nevertheless, we believe that the NTK remains frozen under the depth-width scaling introduced in \eqref{eq:scaling-depth}.}. In that case, the NTK during training is fully determined by the empirical kernel at initialisation, denoted $\widehat{\Theta}^{\alpha\beta}_{L,0}$, and the Taylor expansion \eqref{eq:finite-step-ntk-output} becomes exact. The limiting kernel in the NTK regime will consistently be denoted:
\begin{equation}
   \Theta^{\alpha\beta}_{L}:=\lim_{N\rightarrow \infty} \widehat{\Theta}^{\alpha\beta}_{L,0} \quad \mbox{and} \quad \Theta_L:=\lim_{N\rightarrow \infty} \widehat{\Theta}_{L, 0}
 \end{equation}
 with 
 $$
 \widehat{\Theta}_{L, 0} = (\widehat{\Theta}^{\alpha\beta}_{L, 0})_{\alpha,\beta}.
 $$
After $t$ steps of gradient descent, we obtain
\begin{equation}
  \rvr(\theta_t)
  =
  \left(
  \mathbb I
  -
  \lambda \Theta_L
  \right)^t
  \rvr(\theta_0).
  \label{eq:finite-step-residual-dynamics}
\end{equation}
In order to make apparent the importance of the NTK spectrum, we decompose the residuals as $\rvr(\theta_t) = \sum_k c_{k,t} \boldsymbol{v}_{\Theta,k}$, with $\boldsymbol{v}_{\Theta,k}$ an eigenvector of $\Theta_L$ and $\mu_k$ its corresponding eigenvalue. This leads to the projected dynamics
\begin{equation}
  c_{k,t} = (1 - \lambda \mu_k)^t c_{k,0}. \label{eq:projection-res-evolution}
\end{equation}
Each residual component along an eigenvector $\boldsymbol{v}_{\Theta,k}$ of the NTK thus decays at a rate set by the corresponding eigenvalue $\mu_k$: large eigenvalues are learned quickly, while small eigenvalues correspond to slowly decaying residual modes. This highlights the importance of a well-controlled NTK eigenspectrum for stable training, while the eigenvectors determine the \emph{learning directions}.

In practice, the learning rate $\lambda$ must be chosen small enough to keep every mode stable, that is, $\lambda < 2/\mu_{\max}$, with $\mu_{\max}$ the largest NTK eigenvalue. With this choice, mode $k$ converges after a number of steps of order $\mu_{\max}/\mu_k$, so that the slowest-learned direction, associated with the smallest eigenvalue $\mu_{\min}$, sets the overall training time, of order $\mu_{\max}/\mu_{\min}$. A large $\mu_{\max}$ is therefore not, by itself, problematic as it can always be absorbed by rescaling $\lambda$. However, the \emph{spread} of the NTK spectrum directly controls how many steps are needed to learn every direction in function space. Controlling this spread is thus essential for efficient training.\\

\noindent \textbf{Mean-field expression of the NTK in terms of Jacobians and correlations.} Consistent with the NTK parametrization and the limit $N \to \infty$ considered here, in the next section we take the sequential limit and replace all quantities by their mean-field counterparts.\footnote{We nevertheless expect that the same framework could be readily extended to investigate finite-width effects in the neural tangent kernel.}\\

\noindent \textit{NTK expression --} In the following, we focus on the spectral properties of a given block $\Theta^{\alpha\beta}_L$, which heavily impact the conditioning of the global NTK Gram matrix $\Theta_L$. NTK blocks, given in (\ref{eq:NTK-def}), can be expressed in terms of the covariance kernels and network Jacobians defined in (\ref{eq:empirical_covariance}) and (\ref{eq:ef_jac}), respectively.
\begin{equation}
 \Theta^{\alpha\beta}_L = \frac{1}{\sigma_w^2} \sum_{\ell = 0}^{L}
  K^{\alpha\beta}_{\ell}\, \rmJ_{L \to \ell }^{\alpha}\, \rmJ_{L \to \ell }^{\beta\top}, \label{eq:ntk_compact}
\end{equation}
where we have introduced the partial end-to-end Jacobians:
\begin{equation}
\rmJ_{L \to \ell}^\alpha =
  \begin{cases}
    \prod_{k=0}^{L-\ell-1} \rmJ_{L-k}^\alpha & \text{if } \ell \leq L - 1, \\
    \mathbf{I} & \text{if } \ell = L.
  \end{cases}
\end{equation}
Expression~(\ref{eq:ntk_compact}), which represents the sum of covariances generated at layer $\ell$ and pulled back to layer $L$ by the network Jacobians, was previously derived in \cite{Spectra-NTK-CK} (equation 54 in their appendix); we re-derive it in Appendix~\ref{app:ntk} for completeness, as it plays a central role in the remainder of the paper. Motivated by the behaviour of the individual terms in the sum, we introduce the following rescaling of~\eqref{eq:ntk_compact}, which reveals the relevant structure:
\begin{equation}
  \Theta^{\alpha\beta}_L = \frac{1}{A L} \frac{1}{\sigma_w^2} \sum_{\ell = 0}^{L} \rho^{\alpha\beta}_{\ell} \, {\rmP}^{\alpha \beta}_{L\to \ell},
  \quad \text{where} \quad
  {\rmP}^{\alpha \beta}_{L\to \ell}:= A L \sqrt{ K^{\alpha \alpha}_\ell K^{\beta \beta}_\ell} \,\rmJ_{L \to \ell }^{\alpha} \, \rmJ_{L \to \ell }^{\beta\top} \,.
  \label{eq:ntk_rect}
\end{equation}
In this way, the NTK is now expressed in~\eqref{eq:ntk_rect} in terms of an activation-dependent factor $A$, defined in~\eqref{eq:def_a1}, the sequence of correlation terms $\rho^{\alpha\beta}_\ell$ across layers, and the rescaled \emph{overlap Jacobians} $\rmP^{\alpha\beta}_{L\to\ell}$. Since correlation propagation is controlled at criticality, this suggests that controlling the properties of the NTK amounts to controlling these rescaled Jacobians $\rmP^{\alpha\beta}_{L\to\ell}$, rather than the end-to-end network Jacobian. In fact, we will see that the failure to impose dynamical isometry on the end-to-end Jacobian, found in the previous section, does not prevent proper control of the NTK spectrum, since its original decay with depth can be precisely compensated for the rescaled \emph{overlap Jacobian} (see (\ref{eq:first-moment-rescaled-jacobian})). \\

\noindent \textit{Output scaling --} More precisely, in the \emph{proportional scaling} regime ($n_\textrm{out} = \mathcal{O}(N)$), Gaussian initialisation yields an end-to-end Jacobian with a non-trivial singular spectrum that widens with depth; we thus expect a similarly ill-behaved $\rmJ_{L \to \ell}$, and hence an ill-conditioned $\Theta_L^{\alpha\beta}$. Orthogonal initialisation, by contrast, prevents this widening, and we show below that it indeed yields good control of the NTK block spectrum.

In the \emph{fixed output dimension} regime ($n_\textrm{out} = \mathcal{O}(1)$), the end-to-end Jacobian spectrum is trivial, with $\mathbb{S}[\mathcal{J}^\alpha \mathcal{J}^{\beta\top}] = 0$, and both Gaussian and orthogonal initialisations are expected to control the NTK eigenvalues in the sequential limit. Orthogonal initialisation nonetheless remains valuable: at finite width, the empirical kernel exhibits fluctuations under Gaussian initialisation that are absent in the orthogonal case (Sec.~\ref{sec:finite-size-effect}), leading to markedly different dynamics in~\eqref{eq:ntk_compact} that can substantially undermine the conditioning of the block spectrum.

\subsection{Moments of NTK blocks and NTK convergence to output correlation}
Building on our analysis of correlation propagation (Sec.~\ref{sec:flow-correlation}) and Jacobian transport (Sec.~\ref{sec:jacobian}), we now compute the first two spectral moments of the block NTK using the expression derived in \eqref{eq:ntk_rect}. This analysis will reveal the close connection between the NTK and correlation propagation.\\

\noindent\textbf{First moment.} Using \eqref{eq:moments-definition} and \eqref{eq:ntk_rect}, we find: 
\begin{align} \label{eq:m1-NTK-derivation}
  m^{(1)}_{\Theta_{L}^{\alpha\beta}}
  & = \frac{1}{AL} \frac{1}{\sigma_w^2} \sum_{\ell = 0}^{L} \rho^{\alpha \beta}_\ell \frac{1}{n_\textrm{out}} \E \operatorname{Tr}{\rmP}^{\alpha \beta}_{L\to \ell} \\
  &= \frac{1}{AL} \frac{1}{\sigma_w^2} \left( \rho_\infty^{\alpha \beta} \sum_{\ell = 0}^{L} \frac{1}{n_\textrm{out}} \E\operatorname{Tr}{\rmP}^{\alpha \beta}_{L\to \ell} + \sum_{\ell = 0}^{L} \epsilon_\ell^{\alpha \beta} \frac{1}{n_\textrm{out}}\E \operatorname{Tr}{\rmP}^{\alpha \beta}_{L\to \ell} \right),
\end{align}
where we defined $\epsilon_\ell^{\alpha \beta} = \rho^{\alpha \beta}_\ell - \rho^{\alpha \beta}_\infty$. In addition, it is possible to show using (\ref{eq:chi_m1}) and (\ref{eq:product-susceptibility}) that $\frac{1}{n_\textrm{out}}\E\operatorname{Tr}{\rmP}^{\alpha \beta}_{L\to \ell} \underset{L, \ell \to \infty}{\longrightarrow} 1$. Indeed: 

\begin{align} \label{eq:first-moment-rescaled-jacobian}
  \frac{1}{n_\textrm{out}}\E \operatorname{Tr}{\rmP}^{\alpha \beta}_{L\to \ell} &= A L \sqrt{ K^{\alpha \alpha}_\ell K^{\beta \beta}_\ell} \, \frac{1}{n_\textrm{out}}\operatorname{Tr} \rmJ_{L \to \ell }^{\alpha} \, \rmJ_{L \to \ell }^{\beta\top} \\
  &\underset{L, \ell \to \infty}{\sim} A L \frac{1}{A \ell} \frac{\Xi_L^{\alpha \beta}}{\Xi_\ell^{\alpha \beta}} \qquad \mbox{with} \qquad \frac{\Xi_L^{\alpha \beta}}{\Xi_\ell^{\alpha \beta}} = \prod_{\ell' = \ell+1}^L \chi_{\ell'}^\perp \\
  &\underset{L, \ell \to \infty}{\sim} 1.
\end{align}

\noindent Furthermore, using $\lim_{\ell\to \infty}\epsilon_\ell^{\alpha \beta} = 0$ and the fact that, for every $\ell$, $\left(\E \operatorname{Tr}[\rmP_{L \to \ell}]\right)_{L \ge \ell}$ is bounded, Ces\`aro's theorem leads to:
\begin{equation} \label{eq:first-moment-block-ntk-convergence}
   m^{(1)}_{\Theta_{\infty}^{\alpha\beta}} = \frac{1}{A \sigma_w^2} \rho^{\alpha \beta}_{\infty}.
\end{equation}


\noindent This remarkable identity reveals a fundamental connection between the behavior of the NTK matrix at large depth and correlation propagation. To probe this correspondence further, we now examine the fluctuations around this mean value, before extending the analysis to the full NTK matrix.\\

\noindent\textbf{Second moment and concentration.} We now establish concentration properties around the first moment of $\Theta_{\infty}^{\alpha\beta}$, starting with its second moment:

\begin{align}
  m^{(2)}_{\Theta_{L}^{\alpha\beta}} & = \frac{1}{n_\textrm{out}} \E\Tr \left[( \Theta_{L}^{\alpha\beta})^2 \right]                              \\
  & = \frac{1}{(AL)^2} \frac{1}{\sigma_w^4} \sum_{\ell = 0}^L\sum_{\ell' = 0}^L \rho^{\alpha\beta}_\ell \rho^{\alpha\beta}_{\ell'} \frac{1}{n_\textrm{out}} \E\Tr \left[ {\rmP}^{\alpha \beta}_{L\to \ell} {\rmP}^{\alpha \beta}_{L\to \ell'} \right]. \label{eq:second-moment-ntk}
\end{align}
The computation of the trace bears strong similarities to our previous computations of the second spectral moment of the end-to-end Jacobian. Details are given in Appendix \ref{sec:second-moment-ntk-derivation}, leading to: 
\begin{equation} \label{eq:scalings}
  \textbf{Gaussian:}\quad
  \mathbb{S}\!\left[\widehat{\Theta}_{L}^{\alpha\beta}\right]
  \propto \sqrt{\frac{n_{\mathrm{out}}}{N}L},
  \qquad
  \textbf{Orthogonal:}\quad
  \mathbb{S}\!\left[\widehat{\Theta}_{L}^{\alpha\beta}\right]
  \propto \sqrt{\frac{n_{\mathrm{out}}}{NL}}.
\end{equation}
These relations provide useful prescriptions for choosing the width $N$ as a function of $n_{\mathrm{out}}$ and $L$ to ensure that the spectrum of the NTK block concentrates around
$\rho_{\infty}^{\alpha\beta}/(A\sigma_w^2)$. In the Gaussian case, maintaining spectral concentration requires $N$ to grow with depth. By contrast, orthogonal initialisation leads to increasingly strong spectral concentration as the depth increases, which is a particularly noteworthy feature that we can attribute to criticality. As shown below, in the orthogonal case, this convergence holds in a sense that is even stronger than spectral convergence.\\

\noindent\textbf{Convergence of the NTK block to the network correlation for orthogonal initialisation.}
 Using the previous results on NTK moments, we show that for the orthogonal initialisation under some specific scaling of $n_\textrm{out}$, $N$, and $L$ the NTK block converges in Frobenius norm towards $ \rho^{\alpha \beta}_{\infty}/A \sigma_w^2 \mathbf{I}$. More precisely, we show in Appendix~\ref{sec:app-eigenvectors} that for any \(\gamma \in (0,1)\) such that:
\begin{equation} \label{eq:scaling-depth}
 L \gtrsim \left(\frac{n_\textrm{out}^2}{N}\right)^{1 / (1 - \gamma)}
\end{equation}
the following bound on the Frobenius norm\footnote{It would be more natural to work with the operator norm, which could yield less restrictive and more flexible scaling conditions. However, this would require precise knowledge of the spectrum of $\rmD^{\alpha}$, which is beyond the scope of the present study.
} holds:
\begin{equation}\label{eq:bounding-G}
  \normF{ \frac{1}{L} \sum_{\ell = 1}^{L} {\rmP}^{\alpha \beta}_{L\to \ell} - \mathbf{I} }^2 \lesssim L^{-\gamma},
\end{equation}
In particular, this bound is valid for all $L$ when $n_{\textrm{out}}$ is finite. Then, it follows almost immediately from Cesàro's theorem (see Appendix~\ref{sec:app-eigenvectors}) that:
\begin{equation}
\boxed{
   \Theta^{\alpha\beta}_L \xrightarrow{L \to \infty} \frac{1}{A\sigma_w^2}\rho^{\alpha \beta}_{\infty} \mathbf{I}_{n_\textrm{out}}.
}\label{eq:mainNTKBlock}
\end{equation}
This formula establishes a simple yet profound connection between training dynamics (NTK properties) and information propagation (correlation flow at infinity). This is the major result of this paper. Note that the prefactor \(A\) is known explicitly in terms of the first and third derivatives of the activation function, as derived in equation~(\ref{eq:def_a1}). Another consequence of this formula is the known fact that each NTK block matrices $\Theta^{\alpha\beta}_L$ is diagonal in the infinite width limit \citep{bahri2024houches}, under this specific scaling. In the particular case $\alpha=\beta$, (\ref{eq:mainNTKBlock}) is equivalent to equation (9.71) in \cite{PDLT-2022}, since $\rho^{\alpha\alpha}_\infty=1$.

\subsection{Empirical confirmation of predicted NTK block properties}

\begin{figure}[ht]
  \includegraphics[width = \linewidth]{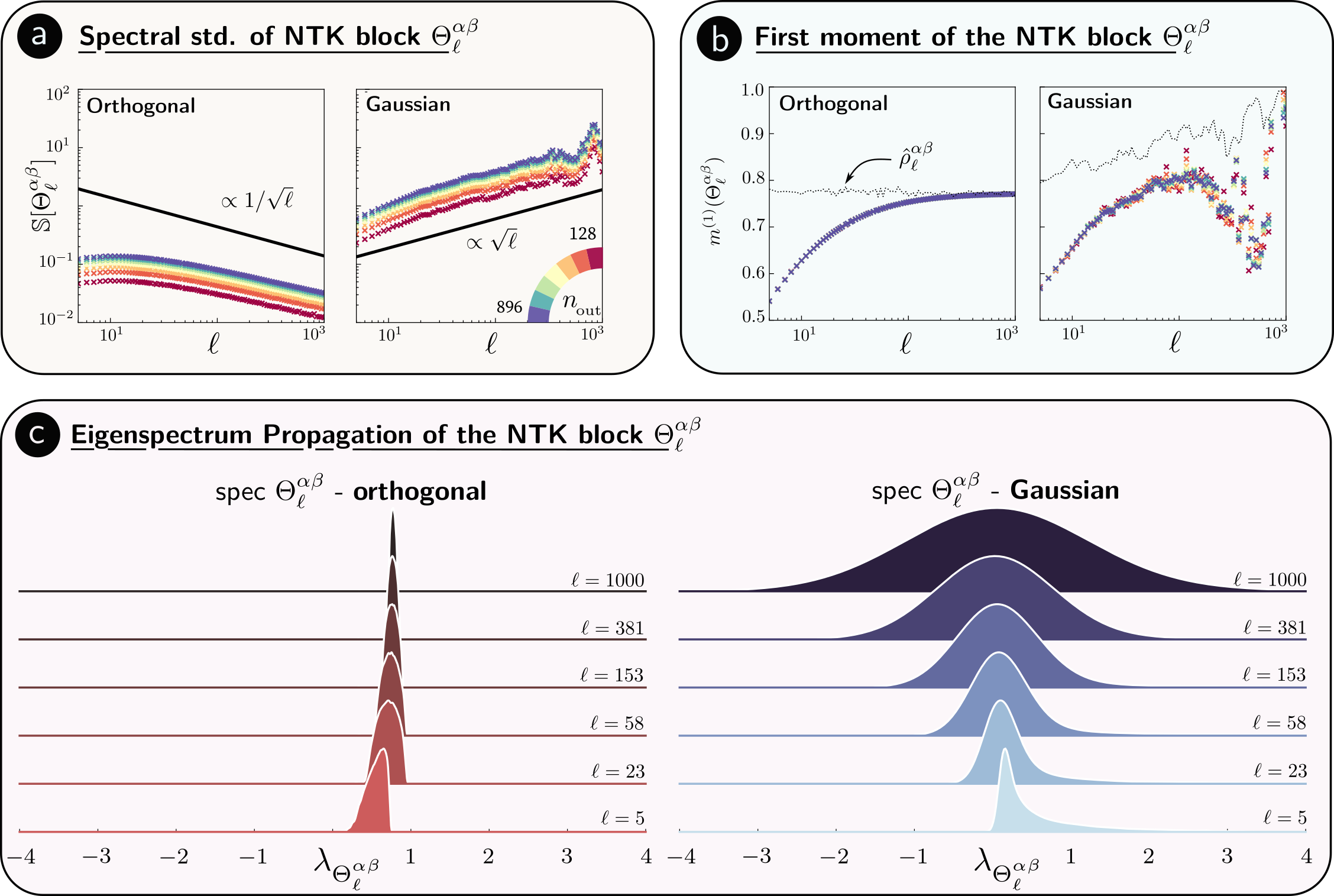}
  \caption{\textbf{Eigenspectrum propagation of the NTK block \(\Theta^{\alpha\beta}_\ell\).} \textbf{a)}
 Spectral standard deviation \(\mathbb{S}[\Theta^{\alpha\beta}_\ell]\) versus depth \(\ell\) for multiple output dimensions \(n_{\text{out}}\) under orthogonal and Gaussian initialisation, the scalings agree with \eqref{eq:scalings}. \textbf{b)}
First spectral moment \(m^{(1)}(\Theta^{\alpha\beta}_\ell)\) versus depth \(\ell\), compared to the empirical correlation \(\hat{\rho}^{\alpha\beta}_\ell\) for multiple output dimensions \(n_{\text{out}}\).
\textbf{c)} Propagation of the ambient NTK block spectrum \(\lambda_{\Theta^{\alpha\beta}_\ell}\) for a fully connected sinusoidal network of width \(N = 1024\) and depth \(L = 1024\) under orthogonal and Gaussian initialisation.
}
  \label{fig:NTK-block}
  \end{figure}

Fig.~\ref{fig:NTK-block} illustrates empirical observations of NTK block spectral properties with either Gaussian or orthogonal initialisation, together with the convergence of the NTK to the network output correlation. \\

\noindent \textbf{Spectral standard deviation.} Fig.~\ref{fig:NTK-block}a shows the empirical spectral standard deviation $\mathbb{S}[\widehat{\Theta}_{L}^{\alpha\beta}]$ for several output dimensions $n_{\mathrm{out}}$ and both initialisation strategies:
\begin{itemize}
  \item \textit{Set-up --} The experiments are conducted on the same fully connected sinusoidal network as in Sec.~\ref{sub:numerical-experiments-jacob}, with $(N,L)=(1024,1024)$, and $n_{\mathrm{out}}$ varies between $1$ and $1024$. 
  
  To preserve the critical bias-free initialisation throughout training, we remove the bias parameters entirely from the network rather than merely initializing them to zero. This avoids taking into account the contribution from their (non-zero) gradients in the NTK. Two inputs, $\boldsymbol{x}^\alpha$ and $\boldsymbol{x}^\beta$, are independently sampled from $\mathcal{N}(0,\mathbf{I}_{n_{\mathrm{in}}})$ and kept fixed throughout the experiment. 
  
  To study the propagation of the NTK through the hidden layers, we compute layer-wise NTK blocks rather than restricting attention to the final output layer. We then define for each selected layer $\ell$ an \emph{intermediate} output variable $\widetilde{\rvz}_\ell \in \mathbb{R}^{n_{\mathrm{out}}}$ by selecting $n_{\mathrm{out}}$ coordinates of $\rvz_\ell$. The corresponding layer-wise NTK block of size $\widetilde{\Theta}_\ell^{\alpha\beta} \in \mathbb{R}^{n_{\mathrm{out}} \times n_{\mathrm{out}}}$ is obtained by replacing $\rvz_\ell$ by $\widetilde{\rvz}_\ell$ in equation \ref{eq:NTK-def}. 
  
  \item \textit{Results --} 
For both Gaussian and orthogonal initialisations, the predicted increase in
$\mathbb{S}[\widetilde{\Theta}_{L}^{\alpha\beta}]$ with $n_{\mathrm{out}}$
is observed, consistent with the earlier Jacobian experiments shown in Fig.~\ref{fig:jacobian-study}. Regarding the dependence on depth, we obtain the following results:

For the \emph{Gaussian} case, the scaling derived in
\eqref{eq:scalings} agrees remarkably well with the empirical results
across all output dimensions. Finite-width effects become visible when $\ell / N \approx 1$, which closely parallels our analysis on finite-size fluctuations.

For the \emph{orthogonal} case, the spectral standard deviation decreases
as $\ell^{-1/2}$ with depth, as predicted by \eqref{eq:scalings}, and this holds for the entire range of output dimensions. 

\end{itemize}

\noindent \textbf{First moment of the NTK block.} The observed evolution of the first spectral moment of the NTK block with depth is displayed in Fig.~\ref{fig:NTK-block}b, and compared with the evolution of the empirical correlation $\widehat{\rho}^{\alpha \beta}_{\ell}$ computed at each layer. 

\begin{itemize}
  \item \textit{Set-up --} We follow the same protocol as before for computing the layer-wise NTK block $\widetilde{\Theta}^{\alpha \beta}_\ell$. We compute the empirical correlation using \eqref{eq:empirical_correlation}, and the full hidden pre-activation $\rvz_\ell$.
  \item \textit{Results --} In the orthogonal case, $\widehat{\rho}^{\alpha \beta}_{\ell}$ converges simultaneously with the first moment (given that $A \sigma_w^2=1$ for sine), matching the prediction \eqref{eq:first-moment-block-ntk-convergence}, and the fluctuations are small regardless of the output size and of network depth, consistent with the discussion in Sec.~\ref{sec:finite-size-effect}. As expected, Gaussian initialisation leads to diverging finite-size fluctuations that disrupt the convergence of the first moment towards correlation as network depth increases.
\end{itemize}

\noindent \textbf{Evolution of the NTK block spectrum.} 
As a direct visual illustration, Fig.~\ref{fig:NTK-block}c shows the evolution of the layer-wise NTK block spectrum with depth.
\begin{itemize}
  \item \textit{Set-up --} We used the same set-up as before, with the output dimension fixed at $n_{\mathrm{out}}=N$.
  
  \item \textit{Results --} In agreement with our theoretical analysis, the spectrum becomes increasingly ill-conditioned with depth under Gaussian initialisation. In contrast, orthogonal initialisation yields a stable spectrum that concentrates without the spurious fluctuations observed in the Gaussian case.
\end{itemize}

\subsection{Global NTK conditioning and dataset alignment \label{sub-sec:NTK-dataset-alignment}}
Now that we have discussed the conditioning of the NTK block components $\Theta^{\alpha\beta}_L$ we will derive in this section results about the first moments of the full NTK matrix $\Theta_L$ and discuss the alignment between the NTK and correlation eigenvectors.\\

\noindent\textbf{Moments of the global NTK.} As a first step, we show that the first two moments of the full and the block NTK are related through: 
  \begin{align}
    m^{(1)}_{ \Theta_{L}} &:= \frac{1}{n_\textrm{out}D} \Tr \E \left[\Theta_{L} \right] = \frac{1}{D} \sum_{\alpha \in \{1,\ldots, D\}} m^{(1)}_{ { \Theta}^{\alpha\alpha}_{L}},\\
    m^{(2)}_{ \Theta_{L}} &:= \frac{1}{n_\textrm{out}D} \Tr \E \left[(\Theta_{L})^2\right]  = \frac{1}{D}\sum_{\alpha, \beta \in \{1,\ldots, D\}} \frac{1}{n_\textrm{out}} \Tr \E \left[ { \Theta}_{L}^{\alpha\beta} { \Theta}_{L}^{\alpha \beta \top} \right] .\label{eq:conditionGammaTheta}
  \end{align}
  
Then using Cauchy-Schwarz inequality, useful bounds are also obtained (see Appendix~\ref{sec:app-theta2gamma}):
  \begin{align} \label{eq:cs-ntk}
  m^{(2)}_{{ \Theta}_{L}^{\alpha\beta} }\leq
    \frac{1}{n_\textrm{out}} \E \Tr \left[ { \Theta}_{L}^{\alpha\beta} { \Theta}_{L}^{\alpha \beta \top} \right] &\leq \sqrt{m^{(2)}_{{ \Theta}_{L}^{\alpha\alpha} }}\sqrt{m^{(2)}_{{ \Theta}_{L}^{\beta\beta} }}.
  \end{align}
When applied to \eqref{eq:conditionGammaTheta}, these bounds lead to two important inequalities, depending on the initialisation choice:
  \begin{align}
   \textbf{Gaussian:}\quad  m^{(2)}_{ \Theta_{L}} &\geq \frac{1}{D} \sum_{\alpha, \beta \in \{1,\ldots, D\}} m^{(2)}_{{ \Theta}_{L}^{\alpha \beta}} \propto \frac{n_\textrm{out}}{N} L ,\\
  \textbf{Orthogonal:}\quad 
     m^{(2)}_{ \Theta_{L}} &\leq \frac{1}{D} \sum_{\alpha, \beta \in \{1,\ldots, D\}} \sqrt{m^{(2)}_{{ \Theta}_{L}^{\alpha\alpha} }}\sqrt{m^{(2)}_{{ \Theta}_{L}^{\beta\beta} }}\propto \frac{n_\textrm{out}}{LN} .
  \end{align}
  The scalings are obtained using equation (\ref{eq:scalings}). This demonstrates the benefits of using an orthogonal initialisation for a controlled conditioning of the NTK in the large depth limit.\\

\noindent \textbf{Learning directions in the sequential limit.} Now that we have characterized the eigenvalues of the global NTK, we turn to its eigenmodes, which define the directions of learning during training, as discussed in Sec.~\ref{sec:ntk_basics}. Using (\ref{eq:mainNTKBlock}), the large depth limit of the mean-field global NTK matrix is expressed as 
\begin{equation}
  \Theta_L \xrightarrow{L \to \infty} \frac{1}{A\sigma_w^2} \boldsymbol{\rho}_{\infty} \otimes \mathbf{I}_{n_{\mathrm{out}}} \quad \text{with} \quad (\boldsymbol{\rho}_{\infty})_{\alpha \beta} \in \mathbb{R}^{D \times D}:= \rho^{\alpha \beta}_{\infty}. \label{eq:final}
\end{equation}
Denoting $\rmV_{\boldsymbol{\rho}} = (\boldsymbol{v}_{\boldsymbol{\rho},k})_{1\leq k \leq D}\in \mathbb{R}^{D \times D}$ the eigenvectors of $\boldsymbol{\rho}_{\infty}$ and $\rmV_{\Theta} \in \mathbb{R}^{n_{\mathrm{out}}D \times n_{\mathrm{out}}D} $ the eigenvectors of $\Theta_\infty$, we obtain: 
\begin{equation}
  \rmV_{\Theta} = \rmV_{\boldsymbol{\rho}} \otimes \mathbf{I}_{n_{\mathrm{out}}}.
\end{equation}

In other words, each eigenvector of $\boldsymbol{\rho}_{\infty}$ generates an $n_{\mathrm{out}}$-dimensional subspace associated with the same eigenvalue. Within each such subspace, there is no preferred direction: the residual components are learned isotropically. Moreover, they are learned at the same rate. Indeed, consider the matrix representation of the initial residual, $\rvr(\theta_0) \in \mathbb{R}^{D \times n_{\mathrm{out}}}$. The projection coefficients onto the subspace generated by $\boldsymbol{v}_{\boldsymbol{\rho},k}$, the $k$-th column of $\rmV_{\rho}$, are collected in the vector
\begin{equation}
  \rvc_k^0
  =
  \rvr(\theta_0)^{\top}
  \boldsymbol{v}_{\boldsymbol{\rho},k}
  \in \mathbb{R}^{n_{\mathrm{out}}}.
\end{equation}
Since the directions corresponding to all components of $\rvc_k^0$ share the eigenvalue $\mu_k$, these components evolve according to the same dynamics, given by \eqref{eq:projection-res-evolution}.\\

\noindent \textbf{Learning in the linearized regime.}
We observe in Sec.~\ref{sub:numerical_experiment} that when the input data are of order one (typically not too large), initialisation at criticality yields $\rho^{\alpha \beta}_\infty \approx \rho^{\alpha \beta}_0$, with small corrections $\eta^{\alpha\beta}_\infty(\rho_0^{\alpha\beta})$. In this regime, each eigenvector $\boldsymbol{v}_{\boldsymbol{\rho},k}$ of $\boldsymbol{\rho}_\infty$ approximately aligns with an eigenvector of $\boldsymbol{\rho}_0$, which describes a principal direction of variation in the dataset. To visualise such effect it is convenient to express the full correlation matrix $\boldsymbol{\rho}_0$ as
\begin{align}
  \boldsymbol{\rho}_0
  &=
  \rmD^{-1/2}\rmX\rmX^\top\rmD^{-1/2} \quad \text{with} \quad  \rmD
  =
  \operatorname{diag}(\rmX\rmX^\top),
  \\
  &=
  \rmV_{\boldsymbol{\rho}_0}
  \boldsymbol{\Sigma}^2
  \rmV_{\boldsymbol{\rho}_0}^\top,
\end{align}
where $\operatorname{diag}$ denotes the diagonal matrix formed from
the diagonal entries of its argument. Here,
$\rmX\in\mathbb{R}^{D\times n_{\mathrm{in}}}$ is the sample-wise
data matrix. We denote by $\boldsymbol{\Sigma}$ the matrix of singular
values of the normalised data matrix $\rmD^{-1/2}\rmX$, and by
$\rmV_{\boldsymbol{\rho}_0}$ the matrix of its left singular vectors.
These vectors are also the eigenvectors of $\boldsymbol{\rho}_0$ and
therefore describe directions of variation across samples, rather than
directions in the feature space $\mathbb{R}^{n_{\mathrm{in}}}$. Consequently, in this regime, the network learns to reduce the residual isotropically across all subspaces aligned with the principal modes of the dataset. \\

\noindent \textbf{Spectral bias from a different perspective.}
The previous results allow us to reinterpret spectral bias. Indeed, we have shown that it is not merely a low-frequency bias, but rather a more general phenomenon that biases learning toward the principal modes of the dataset. Existing explanations typically attribute this phenomenon to kernel-specific structure (e.g., Fourier or spherical harmonic alignment) \citep{freq-bias,freq-bias2,tancik2020}. Here, we identify a more general mechanism. In the large-depth limit, in the linearized regime where $\boldsymbol{\rho}_\infty \approx \boldsymbol{\rho}_0$, the NTK eigenvectors align with the left singular vectors of $\rmD^{-1/2}\rmX$ itself.

To illustrate this, consider a toy dataset with one dimensional inputs: the output of a simplified random Fourier-features layer (see Fig.~\ref{fig:dataset}), which has been highlighted in several works \cite{RF-Rahimi, tancik2020, Wang2021}. We consider a special case where the weights are non-centered:
\begin{equation} \label{eq:fourier-features-enbedding}
  \rmX:= \frac{1}{\sqrt{n_{\textrm{in}}}} \sin \left(\rmW_0 \widetilde{\rmX} + \rvb \right)^\top, \quad \rmW_0 \sim \mathcal{N}(\mu, \sigma_f^2 \mathbf{I}), \quad \rvb \sim \mathcal{U}[0, 2 \pi].
\end{equation}
With $\rmW_0 \in \mathbb{R}^{n_{\mathrm{in}} \times 1}$, $\widetilde{\rmX} \in \mathbb{R}^{1 \times D}$, $\rmX \in \mathbb{R}^{ D\times n_{\textrm{in}}}$. As $n_{\textrm{in}} \to \infty$, the correlation between two rows of $\rmX$, $\boldsymbol{x}^\alpha $ and $\boldsymbol{x}^\beta$ converges to the modulated Gaussian kernel \citep{RF-Rahimi}:
\begin{equation}
  \rho^{\alpha \beta}_0 = \exp \left( - \frac{\sigma_f^2}{2} \| \tilde{x}^{\alpha} - \tilde{x}^{\beta} \|^2 \right) \cos\left(\mu  ( \tilde{x}^{\alpha} - \tilde{x}^{\beta}) \right),
\end{equation}
which is diagonalized by the Fourier basis. Hence, the left singular vectors of $\rmX$ align with Fourier modes, and the eigenvalues decay with frequency:
\begin{equation}
  \lambda(\omega) \propto \exp \left( -\frac{(\omega - \mu)^2}{2 \sigma_f^2} \right) + \exp \left( -\frac{(\omega + \mu)^2}{2 \sigma_f^2} \right).
\end{equation}

The case $\mu=0$ corresponds to the usual spectral bias toward low frequencies, as highlighted in many studies. However, the case $\mu \neq 0$ exhibits different behaviour: frequencies close to $\omega=\mu$ are learned before low frequencies. The preferred alignment with low frequencies is therefore a specificity of the dataset/features we use. Hence, designing $\rho_0$ to encode different harmonic content allows the network to be steered toward arbitrary directions---not just low frequencies.

\subsection{Numerical illustration of the global NTK-correlation matrix relation \label{sec:NTK- numerical-experiments}}

This section empirically demonstrates the predicted relation (\ref{eq:final}) between the global NTK $\Theta_L$ and the correlation kernel $\boldsymbol{\rho}_L$ in the large-$L$ limit. To verify this, we use three complementary approaches: (i) examining the NTK eigenspectrum propagation with depth; (ii) comparing NTK matrix entries with those of the empirical correlation matrix; and (iii) studying the spectral properties of the correlation and NTK matrices.\\

\begin{figure}
  \includegraphics[width = \linewidth]{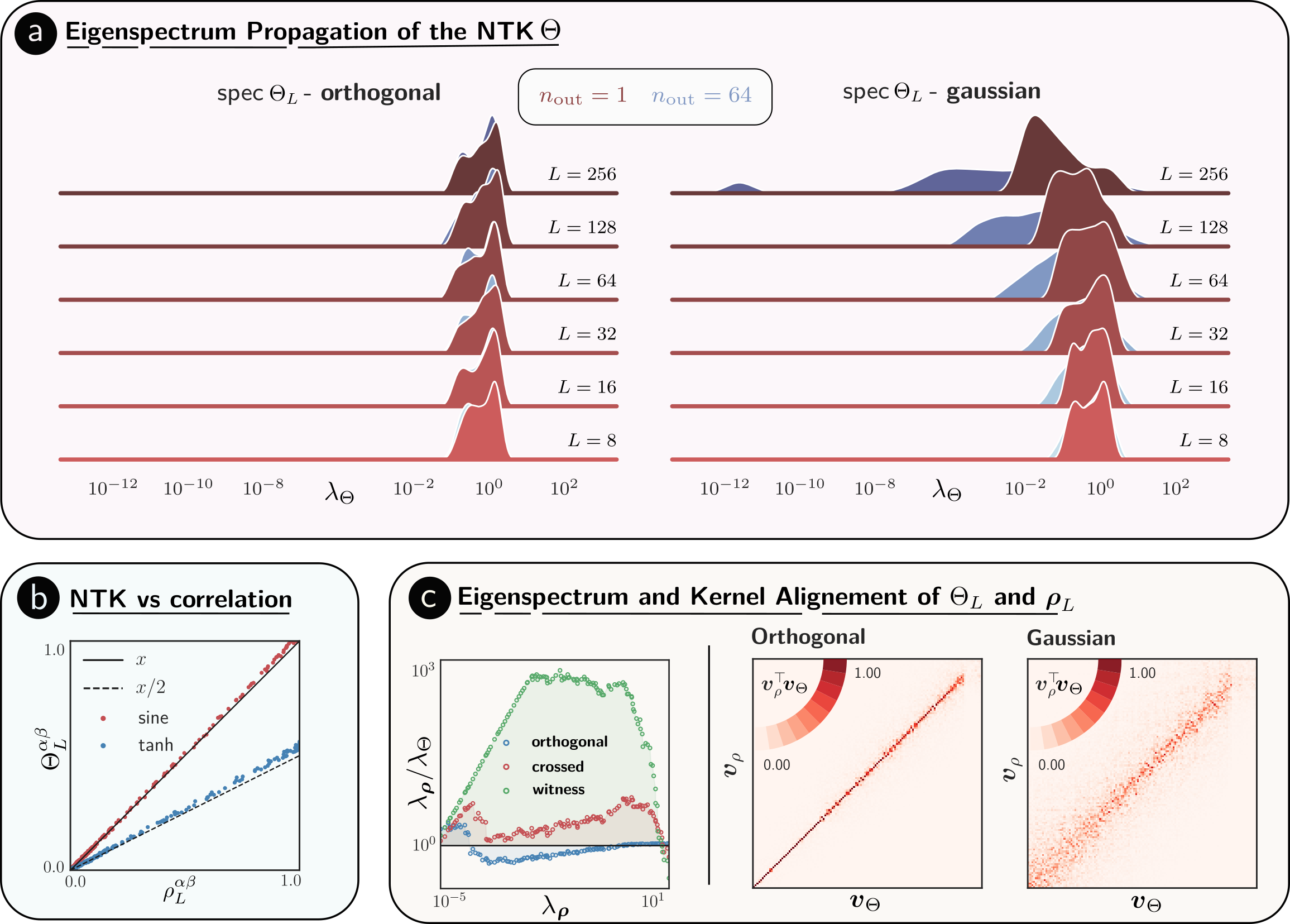}
  \caption{\textbf{NTK and correlation comparison.} \textbf{a) }
Propagation of the NTK eigenspectrum for sinusoidal neural networks for orthogonal initialisation (on the left) and Gaussian initialisation (on the right). We consider both the fixed output (\(n_{\mathrm{out}} = 1\)) and large output scaling (\(n_{\mathrm{out}} = 64\)). \textbf{b) } Scatter plot of the NTK matrix entries \(\Theta_L^{\alpha \beta}\) against the correlation matrix entries \(\rho_L^{\alpha\beta}\) for orthogonal initialised networks with activation functions \texttt{sine} and \texttt{tanh}. The theoretical scalings with prefactor \(1/(A \sigma_w^2)\) are plotted as dashed black lines for both activation functions. \textbf{c) } On the left, we plot the overlay of the eigenspectra of the NTK and correlation matrices for orthogonal and Gaussian initialisations. On the right, we study the alignment of the NTK and correlation eigenvectors for orthogonal and Gaussian initialisation.}
  \label{fig:NTK-full}
\end{figure}

\noindent \textbf{Eigenspectrum propagation of the NTK.} In Fig.~\ref{fig:NTK-full}a, we analyze the evolution of the full NTK spectrum with depth. 

\begin{itemize}
  \item \textit{Set-up --} We consider the same sinusoidal network as before, and a structured random dataset generated via a Fourier-feature embedding, as defined in \eqref{eq:fourier-features-enbedding}. The Fourier-feature frequency is $\sigma_f=8$, and the dataset comprises $D = 256$ normally distributed points in $\mathbb{R}$, with $d = 1$. Here we will analyze the NTK spectrum at depths ranging from $L=8$ to $L=256$, with a fixed width $N=256$. Output dimensions are set to:
\[
  \textbf{fixed-output:} \quad n_{\mathrm{out}} = 1 \quad \text{and} \quad \textbf{proportional scaling:} \quad n_{\mathrm{out}} = 64.
\]
  \item \textit{Results --} For orthogonal initialisation, the NTK spectrum remains stable across depth, with minimal dependence on $n_{\mathrm{out}}$; fluctuations predicted by \eqref{eq:scalings} are small and do not affect the bulk. In contrast, Gaussian initialisation introduces strong spectral fluctuations, visible even for $n_{\mathrm{out}}=1$ and amplified at $n_{\mathrm{out}}=64$. This aligns with finite-size fluctuations of order $\mathcal{O}(L/N)$ perturbing the NTK propagation. For $n_{\mathrm{out}}=64$, the Jacobian transport in \eqref{eq:ntk_rect} is ill-conditioned even at the mean-field level\footnote{We expect finite-size fluctuations to affect Jacobian transport at the same order of magnitude as the mean-field perturbations identified above, since both arise from the same mechanism: the \emph{non-orthogonality} of the weight matrices. A detailed analysis is left for future work.}, causing spectral fluctuations to grow with depth and further distort the eigenspectrum.
\end{itemize}

\noindent \textbf{Correlation and NTK.} In Fig.~\ref{fig:NTK-full}b, we compare the NTK and correlation matrices at the entry level. The goal of this experiment is to test the linear relation between $\Theta_L^{\alpha\beta}$ and $\rho_L^{\alpha\beta}$ predicted in \eqref{eq:final}. Specifically, we seek to check the dependence of this relation on the prefactor $1/A$ which depends on the activation function, comparing \texttt{sine} and \texttt{tanh} networks.

\begin{itemize}
  \item \textit{Set-up --} For the \texttt{sine} network, we use the same parameters as in previous experiments. For the \texttt{tanh} network, initialisation at criticality is
\begin{equation}
  \label{eq:params-tanh-crit}
  (\sigma_w, \sigma_b, \phi_1, \phi_2, \phi_3) = (1, 0, 1, 0, -2).
\end{equation}
The corresponding prefactors are $ A = 1$ for \texttt{sine} and $ A = 2$ for \texttt{tanh}. We restrict this analysis to orthogonally initialised networks to avoid fluctuation effects. 
The dataset is constructed using the same Fourier-feature embedding as before, but with $32$ input points for visualization, and network properties are set to $(L,N,n_\textrm{out})=(128,256,1)$. We then plot the entries $\Theta_L^{\alpha \beta}$ against the corresponding entries $\widehat{\rho}_{L}^{\alpha\beta}$.
\end{itemize}

\noindent \textbf{Eigenspectrum and kernel alignment.} Fig.~\ref{fig:NTK-full}c extends the comparison of the NTK and correlation matrices to their spectral properties. 

\begin{itemize}
   \item \textit{Set-up --} In this experiment, we use the same sinusoidal networks and Fourier-feature inputs as in Fig.~\ref{fig:NTK-full}b, with $D = 256$ input points to better resolve the spectrum. We then plot the ratio of the ordered eigenvalues of the NTK $\Theta_L$ and of the last correlation matrix $\widehat{\boldsymbol{\rho}}_{L}$ against the corresponding correlation eigenvalues. We also assess the alignment of the corresponding eigenvectors. For orthogonal initialisation, we compare the NTK and correlation matrices directly; for Gaussian initialisation, in addition to the Gaussian-correlation / Gaussian-NTK (\textsf{\textbf{witness}} label in the legend) we also compare the Gaussian NTK against the correlation matrix of an orthogonally initialised network (\textsf{\textbf{crossed}} label in the legend), as a mean-field reference.
   \item \textit{Results --} Under orthogonal initialisation, the NTK and correlation spectra agree almost perfectly, with the eigenvalue ratios close to $1$ across the entire spectrum and near-perfect eigenvector alignment -- confirming that orthogonal initialisation at criticality enables learning at similar rates along the directions set by the (near input-like) correlation matrix at the network's end.
   
    The Gaussian case is more subtle. Surprisingly, the Gaussian NTK spectrum agrees closely with the \emph{orthogonal} correlation spectrum, but much less with the Gaussian correlation spectrum itself. In the latter case, both the eigenvalues and eigenvectors are more strongly affected by finite-size fluctuations than those of the NTK. We attribute this effect to a possible self-averaging mechanism in the NTK recursion \eqref{eq:ntk_rect}, which appears to buffer the NTK against fluctuations that more directly affect the correlation kernel (eigenvector comparison not shown).
\end{itemize}

These spectral results corroborate the entrywise comparison of Fig.~\ref{fig:NTK-full}b: orthogonal initialisation yields a clean, close-to-exact alignment between the NTK and correlation kernels, while Gaussian initialisation introduces finite-size fluctuations that mostly blur the correlation spectrum.

\section{Discussion \label{sec:discussion}}

\noindent \textbf{Summary.} Let us unpack what the results above imply in practice, taking them in reverse order of their presentation. First, we established that the NTK governing learning dynamics is, throughout the critical regime studied here and at large depth, exactly proportional to the large-depth correlation \(\rho_\infty^{\alpha\beta}\), with a proportionality constant fixed solely by the activation function (see \eqref{eq:mainNTKBlock}). Remarkably, this holds even though the end-to-end Jacobian itself vanishes algebraically and dynamical isometry fails, which turns out not to be an obstruction to predictable and controlled learning. Second, turning to finite-width corrections, we found that orthogonal initialisation suppresses the leading fluctuations that otherwise accumulate with depth under Gaussian initialisation (see \eqref{eq:perturbations-orthogonal} and \eqref{eq:perturbations-gaussian}). Third, we showed that, at criticality, the limiting correlation is itself an explicit function of the initial correlation \(\rho_0^{\alpha\beta}\) under the condition \(\phi_2 = 0\) (see \eqref{eq:condition-phi2}) -- depending strongly on input normalisation.\\

\noindent \textbf{Key takeaways.} Our results show that criticality links information propagation to learning through the asymptotic correlation--NTK relation, while orthogonal initialization stabilizes this structure against finite-width fluctuations. Together, these results provide a principled basis for using orthogonal initialization at criticality in very wide and deep networks.\\

\noindent \textbf{Implications of correlation--NTK equivalence for learning.} Our results demonstrate the practical relevance of similarity propagation through the network. Because the NTK is asymptotically proportional to the output correlation at criticality, the correlation we characterise is what determines the NTK spectrum and its eigen-directions. Whether a given similarity structure helps or hinders learning depends on the task. For classification, inputs sharing a label are typically similar in a well-chosen representation, and this proximity must survive propagation to remain informative: excessive decorrelation (chaotic regime, or the input-norm-driven decorrelation discussed above) scrambles same-label inputs, while excessive collapse (ordered regime) makes even differently labeled inputs indistinguishable. For regression, the same similarity structure controls the regularity of the learned function, with collapse and decorrelation corresponding to over- and under-smoothing of the target, respectively. 

It is well established that input geometry can be decisive for performance, independently of the network’s bulk architecture: structured data distributions can substantially affect learning and generalization \citep{goldt2020modeling}, random feature maps approximating a task-relevant kernel improve classification accuracy without architectural adaptation \citep{RF-Rahimi}, and Fourier or sinusoidal embeddings counteract spectral bias in regression settings such as implicit neural representations \citep{tancik2020,sitzmann2020}. Our results offer a first-principles account of why such representation-level interventions are effective within the NTK regime: at criticality, the network provably propagates whatever similarity structure is present in its input to its output, without needing to construct task-relevant structure itself -- and, under orthogonal initialisation, this description remains accurate over a practical range of depths. \\

\noindent \textbf{The need for proper dataset normalisation.} The relationship $\rho_\infty^{\alpha\beta}(\rho_0^{\alpha\beta})$ characterised in Sec.~\ref{sub:corr_meanfield} reveals a source of decorrelation distinct from finite-size effects or from initialisation away from criticality: even in the mean-field limit, at criticality, when inputs have large pre-activation norms $K_0^{\alpha\alpha}$ and $K_0^{\beta\beta}$, the pre-activation kernel begins its algebraic decay from a larger value, giving the network's nonlinearity more depth over which to act, so that correlation is lost even between inputs that are not initially close to orthogonal. This gives a concrete, practical justification for input normalisation: anisotropy in input norms can induce an effectively chaotic loss of correlation even at criticality, and even a perfectly isotropic dataset -- distributed on a hypersphere $S^{n_\text{in}-1}$ -- suffers a similar degradation once its radius becomes too large. This raises a natural question: if nonlinearity destroys correlation, why not favour near-linear networks? Likely because the same nonlinearity is what lets the network reshape, rather than transport, input similarity -- a purely linear network preserves correlation perfectly, but cannot build task-relevant structure absent from the input. We leave a precise characterization of this trade-off to future work, which will be a key focus to understand, more precisely, what the features of the NTK are, outside the linearized regime in Sec.~\ref{sub-sec:NTK-dataset-alignment}.\\

\noindent \textbf{Connections to existing initialisation practice.} As a concrete example, consider sine networks (SIREN \citep{sitzmann2020}), commonly used to represent a signal, image, or physical field directly as a continuous function of its coordinates, as in physics-informed neural networks. The original SIREN architecture prescribes a weight variance corresponding to the \textit{chaotic} regime identified in Sec.~\ref{sub-section:large-depth-limit}, while the default \texttt{PyTorch} scheme instead places the same network in the \textit{ordered} regime (its bias variance, $\sigma_b^2 = 1/N$, sits asymptotically close to the vanishing-variance case, but remains in practice within the ordered phase). \citep{combette2025new} recently proposed initializing such networks at criticality instead, reporting improved performance on a range of such tasks -- but justified this choice only as one point along the broader edge of chaos line, without a specific argument for criticality itself. Here we provide this missing justification: criticality is the only choice that keeps the NTK controlled deep into the network while preserving information.\\

\noindent \textbf{Limitations and perspectives.} Several of our simplifying assumptions deserve further investigation. We work throughout with a fixed, finite dataset, which places us in a strongly over-parameterized regime. Bringing the analysis closer to practical settings will require allowing the dataset size to scale sublinearly with the number of network parameters, as observed empirically \citep{kaplan2020scaling}. In addition, for simplicity we restrict our study to fully connected networks. It would be interesting to investigate how the correlation--NTK relation generalizes to commonly used architectures such as MLPs with skip-connections \citep{yang2017}, CNNs \citep{xiao2018dynamical}, and transformers \cite{giorlandino2026two}, building on recent efforts toward a unified theoretical framework \citep{PDLT-2022,simon2026there}. Beyond these extensions, it would be valuable to disentangle the respective roles of architectural design choices and initialisation schemes in determining network performance on a given task.

We also restrict our derivation of finite-size corrections to the sequential limit itself, where $L/N \to 0$; the \emph{joint} limit, where $L/N$ is small but constant as $N$ diverges, has been thoroughly studied \citep{hanin2019finite,PDLT-2022} and is believed to underlie the onset of feature learning beyond the frozen-NTK regime. Our results suggest that this onset is itself initialisation-dependent, since orthogonal initialisation was found to suppress the $L/N$ fluctuations induced by Gaussian initialisation; a direct treatment of the joint limit, however, remains beyond our present scope.

Beyond these limitations, a few further directions stand out. First, while our theoretical framework applies to a large class of commonly used activation functions, our numerical validation has, in most cases, been restricted to the sine activation. Our results suggest a specific, testable prediction beyond sine activations: a SIREN-like network with sinusoidal input encoding but \texttt{tanh} in the bulk layers, initialised at the corresponding \texttt{tanh} critical point, should match the performance of the purely sinusoidal network initialised at criticality. This is consistent with the intuition that the bulk activation is secondary once a suitable frequency-rich input encoding and critical initialisation are chosen; we leave a systematic test of this prediction to future work. Second, extending our theoretical treatment to the class of unbounded, scale-invariant activation functions \citep{PDLT-2022} -- which includes ReLU, and hence the widely used Kaiming initialisation, as a limiting case -- is another natural direction. Finally, an intriguing open question concerns equivariant architectures, where spontaneously broken continuous symmetries can give rise to coherent information propagation \citep{iqbal2026spontaneous}, or to topologically protected defects offering a structurally distinct route to stable signal propagation \citep{iqbal2025topological}. It remains to be seen whether an analogous relation between the correlation structure and the NTK holds in these settings.\\


\textbf{Acknowledgements.}
We warmly thank Karol Kozlowsky for his input on random matrix theory and Eric Simonnet for useful discussions. We gratefully acknowledge the support of the Centre Blaise Pascal’s IT test platform at ENS de Lyon (Lyon, France) for the Machine Learning facilities. The platform operates the SIDUS solution
(Quemener and Corvellec, 2013), developed by Emmanuel Quemener. This work received funding from the French National Research Agency (ANR) under the PINOT project, grant agreement
ANR-25-CE56-2607.

\bibliography{ref}
\appendix
\section{Hypotheses on activation functions\label{app:activation}}

We consider a class of activation functions $\phi$, including
\texttt{tanh} and \texttt{sine}, satisfying the following four assumptions:

\noindent\textbf{H1} -- $\phi$ is bounded and differentiable on $\mathbb{R}$, with Taylor
expansion around the origin given by equation (\ref{eq:activation-expansion}): 
\begin{equation}
  \phi(x)
  \underset{x\to 0}{=}
  \sum_{p\geq 1}\frac{\phi_p}{p!}x^p,
  \qquad \phi_1>0,
\end{equation}

where $\phi_p\in\mathbb{R}$. We set $\phi_0=0$, as a constant term is
redundant with the bias $\rvb_\ell$, and assume $\phi_1>0$ without loss
of generality. 

\noindent \textbf{H2} -- For every $K>0$,
\begin{equation}
  \frac{\partial}{\partial K}
  \left(
    \frac{\E_{z\sim\mathcal{N}(0,K)}
    \left[\phi(z)^2\right]}{K}
  \right)<0.
  \label{eq:cond_uniq}
\end{equation}
This condition will ensure the uniqueness and stability of the positive
fixed point of the diagonal kernel recursion. Here $\E_{z\sim\mathcal{N}(0,K)}[\cdot]$ denotes expectation with respect to a zero-mean Gaussian random variable $z$ with variance $K$.

\noindent\textbf{H3} -- For every $K > 0$, 
\begin{equation}
  \frac{\partial}{\partial K}
  \left(
    \E_{z\sim\mathcal{N}(0,K)}
    \left[\phi(z)^2\right]
  \right) > 0.
  \label{eq:cond_convergence-3}
\end{equation}
This will give us the convergence of the diagonal kernel recursion to a unique fixed point for the whole real line.

\noindent\textbf{H4} -- For every $K>0$ and $K^{\alpha\beta}\in[-K,K]$,
\begin{equation}
  \frac{\partial}{\partial K^{\alpha\beta}}
  \E_{(z^{\alpha}, z^{\beta}) \sim\mathcal{N}(0,\mSigma)}
  \left[\phi(z^\alpha)\phi(z^\beta)\right]>0,
  \qquad
  \mSigma=
  \begin{pmatrix}
    K & K^{\alpha\beta}\\
    K^{\alpha\beta} & K
  \end{pmatrix}.
  \label{eq:cond_convergence-4}
\end{equation}
This condition will be useful for deriving the convergence of the off-diagonal kernel recursion. Here $\E_{(z^{\alpha}, z^{\beta}) \sim\mathcal{N}(0,\mSigma)}[\cdot]$ denotes expectation with respect to the bivariate Gaussian random vector $(z^\alpha,z^\beta)$ with covariance matrix $\mSigma$.

\section{Susceptibilities and Price theorem} \label{app:susceptilibilities}
In this section we give the Gaussian expression of the susceptibilities and the elements to derive them. We begin by recalling Price's theorem \citep{Price-1958}: 
\begin{equation} 
  \partial_{K_{\ell}^{\mu \gamma}} \E \left[ \phi(z_\ell^{\alpha}) \phi(z_\ell^{\beta})\right] = \left(\frac{1}{2}\right)^{\delta_{\mu \gamma}} \E \left[\partial_{z^{\mu}} \partial_{z^{\gamma}} \phi(z_\ell^{\alpha}) \phi(z_\ell^{\beta}) \right] \qquad \mbox{with} \quad \mu,\gamma \in \{\alpha,\beta\}.
\end{equation}
Hence by direct application of the theorem to the definition (\ref{eq:allchi_def}) we have: 

\begin{equation}
\label{eq:chi_terms}
\begin{aligned}
  \chi^{\perp}_\ell &= \sigma_w^2 \, \E \left[ \phi'(z^\alpha_\ell) \phi'(z^\beta_\ell) \right], \\
  \chi^{\alpha}_\ell &= \sigma_w^2 \, \E \left[ \phi'(z_\ell^{\alpha})^2 + \phi(z_\ell^{\alpha}) \phi''(z_\ell^{\alpha}) \right], \\
  \chi^{\beta}_\ell &= \sigma_w^2 \, \E \left[ \phi'(z_\ell^{\beta})^2 + \phi(z_\ell^{\beta}) \phi''(z_\ell^{\beta}) \right], \\
  \chi^{\bullet}_\ell &= \sigma_w^2 \, \E \left[ \phi''(z^\alpha_\ell) \phi(z^\beta_\ell) \right], \\
  \chi^{\circ}_\ell &= \sigma_w^2 \, \E \left[ \phi(z^\alpha_\ell) \phi''(z^\beta_\ell) \right].
\end{aligned}
\end{equation}
\section{Covariance Matrix and Fixed point}\label{app:fixed-point-discussion}
In this section, we show that under the assumption we stated in Sec.~\ref{sec:architecture}, several properties can be unearthed on the convergence of the covariance matrix $\mSigma_\ell^{\alpha \beta}$ with depth towards one of its fixed points $\mSigma_\star^{\alpha \beta}$ of the form given in \eqref{eq:gaussian-kernel}. We start by decomposing $\mathscr{F}$ as a diagonal $\mathscr{F}_K$ and an off-diagonal component $\mathscr{F}_C$, exposed in \eqref{eq:rec-variance} and \eqref{eq:rec-covariance}, namely: 
\begin{equation} \label{eq:covariance-restriction}
  \begin{cases}
    K^{\alpha \alpha}_{\ell + 1} = \mathscr{F}_K(K^{\alpha \alpha}_{\ell}) ,\\
    K^{\alpha \beta}_{\ell + 1} = \mathscr{F}_C(K^{\alpha \beta}_{\ell}, K^{\alpha \alpha}_{\ell}, K^{\beta \beta}_{\ell}).
  \end{cases}
\end{equation}

\subsection{Existence, uniqueness and stability of the diagonal fixed point}

The primary focus of this first part is to demonstrate why the diagonal elements of any fixed point $\mSigma_{\star}^{\alpha \beta}$ are equal for nonzero inputs $\boldsymbol{x}^{\alpha}, \boldsymbol{x}^{\beta}$, implying a unique stable fixed point for the variance. We begin by rewriting the fixed point condition:
\begin{equation}
  \label{eq:fixed-point-cond}
  K_{\star} = \mathscr{F}_K(K_{\star}):= \sigma_w^2 \E \left[ \phi(z)^2 \right] + \sigma_b^2 \quad \text{with} \quad z \sim \mathcal{N}(0, K_{\star}). 
\end{equation}
We consider two distinct cases depending on the value of $\sigma_b^2$.

\paragraph{Case 1: $\sigma_b^2 > 0$.} In this scenario, we observe that $K_{\star} \ge \sigma_b^2$. We introduce the function
\begin{align}
  g \colon 
    [\sigma_b^2, \infty) &\to (0,\eta], \\
    K &\mapsto \frac{\E\left[ \phi(z)^2 \right]}{K},
\end{align}
where $\eta > 0$. We use the fact that $\phi$ is bounded (\textbf{H1}) to derive $\lim_{K\to \infty}g(K) = 0$. A supplementary condition on $\phi$ that ensures the existence of a unique fixed point $K_{\star} \in [\sigma_b^2, \infty)$ using the bijection theorem is given by hypothesis (\textbf{H2}) that can be reformulated as: 
\begin{equation} \label{eq:hypothesis-activation-original}
  \partial_K \E\left[ \phi(z)^2 \right] K - \E\left[ \phi(z)^2 \right] < 0,
\end{equation}
which ensures that $g$ is decreasing.
Using Price's theorem and Gaussian integration by parts, the left-hand side writes
\begin{equation} 
\label{eq:intermediate}
  \partial_K \left(\E\left[ \phi(z)^2 \right]\right) K = \E\left[ z \phi(z) \phi(z)'\right].
\end{equation}
One can rewrite \eqref{eq:hypothesis-activation-original} using the previous equality: 
\begin{equation} \label{eq:hypothesis-activation}
  \E\left[ \phi(z) \left(z\phi'(z) - \phi(z)\right) \right] < 0.
\end{equation}
Furthermore, the stability of this fixed point can be analyzed as follows. Using \eqref{eq:fixed-point-cond} and \eqref{eq:intermediate} leads to
\begin{equation}
   \frac{\partial \mathscr{F}_K}{\partial K} = \frac{\sigma_w^2}{K} \E \left[ \phi(z) \phi'(z) z \right] < \frac{\sigma_w^2}{K} \E\left[ \phi(z)^2 \right],
\end{equation}
where the second inequality is a consequence of \eqref{eq:hypothesis-activation}. At the fixed point $K_{\star}$, using \eqref{eq:fixed-point-cond}, we obtain:
\begin{equation} \label{eq:stability-variance}
  \chi^{\parallel}_\star =  \frac{\partial \mathscr{F}}{\partial K} \bigg\vert_{K_\star} < 1 - \frac{\sigma_b^2}{K_{\star}}.
\end{equation}
Using $K_{\star} > \sigma_b^2$ and (\textbf{H3}) gives us $0< \chi^{\parallel}_\star< 1$. This confirms that the unique fixed point $K_{\star}$ is stable.

\paragraph{Case 2: $\sigma_b^2 = 0$.} The uniqueness of a global fixed point is no longer guaranteed. However, we can still establish the uniqueness of a \textit{stable} fixed point for every $\sigma_w$. $K_{\star} = 0$ is always a fixed point, given that $\phi(0)=0$. Its stability depends on the condition $\sigma_w \phi_1 < 1$, indeed from (\ref{eq:chi_terms}), a direct computation leads to: 
\begin{equation}
  \chi^{\parallel}_\star = \sigma_w^2 \phi_1^2.
\end{equation}
To study its uniqueness and the emergence of other fixed points, we follow an approach similar to that used above. The function $g(K)$ remains strictly decreasing under our hypothesis, but its domain must be redefined as:
\begin{align}
  g \colon (0, \infty) &\to (0,\eta), \\
    K &\mapsto \frac{\E\left[ \phi(z)^2 \right]}{K},
\end{align}
where the constant $\eta$ is now given by:
\begin{equation}
  \eta = \lim_{K \to 0^+} g(K) = \phi_1^2.
\end{equation}
\noindent The value of $\sigma_w \phi_1$ plays here also a critical role:
\begin{itemize}
  \item If $\sigma_w \phi_1 < 1$, then $g(K) < \frac{1}{\sigma_w^2}$ for all $K > 0$, and no strictly positive fixed point exists. And $K_\star = 0$ is the unique stable fixed point
  \item If $\sigma_w \phi_1 = 1$, then $K_\star = 0$, its stability is however not prescribed.
  \item If $\sigma_w \phi_1 > 1$, there exists a strictly positive fixed point $K_{\star}$, and the stability result from \eqref{eq:stability-variance} confirms that this fixed point is stable.
\end{itemize}
Combining these results, we conclude that if there exists a stable fixed point $K_\star$ it is always unique, and its existence is obtained in the entire plane $(\sigma_w, \sigma_b) \setminus \{(1 / \phi_1, 0)\}$. This analysis extends the results of \cite{hayou2019impact}.

\subsection{Basin of attraction of the diagonal fixed point}
In this section, we show that for each $(\sigma_w, \sigma_b)$ and for non-zero inputs, there is exactly one basin of attraction covering $\mathbb{R}^{+*}$. More precisely, for non-zero inputs, the recurrence relation (\ref{eq:rec-variance}) converges to the unique stable fixed point $K_\star$ across the entire plane $(\sigma_w, \sigma_b) \setminus \{(1 / \phi_1, 0)\}$, and to $K_\star = 0$ at the point $(1 / \phi_1, 0)$.

By hypothesis (\textbf{H3}), $\mathscr{F}_K$ is strictly increasing. Let $K_0^{\alpha \alpha} \in \mathbb{R}^{+*}$ be the first term of the sequence (\ref{eq:rec-variance}). This directly implies that $K^{\alpha \alpha}_\ell$ remains on the same side of $K_\star$ as $K^{\alpha \alpha}_0$. We now consider two cases:

\paragraph{Case 1: $K_\star > 0$}
Using (\textbf{H2}), we rewrite the expression for $\mathscr{F}_K$:
  \begin{equation}
  \mathscr{F}_K - K = \sigma_w^2 K \left[ g(K) - g(K_\star)\right] + (1 - \frac{K}{K_\star}) \sigma_b^2,
\end{equation}
which directly implies that $K_\ell$ decreases if $K^{\alpha \alpha}_0 > K_\star$ and increases otherwise. Thus, for every initial condition $K^{\alpha \alpha}_0$, the sequence converges, as it is bounded and monotone. By the continuity of $\mathscr{F}_K$, it converges to $K_\star$.

\paragraph{Case 2: $K_\star = 0$}
In this case, we derive a similar expression using $\lim_{K \to 0} g(K) = (\phi_1)^2$ and recalling that $\sigma_b = 0$:
\begin{equation}
   \mathscr{F}_K - K = \sigma_w^2 K \left[ g(K) - g(K_\star)\right] + K \left((\sigma_w \phi_1)^2 - 1\right)
\end{equation}
which converges to $K_\star = 0$ for all $K^{\alpha \alpha}_1 \in \mathbb{R}^{+*}$ only if $\sigma_w \phi_1 \le 1$. If $\sigma_w \phi_1 > 1$, $K_\star = 0$ is unstable, and we return to the previous case, considering the unique stable fixed point $K_\star > 0$.

\subsection{Finite Number of Fixed Points for $\mathscr{F}$}
Now that we have characterized the diagonal recurrence relation generated by $\mathscr{F}_K$, to fully determine the global recurrence, we must also study $\mathscr{F}_C$ in \eqref{eq:covariance-restriction}, the restriction of $\mathscr{F}$ to the off-diagonal elements.

\paragraph{Asymptotic Autonomy}
First, note that since $\mathscr{F}_C$ is continuous (as it is a Gaussian integral), we can focus on the limiting autonomous map $\mathscr{F}_C(\, \cdot \, , K_\star, K_\star)$ to analyze the convergence behaviour of the non-autonomous sequence.

\paragraph{Finiteness of the Fixed-Point Set}
Another key observation is that $K_\star$ is already a fixed point of $\mathscr{F}_C(\, \cdot \, , K_\star, K_\star)$. Thus, we only need to show that there is a finite number of such fixed points. This follows directly from the analyticity of $\mathscr{F}_C(\, \cdot \, , K_\star, K_\star)$, assuming it is not the identity map. Indeed, the function $K^{\alpha \beta} \mapsto \mathscr{F}_C(K^{\alpha \beta}, K_\star, K_\star) - K^{\alpha \beta}$ is analytic on the interval $[-K_\star, K_\star]$ and non-zero, so it admits a finite number of fixed points. Therefore, $\mathscr{F}_C$ also has a finite number of fixed points, and the same conclusion holds for the global map $\mathscr{F}$.

\subsection{Basin of convergence of the off-diagonal recursion}

Now that we know that there is a finite number of fixed points, let us determine whether their basins of convergence cover the entire admissible covariance interval $[-K_\star,K_\star]$. In other words, we want to show that, for every $K^{\alpha\beta}_0$ in this interval, the sequence generated by the limiting map converges. Hypothesis~(\textbf{H4}) ensures that $\mathscr{F}_C(\,\cdot\,,K_\star,K_\star)$ is increasing. Consequently, the sequence $(K^{\alpha\beta}_\ell)_{\ell\in\mathbb{N}}$ is monotone for every initial condition. Since it is also bounded, it converges. Therefore, the basins of convergence of these fixed points cover the entire admissible interval.

As stated previously for the non-autonomous recursion, the convergence of the diagonal sequences to $K_\star$, together with the continuity of $\mathscr{F}_C$, ensures that the layer-dependent maps asymptotically approach the limiting map. Giving the same fixed point and ensuring that the basins of convergence covers the admissible interval. Nevertheless, the two recursions do not necessarily have the same basins of convergence, since the transient dynamics may affect which fixed point is ultimately selected.

\subsection{Discussion on Hypotheses (\textbf{H1-H4})}

\paragraph{Discussion of Hypothesis \textbf{H2}}
We verify that hypothesis (\textbf{H2}), equivalently stated in \eqref{eq:hypothesis-activation}, is reasonable and holds for common activation functions.

\subparagraph{Common sigmoidal functions}
A sufficient condition for \eqref{eq:hypothesis-activation} is that $\phi$ satisfies:
\begin{equation} \label{eq:activation-pointwise-hypothesis}
 \forall x \neq 0 \quad \frac{d}{dx}\frac{\phi(x)}{\|x\|} < 0 \quad \text{and} \quad \phi(x) x > 0 .
\end{equation}
Indeed, these conditions directly imply:
\begin{equation} \label{eq:reduced-hypothesis-activation}
  \forall x \in \mathbb{R}^* \quad \phi(x) \left(\phi'(x)x - \phi(x)\right) < 0
\end{equation}

\begin{figure}[ht]
  \centering
  \includegraphics[width = \linewidth]{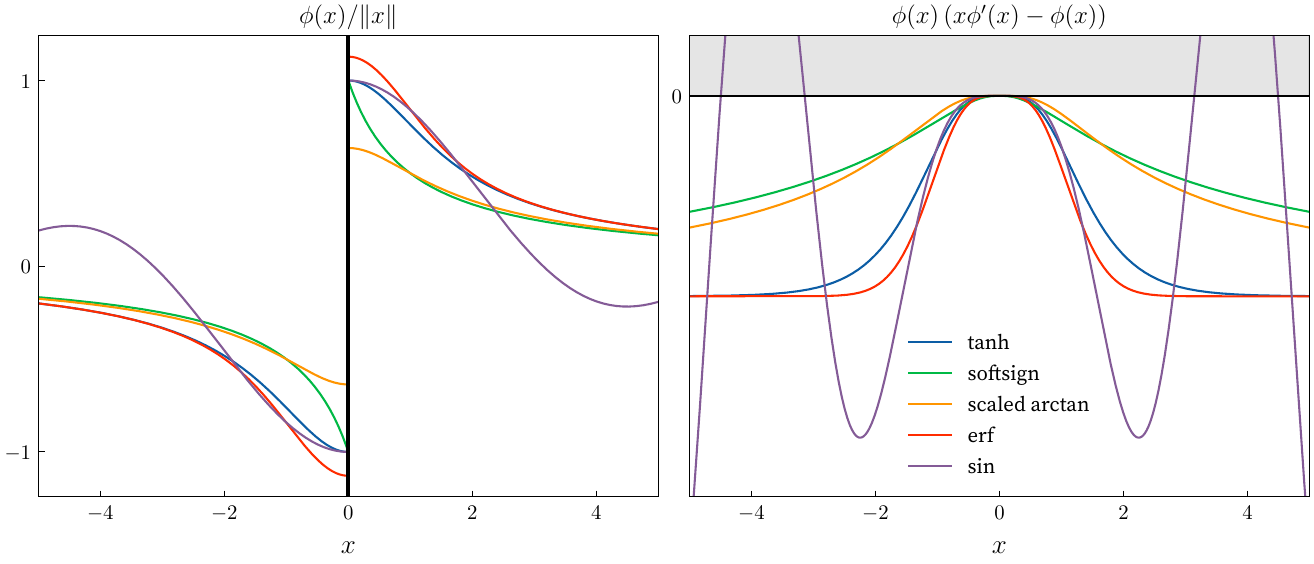}
  \caption{The two sufficient conditions \eqref{eq:reduced-hypothesis-activation} and \eqref{eq:activation-pointwise-hypothesis} are plotted for several activation functions. They are satisfied by all the sigmoidal functions considered here. For the \texttt{sine} activation function, the pointwise inequality does not hold, and the Gaussian average must therefore be evaluated explicitly.
  \label{fig:reduced-hypothesis}}
\end{figure}

Since this inequality holds pointwise for all $x \neq 0$, it remains valid after taking the Gaussian average. Therefore, hypothesis \eqref{eq:hypothesis-activation} is satisfied. As illustrated in Fig.~\ref{fig:reduced-hypothesis}, this condition applies to common sigmoidal activation functions such as \texttt{tanh} and its variants, including \texttt{erf} and \texttt{softsign}, as can be checked directly.

\subparagraph{The \texttt{sine} Function}
For the sine function, such a pointwise inequality is not available. However, the expectation can be computed explicitly:
\begin{equation}
  \E\left[\sin(z)^2\right] = \frac{1}{2}\left(1 - e^{-2K} \right).
\end{equation}
Using the equivalence between \eqref{eq:hypothesis-activation-original} and \eqref{eq:hypothesis-activation}, we obtain:
\begin{align}
   \E\left[ \phi(z) \left(z\phi'(z) - \phi(z)\right) \right] &= e^{-2 K} K - \frac{1}{2}\left(1 - e^{-2K} \right) \\
   &= \frac{1}{2} \left( (2 K + 1) e^{-2K} - 1 \right).
\end{align}
Since $e^{2K} > 2K + 1$ for all $K > 0$, the desired inequality \eqref{eq:hypothesis-activation} follows.

\paragraph{Discussion of Hypothesis (\textbf{H3})}
We now verify that this hypothesis holds for the activation functions considered above. Equation~\eqref{eq:intermediate} shows that it is sufficient for $\phi$ to satisfy:
\begin{equation}
  \forall x \in \mathbb{R}, \quad \frac{d}{dx} \phi(x) > 0 \quad \text{and} \quad \phi(x) x > 0.
\end{equation}
These conditions are readily verified for \texttt{tanh}, \texttt{erf}, and similar sigmoidal functions, and therefore imply (\textbf{H3}). For the \texttt{sine} activation function, the pointwise argument does not apply, but a direct computation gives:
\begin{equation}
  \E\left[ \phi(z) z\phi'(z) \right] = K e^{-2K} > 0.
\end{equation}
Thus, (\textbf{H3}) also holds for the sine function.

\paragraph{Discussion of Hypothesis (\textbf{H4})}
We finally verify (\textbf{H4}) for the same activation functions. From the previous discussion of the susceptibilities (see \eqref{eq:chi_terms}), for all $K^{\alpha \alpha} > 0$ and $K^{\alpha \beta} \in \mathbb{R}$, we have:
\begin{equation}
  \frac{\partial}{\partial K^{\alpha\beta}}
  \E_{\vz\sim\mathcal{N}(0,\mSigma)}
  \left[\phi(z^\alpha)\phi(z^\beta)\right] = \E_{\vz\sim\mathcal{N}(0,\mSigma)}
  \left[\phi'(z^\alpha)\phi'(z^\beta)\right]
\end{equation}
with $\mSigma = \begin{pmatrix}
 K^{\alpha \alpha} & K^{\alpha \beta} \\ K^{\alpha \beta} & K^{\alpha \alpha}
\end{pmatrix}$. Hence, if $\phi' > 0$, hypothesis (\textbf{H4}) follows directly. This applies, in particular, to the sigmoidal functions considered above, such as \texttt{tanh} and \texttt{erf}. For the \texttt{sine} activation function, the expectation can instead be evaluated explicitly:
\begin{equation}
   \E_{\vz\sim\mathcal{N}(0,\mSigma)}
  \left[\phi'(z^\alpha)\phi'(z^\beta)\right] = \cosh(K^{\alpha \beta})e^{-K^{\alpha \alpha}} 
\end{equation}
This quantity is strictly positive, so (\textbf{H4}) is also satisfied.
\section{Correlation \label{app:correlation}}
\subsection{Recurrence relation} \label{eq:corr_rec}
Here we derive the correlation expression and show \eqref{eq:rho-map}. Let's delve into the calculations, using the recurrence relation \eqref{eq:covariance-recursion}: 
\begin{align}
  \rho_{\ell + 1}^{\alpha \beta} & = \frac{K_{\ell+1}^{\alpha\beta}}{\sqrt{K_{\ell+1}^{\alpha\alpha} K_{\ell+1}^{\beta\beta}}} \\ &=\frac{K_{\ell}^{\alpha\beta}}{\sqrt{K_{\ell}^{\alpha\alpha} K_{\ell}^{\beta\beta}}} \left[\frac{1 + \tfrac{\phi_{2}^2}{4\phi_{1}^2}( \frac{K_{\ell}^{\alpha \alpha} K_{\ell}^{\beta\beta}}{K_{\ell}^{\alpha\beta}} + 2 K_\ell^{\alpha \beta}) + \tfrac{\phi_{3}}{2\phi_{1}} (K_{\ell}^{\alpha\alpha} + K_{\ell}^{\beta\beta}) + \mathcal{O}(K^2)}{1 + \tfrac{A}{2}( K^{\alpha\alpha}_{\ell} + K^{\beta\beta}_{\ell}) + \mathcal{O}(K^2)} \right]
  \\
                  & = \rho_{\ell}^{\alpha \beta} \left[  1 + (\tfrac{\phi_2}{\phi_{1}})^2 \left(\tfrac{1}{4}(\rho_{\ell}^{-2} + 2) K^{\alpha \beta }_\ell - \tfrac{3}{8}(K^{\beta \beta }_{\ell } + K^{\alpha \alpha }_{\ell} ) \right) + \mathcal{O}(K^2)\right],
\end{align}
where A is defined in \eqref{eq:def_a1}.
\subsection{Finite size effects} \label{app-finite-size}

\paragraph{Induced error terms} 
\label{par:induced_err}
The mean-field perturbed recurrence for the empirical kernel $\widehat{K}^{\alpha \beta}$ is derived as follows:
\begin{align}
    \widehat{K}_{\ell+1}^{\alpha\beta} &= \frac{1}{N} \rvz^{\alpha \top}_{\ell + 1} \rvz^{\beta}_{\ell + 1}\\
    &= \frac{1}{N} \phi(\rvz^{\alpha}_{\ell})^\top \frac{\rmW_{\ell + 1}^\top \rmW_{\ell + 1}}{N} \phi(\rvz^{\beta}_{\ell}) \\
    &= \phi(\rvz^{\alpha})^\top \left( \frac{\rmW_{\ell + 1}^\top \rmW_{\ell + 1}}{N} - \sigma_w^2 \mathbf{I} \right)\phi(\rvz^{\beta}) + \frac{\sigma_w^2}{N} \phi(\rvz^{\alpha})^\top \phi(\rvz^{\beta}) \label{eq:empirical-kernel-derivation-1} \\
    &= \sigma_w^2 \sum_{p, q \ge 0} \frac{\phi_p \phi_q}{q! p!} M_\ell^{\alpha \beta}(p,q) + \sigma_w^2 \sum_{p, q \ge 0} \frac{\phi_p \phi_q}{q! p!}  \left[\widehat{M}_\ell^{\alpha \beta}(p,q) - M_\ell^{\alpha \beta}(p,q)\right] \label{eq:empirical-kernel-derivation-2}
  \\ &+ \phi(\rvz^{\alpha})^\top \left( \frac{\rmW_{\ell + 1}^\top \rmW_{\ell + 1}}{N} - \sigma_w^2\mathbf{I}\right)\phi(\rvz^{\beta}). \notag \\
  &= \underbrace{\mathscr{F}_C(\widehat{K}_{\ell}^{\alpha\beta}, \widehat{K}_{\ell}^{\alpha\alpha}, \widehat{K}_{\ell}^{\beta\beta})}_{\text{MF}} + \underbrace{\sigma_w^2 \sum_{p, q \ge 0} \frac{\phi_p \phi_q}{q! p!}  \left[\widehat{M}_\ell^{\alpha \beta}(p,q) - M_\ell^{\alpha \beta}(p,q)\right] }_{\varepsilon_{\text{Wick}}}
  \\ &+ \underbrace{\phi(\rvz^{\alpha})^\top \left( \frac{\rmW_{\ell + 1}^\top \rmW_{\ell + 1}}{N} - \sigma_w^2\mathbf{I}\right)\phi(\rvz^{\beta})}_{\varepsilon_{\text{Weight}}}. \notag
  \end{align}

To go from (\ref{eq:empirical-kernel-derivation-1}) to (\ref{eq:empirical-kernel-derivation-2}), we use the definition of $\phi$ in \eqref{eq:activation-expansion}, and define $\widehat{M}_\ell^{\alpha \beta}(p,q)$ the empirical mixed moment of order $(p,q)$ as 
\begin{equation}
  \widehat{M}_\ell^{\alpha \beta}(p,q) = \frac{1}{N} \sum_{i = 1}^{N} (z_{\ell,i}^{\alpha})^p (z_{\ell,i}^{\beta})^q, \label{eq:app_Mhat}
\end{equation} 
and \(M_\ell^{\alpha \beta}(p,q)\) its Wick contraction in the Gaussian case:
\begin{equation}
  M^{\alpha \beta}_{\ell}(p,q) = \sum_{\pi \in \mathcal{P}^{\alpha \beta}(p,q)} \prod_{\{\gamma , \delta \} \in \pi} \widehat{K}^{\gamma \delta}_\ell, \label{eq:app_M}
\end{equation}
where $\mathcal{P}^{\alpha \beta}(p,q)$ denotes all the pairings (partitions of pairs) of $ \{\overbrace{\alpha,\ldots,\alpha}^p,\overbrace{\beta, \ldots,\beta}^q\}$ and $\pi$ a partition of $\mathcal{P}^{\alpha \beta}(p,q)$.
The final equality comes from noticing that $\mathscr{F}$ is exactly defined using the Wick contraction on the Gaussian moments. 

\paragraph{Error propagation}
\label{par:error_prop} 
In order to establish the relation between $ \delta_\ell^{\alpha \beta}$ and $ \varepsilon_\ell^{\alpha \beta}$, we start by combining \eqref{eq:delta} and \eqref{eq:reckernel}:
\begin{align}
   \begin{pmatrix}\delta_{\ell+1}^{\alpha \beta} \\
      \delta_{\ell+1}^{\alpha \alpha} \\
      \delta_{\ell+1}^{\beta \beta}
     \end{pmatrix} &=  \begin{pmatrix} \widehat{K}_{\ell+1}^{\alpha \beta} \\
      \widehat{K}_{\ell+1}^{\alpha \alpha} \\
      \widehat{K}_{\ell+1}^{\beta \beta}
     \end{pmatrix} - \begin{pmatrix} K_{\ell+1}^{\alpha \beta} \\
      K_{\ell+1}^{\alpha \alpha} \\
      K_{\ell+1}^{\beta \beta}
     \end{pmatrix} \\
   &= \mathscr{F}(\widehat{K}_{\ell}^{\alpha\beta}, \widehat{K}_{\ell}^{\alpha\alpha}, \widehat{K}_{\ell}^{\beta\beta}) + \begin{pmatrix}
     \varepsilon_{\ell + 1}^{\alpha \beta} \\
      \varepsilon_{\ell + 1}^{\alpha \alpha} \\
      \varepsilon_{\ell + 1}^{\beta \beta}
     \end{pmatrix} -  \mSigma_{\ell + 1}^{\alpha \beta}  \\
     &= \mathscr{F}(\mSigma_{\ell}^{\alpha \beta} ) + \rmJ_{\mathscr{F}}(\mSigma_\ell^{\alpha \beta}) \begin{pmatrix}\delta_{\ell}^{\alpha \beta} \\
      \delta_{\ell}^{\alpha \alpha} \\
      \delta_{\ell}^{\beta \beta}
     \end{pmatrix} + \begin{pmatrix}
     \varepsilon_{\ell + 1}^{\alpha \beta} \\
      \varepsilon_{\ell + 1}^{\alpha \alpha} \\
      \varepsilon_{\ell + 1}^{\beta \beta}
     \end{pmatrix} - \mSigma_{\ell + 1}^{\alpha \beta} + \mathcal{O}(\delta_\ell^2)
     \\
  &= \rmJ_{\mathscr{F}}(\mSigma_\ell^{\alpha \beta}) \begin{pmatrix}\delta_{\ell}^{\alpha \beta} \\
      \delta_{\ell}^{\alpha \alpha} \\
      \delta_{\ell}^{\beta \beta}
     \end{pmatrix} + \begin{pmatrix}
     \varepsilon_{\ell + 1}^{\alpha \beta} \\
      \varepsilon_{\ell + 1}^{\alpha \alpha} \\
      \varepsilon_{\ell + 1}^{\beta \beta}
     \end{pmatrix} + \mathcal{O}(\delta_\ell^2)
\end{align}
Then, noticing that $ \forall \alpha, \beta ,\; \delta_1^{\alpha \beta} = \varepsilon^{\alpha \beta}_1$ by immediate recursion, we get:
\begin{equation} \label{eq:perturbation-transport-matrix}
  \begin{pmatrix}
     \delta_{\ell+1}^{\alpha \beta} \\
      \delta_{\ell+1}^{\alpha \alpha} \\
      \delta_{\ell+1}^{\beta \beta}
     \end{pmatrix}
     = \sum_{\ell' = 1}^{\ell + 1} \prod_{\ell'' = \ell'}^{\ell} \mathbf{J}_\mathscr{F}(\mSigma_{\ell''}^{\alpha \beta})
     \begin{pmatrix}
     \varepsilon_{\ell'}^{\alpha \beta} \\
      \varepsilon_{\ell'}^{\alpha \alpha} \\
      \varepsilon_{\ell'}^{\beta \beta}
     \end{pmatrix} ,
\end{equation}
where the convention $\prod_{\ell + 1}^{\ell} \cdot = 1$ has been used. The matrix $\mathbf{J}_\mathscr{F}(\mSigma_{\ell}^{\alpha \beta})$ is defined in \eqref{eq:jaccov}. From \eqref{eq:perturbation-transport-matrix}, it is clear that understanding the depth dependence of $\chi^{\perp}_\ell$, $\chi^{\parallel}_{\ell}$, $\chi^{\bullet}_{\ell}$, and $\chi^{\circ}_{\ell}$ is crucial.

\paragraph{Scaling at criticality}
\label{par:scaling_criticality}
At criticality, their expressions are remarkably simple. Indeed, from \eqref{eq:covariance-recursion}, we directly obtain:
\begin{equation} \label{eq:scaling-chis}
  \begin{cases}
  \chi^{\perp}_\ell = 1 + \frac{\phi_3}{2 \phi_1}\left(K^{\alpha \alpha}_\ell + K^{\beta \beta}_\ell \right) + \left(\frac{\phi_2}{\phi_1} \right)^2K^{\alpha \beta}_\ell + \dots \\
  \chi^{\parallel}_\ell = 1 - 2 A K^{\alpha \alpha}_\ell + \dots \\
  \chi^{\bullet}_{\ell}= \frac{\phi_3}{2 \phi_1}K^{\alpha \beta}_\ell + \dots \\
  \chi^{\circ}_{\ell}= \frac{\phi_3}{2 \phi_1} K^{\alpha \beta}_\ell + \dots
  \end{cases},
\end{equation}
where $A$ is introduced in \eqref{eq:def_a1} and we replace $\chi^{\alpha}_\ell$ and $\chi^{\beta}_\ell$ by the parallel susceptibility $\chi^{\parallel}_\ell$ since both scale equally at criticality. Here and in the following ``\dots'' denotes terms of order $O((\log \ell) / \ell^2)$. We are now interested in the behaviour of the diagonal products, appearing in \eqref{eq:perturbation-transport-matrix} $\prod_{\ell'' = \ell'}^\ell \chi^{\perp}_{\ell''}$ and $\prod_{\ell'' = \ell'}^\ell \chi^{\parallel}_{\ell''}$. From the expansions derived above \eqref{eq:scaling-chis}, we see that these products can be easily studied through their logarithmic sum, this gives us for $\chi^{\perp}_\ell$: 
\begin{align} 
  \log \chi^{\perp}_\ell &= \frac{\phi_3}{2 \phi_1}\left(K^{\alpha \alpha}_\ell + K^{\beta \beta}_\ell \right) + \left(\frac{\phi_2}{\phi_1} \right)^2K^{\alpha \beta}_\ell + \dots \\
  &= \frac{1}{A\ell} \left(\left(\tfrac{\phi_{2}}{\phi_{1}}\right)^2 \rho^{\alpha \beta}_{\ell} + \tfrac{\phi_{3}}{\phi_{1}}\right) + \dots \\
  &= \frac{B^{\alpha \beta}}{\ell} + \dots \quad \text{with} \quad B^{\alpha \beta} = -\frac{{\phi_3}/{\phi_1} + \rho^{\alpha\beta}_{\infty} 
  \left({\phi_2}/{\phi_1}\right)^{\!2}}{
  {\phi_3}/{\phi_1} + \tfrac{3}{4}\left({\phi_2}/{\phi_1}\right)^{\!2}},
\end{align}
where we used the algebraic convergence of $\rho^{\alpha \beta}_\ell = \rho^{\alpha \beta}_\infty + \dots$, yielding a higher order term. One can notice that setting $\phi_2 = 0$ to preserve the correlation flow at criticality we have $B^{\alpha \beta} = -1$. In the particular case $\alpha=\beta$, we recover a formula previously computed by \cite{PDLT-2022}. Similarly for $\chi^{\parallel}_\ell$: 
\begin{align} 
  \log \chi^{\parallel}_\ell &= - 2AK^{\alpha \alpha} + \dots \\
  &= \frac{-2}{
  \ell} + \dots \,.
\end{align}
Hence, we have directly for their product: 
\begin{equation} \label{eq:product-susceptibility-pre}
\begin{cases}
\prod_{\ell' = 1}^{\ell}\chi^{\perp}_{\ell'} = \exp \left[ \sum_{\ell' =1}^\ell \left( \frac{-1}{
  \ell'} + \dots \right)\right] = \mathcal{O}( 1 / \ell), \\
  \prod_{\ell' = 1}^{\ell} \chi^{\parallel}_{\ell'}= \exp \left[ \sum_{\ell' =1}^\ell \left( \frac{-2}{
  \ell'} + \dots \right)\right] = \mathcal{O}( 1 / \ell^2).
\end{cases}
\end{equation}
Considering now the truncated products, we have for both cases: 
\begin{equation} \label{eq:product-susceptibility}
\begin{cases}
\prod_{\ell'' = \ell'}^{\ell}\chi^{\perp}_{\ell''} \overset{\ell, \ell' \to \infty}{\sim} \ell'/\ell , \\
\prod_{\ell'' = \ell'}^{\ell}\chi^{\parallel}_{\ell''} \overset{\ell, \ell' \to \infty}{\sim} (\ell'/\ell )^2 .
\end{cases}
\end{equation}
These different exponents are related to the critical exponent reported in \cite{PDLT-2022}.

\paragraph{Study of $\varepsilon_{\text{Wick}}$}
\label{par:eps_wick}

 Wick defects are controlled by the first non-zero term, where we dropped all unnecessary coefficients for the remaining derivations:
 
 \begin{equation} \label{eq:wick-deffects}
  \varepsilon_{\text{Wick}}^1 = \frac{1}{N} \sum_{i = 1}^{N} (z_{\ell, i}^{\alpha})^4 - 3(\widehat{K}^{\alpha \alpha}_\ell)^2.
\end{equation}
where \begin{equation}
(\widehat{K}^{\alpha \alpha}_\ell)^2 = \frac{1}{N^2} \sum_{p,p'} (z_{\ell,p}^{\alpha})^2 (z_{\ell,p'}^{\alpha})^2.
\end{equation}
We now explicitly compute its bias and variance. At first order, one can consider for such perturbations, the leading Gaussian case: \(\rvz \sim \mathcal{N}(0, K^{\alpha \alpha} \mathbf{I})\), which will considerably simplify the computation. 
\begin{itemize}
  \item \textit{Bias study} -- 
By Wick's theorem, $\E[(z_{\ell,i}^{\alpha})^4] = 3(K^{\alpha \alpha}_\ell)^2$. For the second term, we distinguish two cases: if $p \neq p'$, $\E[(z_{\ell,p}^{\alpha})^2 (z_{\ell,p'}^{\alpha})^2] = (K^{\alpha \alpha}_\ell)^2$; if $p = p'$, $\E[(z_{\ell,p}^{\alpha})^4] = 3(K^{\alpha \alpha}_\ell)^2$. Combining these results, we obtain:
\begin{equation} \label{eq:bias-wick}
\E\left[\varepsilon_{\text{Wick}}^1\right] = -\frac{6(K^{\alpha \alpha}_\ell)^2}{N}.
\end{equation}

\item \textit{Variance study} --
The variance is given by:
\begin{equation}
\Var\left[\varepsilon_{\text{Wick}}^1\right] = \E\left[\left(\varepsilon_{\text{Wick}}^1\right)^2\right] - \E\left[\varepsilon_{\text{Wick}}^1\right]^2.
\end{equation}
Expanding the square using \eqref{eq:wick-deffects}, we have:
\begin{equation}
  \begin{split}
  \left(\frac{1}{N} \sum_{p=1}^{N} (z_{\ell,p}^{\alpha})^4 - 3(\widehat{K}^{\alpha \alpha}_\ell)^2 \right)^2 = &\frac{1}{N^2} \Big[\sum_{p,p'} (z_{\ell,p}^{\alpha})^4 (z_{\ell,p'}^{\alpha})^4 - \frac{6}{N} \sum_{p,p',k} (z_{\ell,p}^{\alpha})^4 (z_{\ell,p'}^{\alpha})^2 (z_{\ell,k}^{\alpha})^2 \\ 
  &+ \frac{9}{N^2} \sum_{p,p',k,k'} (z_{\ell,p}^{\alpha})^2 (z_{\ell,p'}^{\alpha})^2 (z_{\ell,k}^{\alpha})^2 (z_{\ell,k'}^{\alpha})^2 \Big].
  \end{split}
\end{equation}
These three summations can be evaluated via successive Wick contractions:
\begin{equation}
\begin{aligned}
&\E\left[\sum_{p,p'} (z_{\ell,p}^{\alpha})^4 (z_{\ell,p'}^{\alpha})^4\right] = (K^{\alpha \alpha}_\ell)^4 (9N^2 + 96N), \\
&\E\left[\frac{6}{N} \sum_{p,p',k} (z_{\ell,p}^{\alpha})^4 (z_{\ell,p'}^{\alpha})^2 (z_{\ell,k}^{\alpha})^2\right] = (K^{\alpha \alpha}_\ell)^4 (18N^2 + 180N + 432), \\
&\E\left[\frac{9}{N^2} \sum_{p,p',k,k'} (z_{\ell,p}^{\alpha})^2 (z_{\ell,p'}^{\alpha})^2 (z_{\ell,k}^{\alpha})^2 (z_{\ell,k'}^{\alpha})^2\right] = (K^{\alpha \alpha}_\ell)^4 \left(9N^2 + 108N + 396 + \frac{432}{N}\right).
\end{aligned}
\end{equation}
Combining these results with the bias term \eqref{eq:bias-wick}, we obtain, to leading order in $1/N$:
\begin{equation}
\Var\left[\varepsilon_{\text{Wick}}^1\right] = \frac{24}{N}(K^{\alpha \alpha}_\ell)^4 + \mathcal{O}\left(\frac{1}{N^2}\right).
\end{equation}
\end{itemize}
\paragraph{Study of $\varepsilon_{\text{Weight}}$}
\label{par:eps_weight}
In contrast to $\varepsilon_{\text{Wick}}$, we show that, while this perturbation is unbiased, its variance is more ill-conditioned.

\begin{itemize}
  \item \textit{Bias study} -- 
The perturbation can be expanded as:
\begin{equation}
\frac{1}{N}\phi(\rvz^{\alpha})^\top \left( \frac{1}{N}\rmW_{\ell+1}^\top \rmW_{\ell+1} - \sigma_w^2\mathbf{I}\right)\phi(\rvz^{\alpha}) = \frac{1}{N}\sum_{i,j} \phi(\rvz^{\alpha}_{\ell,i})^\top \rmA_{ij}\phi(\rvz^{\alpha}_{\ell,j}),
\end{equation}
where
\begin{equation}
\rmA_{ij} = \left( \frac{1}{N}\rmW_{\ell+1}^\top \rmW_{\ell+1} - \sigma_w^2\mathbf{I}\right)_{ij} = \frac{1}{N}\sum_k \rmW_{\ell+1,ki}\rmW_{\ell+1,kj} - \delta_{ij} \sigma_w^2.
\end{equation}
The expectation of the perturbation is:
\begin{equation}
\E[\varepsilon_{\text{Weight}}] = \frac{1}{N}\sum_{i,j} \E[\phi(\rvz^{\alpha}_{\ell,i}) \phi(\rvz^{\alpha}_{\ell,j})] \E[\rmA_{ij}].
\end{equation}
Since $\E[\rmA_{ij}] = 0$, we have $\E[\varepsilon_{\text{Weight}}] = 0$, confirming that the perturbation is unbiased.
 \item \textit{Variance study} -- 
The variance is given by:
\begin{equation}
\Var[\varepsilon_{\text{Weight}}] = \frac{1}{N^2}\sum_{p,p',k,k'} \E\left[\phi(\rvz^{\alpha}_{\ell,p}) \phi(\rvz^{\alpha}_{\ell,p'}) \phi(\rvz^{\alpha}_{\ell,k}) \phi(\rvz^{\alpha}_{\ell,k'}) \right] \E[\rmA_{pp'} \rmA_{kk'}].
\end{equation}
At leading order in $K^{\alpha \alpha}_\ell$, a Taylor expansion of $\phi$ yields:
\begin{equation} \label{eq:nonlinearity-weight-defect}
\E\left[\phi(\rvz^{\alpha}_{\ell,p}) \phi(\rvz^{\alpha}_{\ell,p'}) \phi(\rvz^{\alpha}_{\ell,k}) \phi(\rvz^{\alpha}_{\ell,k'}) \right] = \phi_1^4 (K^{\alpha \alpha}_\ell)^2\left(\delta_{kk'}\delta_{p p'} + \delta_{kp}\delta_{k'p'} + \delta_{k'p}\delta_{kp'}\right) + \dots
\end{equation}
Assuming the weight matrix has finite fourth moments and introducing the fourth cumulant $\kappa_4$, we have:
\begin{equation} \label{eq:weight-defects}
\E[\rmA_{pp'} \rmA_{kk'}] = \frac{\sigma_w^4}{N} (\delta_{pk} \delta_{p'k'} + \delta_{p'k} \delta_{p k'}) + \frac{\kappa_4}{N} \delta_{pp'}\delta_{k k'} \delta_{pk}.
\end{equation}

Multiplying \eqref{eq:nonlinearity-weight-defect} and \eqref{eq:weight-defects}, we obtain, to leading order in $1/N$ and $K^{\alpha \alpha}_\ell$:
\begin{equation}
\Var[\varepsilon_{\text{Weight}}] = \frac{2 (\sigma_w \phi_1)^4}{N} (K^{\alpha \alpha}_\ell)^2.
\end{equation}

\end{itemize}

\subsection{Taxonomy of network effects at criticality \label{app:taxonomy-network}} 

In this section, we discuss in detail the flow of correlation at large depth for various input norms and correlations as explained in Sec.~\ref{sec:flow-correlation}. To this end, we restrict our study to six well-chosen pairs of inputs on two spheres of radii \( r_1 \) and \( r_2 \) (with \( r_1 < r_2 \)), denoted as \( S_1 \) and \( S_2 \), respectively. These pairs are defined below: 

\[
\begin{array}{l|l|l}
{\color{red!75!black}(x_1^\alpha,x_1^\beta)}
\in S_1 \times S_1,\;
x_1^\alpha \nparallel x_1^\beta
&
{\color{orange!90!black}(x_2^\alpha,x_2^\beta)}
\in S_2 \times S_2,\;
x_2^\alpha \nparallel x_2^\beta
&
{\color{green!50!black}(x_1^\gamma,x_1^\mu)}
\in S_1 \times S_1,\;
x_1^\gamma \parallel x_1^\mu
\\[1.2em]
{\color{teal!70!black}(x_2^\gamma,x_2^\mu)}
\in S_2 \times S_2,\;
x_2^\gamma \parallel x_2^\mu
&
{\color{blue!75!black}(x_3^\alpha,x_3^\beta)}
\in S_2 \times S_1,\;
x_3^\alpha \nparallel x_3^\beta
&
{\color{magenta!80!black}(x_3^\gamma,x_3^\mu)}
\in S_2 \times S_1,\;
x_3^\gamma \parallel x_3^\mu
\end{array}
\]

The first pair, \textcolor{red!75!black}{\((x_1^{\alpha}, x_1^{\beta})\)}, corresponds to inputs with reasonable norms that are initially not aligned. The resulting pair of pre-activations, \textcolor{red!75!black}{\((z_1^{\alpha}, z_1^{\beta})\)}, exhibits almost the same angle. The second pair, \textcolor{orange!90!black}{\((x_2^{\alpha}, x_2^{\beta})\)}, consists of large-norm inputs that are also initially not aligned. This yields a stronger misalignment for \textcolor{orange!90!black}{\((z_2^{\alpha}, z_2^{\beta})\)}, which becomes asymptotically decorrelated as \( r_2 \to \infty \). For the third pair, \textcolor{green!50!black}{\((x_1^{\gamma}, x_1^{\mu})\)}, the inputs are aligned vectors on the smaller sphere \( S_1 \). These remain aligned in the ambient space, as reflected in \textcolor{green!50!black}{\((z_1^{\gamma}, z_1^{\mu})\)}. The fourth pair, \textcolor{teal!70!black}{\((x_2^{\gamma}, x_2^{\mu})\)}, behaves identically to the third case, with alignment preserved in the ambient space. The fifth pair, \textcolor{blue!75!black}{\((x_3^{\alpha}, x_3^{\beta})\)}, involves two non-aligned vectors, one on each sphere. In the ambient space, \textcolor{blue!75!black}{\((z_3^{\alpha}, z_3^{\beta})\)} tends to decorrelate as \( r_2 \to \infty \), similar to the second pair. Most interestingly, for the final pair, \textcolor{magenta!80!black}{\((x_3^{\gamma}, x_3^{\mu})\)}, the inputs are aligned but lie on different spheres. Here, the dynamics of the pre-activations \textcolor{magenta!80!black}{\((z_3^{\gamma}, z_3^{\mu})\)} are more complex: as \( r_2 \to \infty \), the alignment is lost in the ambient space.

\begin{figure}[ht]
  \centering
  \includegraphics[width=\linewidth]{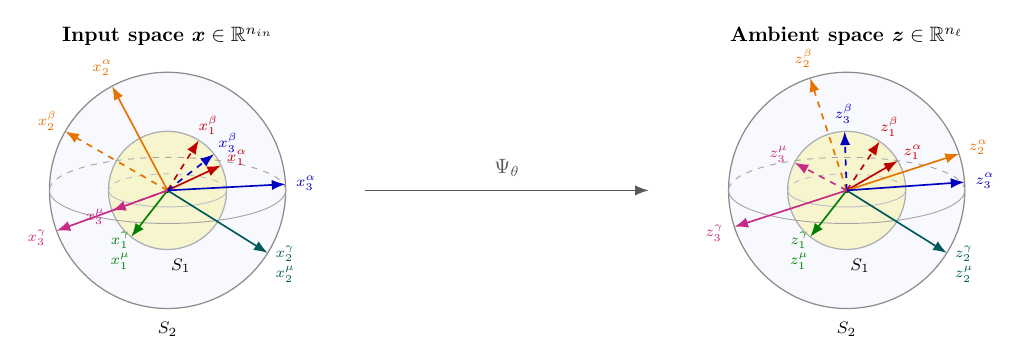}
  \caption{Taxonomy of the network's behaviour at criticality for different input pairs. This figure should be interpreted pair by pair, as the inter-pair dynamics cannot be represented in three dimensions.}
  \label{fig:sphere_taxonomy}
\end{figure}

\section{Jacobian \label{app:jac}}

\subsection{Asymptotic Freeness and its consequences on first and second moments of matrix products\label{app:freeness}}

This appendix derives equation~\eqref{eq:rel-var} using free probability. We begin by introducing the necessary definitions of \textit{free cumulants} and 
\textit{asymptotic freeness}, and establish key identities for mixed moments of 
products of free matrices such as $\prod_{\ell=1}^L \rmX_\ell$. Throughout, computing such moments reduces to 
evaluating the normalised trace, denoted
\begin{equation}
  \operatorname{tr}\!\left(\prod_{\ell=1}^L \rmX_\ell\right)
  :=\frac{1}{N} \operatorname{Tr}\!\left(\prod_{\ell=1}^L \rmX_\ell\right).
  \label{def:trace}
\end{equation}
We proceed in two steps: we first treat the simpler 
case of square matrix products, then extend the results to products of 
rectangular matrices of the form~(\ref{eq:defZApp}), which arise naturally 
in the expression of the end-to-end Jacobian.
We refer to \cite{mingo2018free, speicher2014freeprobabilityrandommatrices} 
for a complete introduction to free probability.\\

\noindent\textbf{Definition of free cumulants.} Let $\kappa_{\ell}(\rmX_1,\ldots,\rmX_\ell)$ denote the $\ell$-th order free 
cumulant of $(\rmX_1,\ldots,\rmX_\ell)$, where we assume $\rmX_{\ell} \in 
\mathbb{R}^{N \times N}$ for all $\ell$. The first two free cumulants are defined as:
\begin{align}
  \kappa_1(\rmX_1) &= \operatorname{tr}(X_1), \label{eq:cumul1}\\
  \kappa_2(\rmX_1,\rmX_2) &= \operatorname{tr}(X_1 X_2)
  - \operatorname{tr}(X_1)\operatorname{tr}(X_2). \label{eq:cumul2}
\end{align}
The first cumulant is the normalised trace, and the second cumulant is the 
free covariance. More generally, free cumulants can be defined recursively 
through the moment-cumulant relation, providing the relation between the normalised trace and free cumulants:
\begin{equation}
  \operatorname{tr}(\rmX_1 \cdots \rmX_L) = \sum_{\pi \in \mathrm{NC}(L)} 
  \prod_{\substack{B \in \pi \\ B = (i_1, \dots, i_b)}} 
  \kappa_{b}(\rmX_{i_1}, \dots, \rmX_{i_b}), \label{eq:freecumulants}
\end{equation}

where $\mathrm{NC}(L)$ denotes the set of \emph{non-crossing partitions} of $S=\{1, \dots, L\}$. A partition $\pi$ of $S$ is a collection of non-empty, mutually disjoint subsets of $S$, called \emph{blocks} and denoted $B$, whose union is $S$. For each block $B \in \pi$, we write $(i_1, \dots, i_b)$ for its elements listed in increasing order, with $b = |B|$ the block size; the contribution of $B$ to the product is $\kappa_b(\rmX_{i_1}, \dots, \rmX_{i_b})$.

A partition $\pi$ is called \emph{non-crossing} if its blocks do not interleave, meaning that there do not exist four elements $a_1 < a_2 < a_3 < a_4$ such that $a_1$ and $a_3$ belong to one block while $a_2$ and $a_4$ belong to another (see \cite{mingo2018free} for a formal definition). For example, consider the two partitions $\{\{1\}, \{2,4\}, \{3\}\}$ and 
$\{\{1,3\}, \{2,4\}\}$ of $\{1,2,3,4\}$, represented graphically as
{
\setlength{\unitlength}{0.35cm}
\[
\begin{picture}(5.5,3)
\thicklines
\put(0,1){\line(0,1){1}}
\put(3,1){\line(0,1){1}}
\put(1.5,0){\line(1,0){1.5}}
\put(1.5,0){\line(0,1){2}}
\put(3,0){\line(1,0){1.5}}
\put(4.5,0){\line(0,1){2}}
\put(0,2.7){\makebox(0,0){$1$}}
\put(1.5,2.7){\makebox(0,0){$2$}}
\put(3,2.7){\makebox(0,0){$3$}}
\put(4.5,2.7){\makebox(0,0){$4$}}
\end{picture}
\qquad
\begin{picture}(5.5,3)
\thicklines
\put(0,1){\line(1,0){3}}
\put(1.5,0){\line(1,0){3}}
\put(0,1){\line(0,1){1}}
\put(3,1){\line(0,1){1}}
\put(4.5,0){\line(0,1){2}}
\put(1.5,0){\line(0,1){2}}
\put(0,2.7){\makebox(0,0){$1$}}
\put(1.5,2.7){\makebox(0,0){$2$}}
\put(3,2.7){\makebox(0,0){$3$}}
\put(4.5,2.7){\makebox(0,0){$4$}}
\end{picture}
\]
}
We see from this representation that only the first partition is non-crossing and therefore belongs to $NC(4)$. 
The product of free cumulants associated with this particular partition is
\[ 
\prod_{\substack{B \in \{ \{1\}, \{2,4\}, \{3\} \} \\ B = (i_1, \dots, i_b)}} \kappa_{b}(\rmX_{i_1}, \dots, \rmX_{i_b}) = \kappa_1(\rmX_1) \kappa_2(\rmX_2, \rmX_4) \kappa_1(\rmX_3)
.\]

The relation~(\ref{eq:freecumulants}) can be inverted recursively to obtain the free cumulants $\kappa_L(\rmX_1,\ldots,\rmX_L)$. The procedure consists in subtracting from $\operatorname{tr}(\rmX_1 \cdots \rmX_L)$ the contributions of all partitions involving only cumulants of order strictly less than $L$. The first two cases, given above in \eqref{eq:cumul1} and \eqref{eq:cumul2}, follow directly from this procedure. \\

\noindent \textbf{Freeness property and its consequences.} Two matrices $\rmX_i$ and $\rmX_j$ are called \emph{free} if all mixed free cumulants $\kappa_b$ involving both $\rmX_i$ and $\rmX_j$ vanish exactly. In practice, exact freeness does not hold at finite $N$; one says instead that two random matrix ensembles are \emph{asymptotically free} if their mixed cumulants vanish in the large-$N$ limit \cite{mingo2018free}. The moment-cumulant relation~(\ref{eq:freecumulants}) serves as the definition of free cumulants for any $N$; the asymptotic freeness property is what makes the resulting simplifications practically useful.

For example, if $\rmX_1$ and $\rmX_2$ are free, then it implies $\kappa_2(\rmX_1, \rmX_2)=0$, and the normalised trace of their product can be factorised as the product of their normalised traces: 
\begin{align}
  \operatorname{tr}(\rmX_1 \rmX_2) 
  &= \kappa_2(\rmX_1, \rmX_2) + \kappa_1(\rmX_1)\kappa_1(\rmX_2) \\
  &= \operatorname{tr}(\rmX_1)\operatorname{tr}(\rmX_2). \label{eq:tr_ruleApp}
\end{align}
This property is analogous to the independence of scalar random variables, upon identifying the normalised trace with the expectation, and will be a key ingredient in establishing the first equality in \eqref{eq:rel-var}. 

We now derive a corresponding identity for second-order moments, again first in the square case, which will play the same role for the second equality in~(\ref{eq:rel-var}). Let us consider 
$\operatorname{tr}(\rmX_1 \rmX_2 \rmX_1 \rmX_2)$, assuming $\rmX_1$ and $\rmX_2$ are free. By the freeness property, only non-crossing partitions whose blocks contain indices corresponding to the same matrix contribute. There are only three such partitions, namely:
{
\setlength{\unitlength}{0.35cm}
\[
\begin{picture}(5.5,3)
\thicklines
\put(0,0){\line(0,1){2}}
\put(3,1){\line(0,1){1}}
\put(1.5,0){\line(1,0){1.5}}
\put(1.5,0){\line(0,1){2}}
\put(3,0){\line(1,0){1.5}}
\put(4.5,0){\line(0,1){2}}
\put(0,2.7){\makebox(0,0){$\rmX_1$}}
\put(1.5,2.7){\makebox(0,0){$\rmX_2$}}
\put(3,2.7){\makebox(0,0){$\rmX_1$}}
\put(4.5,2.7){\makebox(0,0){$\rmX_2$}}
\end{picture}
\qquad
\begin{picture}(5.5,3)
\thicklines
\put(0,0){\line(1,0){3}}
\put(0,0){\line(0,1){2}}
\put(3,0){\line(0,1){2}}
\put(4.5,0){\line(0,1){2}}
\put(1.5,1){\line(0,1){1}}
\put(0,2.7){\makebox(0,0){$\rmX_1$}}
\put(1.5,2.7){\makebox(0,0){$\rmX_2$}}
\put(3,2.7){\makebox(0,0){$\rmX_1$}}
\put(4.5,2.7){\makebox(0,0){$\rmX_2$}}
\end{picture}
\qquad
\begin{picture}(5.5,3)
\thicklines
\put(0,0){\line(0,1){2}}
\put(1.5,0){\line(0,1){2}}
\put(3,0){\line(0,1){2}}
\put(4.5,0){\line(0,1){2}}
\put(0,2.7){\makebox(0,0){$\rmX_1$}}
\put(1.5,2.7){\makebox(0,0){$\rmX_2$}}
\put(3,2.7){\makebox(0,0){$\rmX_1$}}
\put(4.5,2.7){\makebox(0,0){$\rmX_2$}}
\end{picture}
\]
}
This leads to 
\begin{equation}
  \operatorname{tr}\left(\left(\rmX_1 \rmX_2\right)^2\right) = \kappa_1^2(\rmX_1) \kappa_2(\rmX_2, \rmX_2) + \kappa_1^2(\rmX_2) \kappa_2(\rmX_1, \rmX_1) + \kappa_1^2(\rmX_1) \kappa_1^2(\rmX_2).\label{eq:interTrace}
\end{equation}
Using definitions \eqref{eq:cumul1} and \eqref{eq:cumul2}, recalling the relative variance definition 
\begin{equation}
  \mathbb{V}[\rmX] = \frac{\operatorname{tr}(\rmX^2)}
  {\bigl(\operatorname{tr}(\rmX)\bigr)^2} - 1,
  \label{eq:relVarTraceApp}
\end{equation}
equation~\eqref{eq:interTrace} gives 
the variance rule for free square matrix products:
\begin{equation}
    \mathbb{V}[\rmX_1 \rmX_2] =   \mathbb{V}[\rmX_1] +   \mathbb{V}[\rmX_2].\label{eq:VarSquareApp}
\end{equation}
That is, the relative variance of a product of free square matrices is the sum of their individual relative variances. \\

\noindent\textbf{Generalization to rectangular matrices. }
We now show how the results above extend to the more complicated matrix product considered in the main text, equation (\ref{eq:defZApp}), namely 
\begin{equation}
  \rmZ_L = \left(\prod_{\ell=1}^L \rmX_\ell\right)
  \left(\prod_{\ell=1}^L \rmY_\ell\right)^\top,
  \quad \rmX_\ell,\,\rmY_\ell \in \mathbb{R}^{n_\ell \times n_{\ell-1}},
  \label{eq:defZApp}
\end{equation}
where the matrices $\rmX_\ell \rmY_\ell^\top$ are assumed asymptotically 
free. 

We proceed recursively by computing the moments of $\rmX_L \rmZ_{L-1} \rmY_L^\top$, 
with $\rmX_L, \rmY_L \in \mathbb{R}^{n_L \times n_{L-1}}$ and $\rmZ_{L-1} \in \mathbb{R}^{n_{L-1} 
\times n_{L-1}}$, using freeness of $\rmZ_{L-1}$ and $\rmX_L \rmY_L^\top$. To simplify presentation, we drop in the following the index $L$, introduce $m:=n_{L-1}$, and introduce $\rmA:=\rmZ_{L-1}$.

For any 
$k \in \mathbb{N}$, the cyclic property of the trace and 
definition~(\ref{def:trace}) give
\begin{equation}
  \operatorname{tr}\!\left((\rmX \rmA \rmY^\top)^k\right) 
  = \frac{m}{n}\,\operatorname{tr}\!\left((\rmY^\top \rmX \rmA)^k\right),
  \label{eq:rec2square}
\end{equation}
which reduces the computation of moments of a rectangular matrix product to 
that of a mixed moment of two square matrices, allowing us to apply the 
machinery developed above.

For $k=1$, asymptotic freeness of $\rmX\rmY^\top$ and $\rmA$ yields
\begin{align}
  \operatorname{tr}(\rmX \rmA \rmY^\top) 
  &= \frac{m}{n}\,\operatorname{tr}(\rmY^\top \rmX)\,\operatorname{tr}(\rmA)  \label{eq:traceFinalApp}\\
  &= \operatorname{tr}(\rmX \rmY^\top)\,\operatorname{tr}(\rmA),
\end{align}
which generalizes the trace product rule to the rectangular case. 

For $k=2$, setting $\rmB = \rmY^\top \rmX$ and applying 
equation~(\ref{eq:freecumulants}) together with freeness of $\rmB$ and $\rmA$,
\begin{align}
  \operatorname{tr}\!\left((\rmX \rmA \rmY^\top)^2\right)
  &= \frac{m}{n}\,\operatorname{tr}(\rmB \rmA \rmB \rmA) \notag\\
  &= \frac{m}{n}\Bigl(
    \kappa_1^2(\rmB)\,\kappa_2(\rmA,\rmA)
    + \kappa_2(\rmB,\rmB)\,\kappa_1^2(\rmA)
    + \kappa_1^2(\rmB)\,\kappa_1^2(\rmA)
   \Bigr). \label{eq:lastTraceApp}
\end{align}
Using the relative 
variance defined in \eqref{eq:relVarTraceApp}, combined with \eqref{eq:lastTraceApp}, \eqref{eq:traceFinalApp}, and  \eqref{eq:cumul1} with \eqref{eq:cumul2}, gives 
the rectangular variance rule:
\begin{equation}
  \mathbb{V}[\rmX \rmA \rmY^\top]
  = \frac{n}{m}\Bigl(\mathbb{V}[ \rmY^\top \rmX] + \mathbb{V}[\rmA]\Bigr) 
  + \frac{n}{m} - 1. \label{eq:varFinalApp}
\end{equation}
%
The previous relations applied to the case (\ref{eq:defZApp}) give: 

\begin{equation}
  \operatorname{tr}(\rmZ_{L}) = \operatorname{tr}( \rmX_L \rmY_L^\top)\,\operatorname{tr}(\rmZ_{L-1}),
\end{equation} 
and
  \begin{equation}
  \mathbb{V}[\rmZ_{L}]
  = \frac{n_{L}}{n_{L-1}}\Bigl(\mathbb{V}[\rmY_L^\top\rmX_L ] + \mathbb{V}[\rmZ_{L-1}]\Bigr) 
  + \frac{n_{L}}{n_{L-1}} - 1. 
\end{equation}
These recurrence equations lead to the expected result, equation~(\ref{eq:rel-var}) of the main text.

\subsection{Gaussian weights moments} \label{sec:appendix-gaussian-weight}
 In this section, we derive the first spectral moment of $\frac{1}{n_{\ell-1}}\rmW_{\ell}^\top \rmW_{\ell}$ and its relative variance $\mathbb{V}[\frac{1}{n_{\ell-1}}\rmW_{\ell}^\top \rmW_{\ell}]$. From the Marchenko-Pastur law:
\begin{equation}
  m^{(1)}_{\frac{1}{n_{\ell-1}} \rmW_\ell \rmW_\ell^\top} = \sigma_w^2 \quad \mbox{and} \quad \mathbb{V}[\frac{1}{n_{\ell-1}}\rmW_{\ell} \rmW_{\ell}^\top] = \frac{n_\ell}{n_{\ell-1}}.
\end{equation}
Following similar reasoning to that in the previous section, we have:
\begin{equation}
  m^{(k)}_{\frac{1}{n_{\ell-1}} \rmW_\ell \rmW_\ell^\top} = \frac{n_{\ell - 1}}{n_{\ell}} m^{(k)}_{\frac{1}{n_{\ell-1}} \rmW_\ell^\top \rmW_\ell} ,
\end{equation}
leading to
\begin{equation}
  m^{(1)}_{\frac{1}{n_{\ell-1}} \rmW_\ell^\top \rmW_\ell} = \sigma_w^2\frac{n_\ell}{n_{\ell-1}}.
\end{equation}
For the relative variance we get the result almost immediately from \eqref{eq:varFinalApp}: 
\begin{align}\label{eq:rel-variance-weight}
  \mathbb{V}[\frac{1}{n_{\ell-1}}\rmW_{\ell}^\top \rmW_{\ell}] &= \frac{n_{\ell-1 }}{n_{\ell}}(\mathbb{V}[\frac{1}{n_{\ell-1}} \rmW_\ell \rmW_\ell^\top] + 1) - 1\\
  &= \frac{n_{\ell-1 }}{n_{\ell}}.
\end{align}

\subsection{Convergence of $\sum_{\ell = 1}^L \mathbb{V}[\rmD_{\ell}^{\alpha}\rmD_{\ell}^{\beta}]$} \label{sec:appendix-sum-convergence}

For this, it is sufficient to recall the definition \eqref{eq:rel-variance-def} which yields : 

\[\sum_{\ell = 1}^L \mathbb{V}[\rmD_{\ell}^{\alpha}\rmD_{\ell}^{\beta}]=\sum_{\ell=1}^L \left(\frac{m_{\rmD_{\ell}^{\alpha}\rmD_{\ell}^{\beta}}^{(2)}}{\left(m_{\rmD_{\ell}^{\alpha}\rmD_{\ell}^{\beta}}^{(1)}\right)^2} - 1\right)\]
For $\phi_2=0$, by Taylor expanding at criticality we get
\begin{align}
  m_{\rmD_{\ell}^{\alpha}\rmD_{\ell}^{\beta}}^{(1)} & = \phi_1^2\left[1 + \frac{\phi_{3}}{\phi_{1}} \tfrac{1}{2}\left( K^{\alpha \alpha}_\ell + K^{\beta \beta}_{\ell} \right)+ \mathcal{O}(K_{\ell}^2)\right] , \\
  m_{\rmD_{\ell}^{\alpha}\rmD_{\ell}^{\beta}}^{(2)} & =  \phi_1^4\left[1 + \frac{2\phi_{3}}{\phi_{1}} \tfrac{1}{2}\left( K^{\alpha \alpha}_\ell + K^{\beta \beta}_{\ell} \right)+ \mathcal{O}(K_{\ell}^2 )\right],
  \end{align}
  where the absence of superscripts in \(K_{\ell}\) denotes all possible superscripts. This gives the scaling \[\left( {m_{\rmD_{\ell}^{\alpha}\rmD_{\ell}^{\beta}}^{(2)}}/{(m_{\rmD_{\ell}^{\alpha}\rmD_{\ell}^{\beta}}^{(1)})^2}-1\right) = \mathcal{O}(\ell^{-2}),\] 
  giving us the convergence of the studied sum.

\subsection{Concentration for fixed $n_\textrm{out}$ \label{app:fixed_nout}}
In this section we  prove that in the most simple case $n_\textrm{out}$ finite and $n_\textrm{in}, N \to \infty$, the spectral standard deviation of the overlap end-to-end Jacobian vanishes:
\begin{equation}
   \mathbb{S}[\mathcal{J}_L^\alpha \mathcal{J}_L^{\beta\top}] = 0 , \label{eq:appResToBeProved_S_Jac}
\end{equation}

To do so, let us consider a matrix element $\left(\mathcal{J}_L^\alpha \mathcal{J}_L^{\beta\top}\right)_{ij}$ 
\begin{equation}
  \left(\mathcal{J}_L^\alpha \mathcal{J}_L^{\beta\top}\right)_{ij} = \frac{1}{n_{L-1}}\sum_{k,k'} (\rmW_L)_{ik}(\rmW_L^\top)_{k'j} \rmC^{\alpha \beta}_{kk'},
\end{equation}
where $\rmC^{\alpha \beta}$ contains all remaining terms in the Jacobian and we saw in the previous section that it admits a finite second spectral moment $m^{(2)}_{\rmC^{\alpha \beta}} < \infty$. Recalling that we have $\mathbb{E}[\rmW_{ik} \rmW_{jk'}] = \delta_{ij}\delta_{kk'}\sigma_w^2$ and $\rmW_L$ independent of $\rmC^{\alpha \beta}$ we have directly: 

\begin{align}
  \mathbb{E}\left(\mathcal{J}_L^\alpha \mathcal{J}_L^{\beta\top}\right)_{ij} &= \delta_{ij}\frac{\sigma_w^2}{n_{L-1}}\sum_{k} \mathbb{E}[\rmC^{\alpha \beta}_{kk}]\\
  &= \delta_{ij}\sigma_w^2 m^{(1)}_{\rmC^{\alpha \beta}}.
\end{align}

Now we can move on to the calculation of the variance of $ \left(\mathcal{J}_L^\alpha \mathcal{J}_L^{\beta\top}\right)_{ij}$ to see that it concentrates around its average. First let's introduce: 

\begin{equation}
  \left(\mathcal{J}_L^\alpha \mathcal{J}_L^{\beta\top}\right)_{ij}^2 = \frac{1}{n_{L-1}^2}\sum_{k,k',p,p'} (\rmW_L)_{ik}(\rmW_L^\top)_{k'j} (\rmW_L)_{ip}(\rmW_L^\top)_{p'j}\rmC^{\alpha \beta}_{pp'}\rmC^{\alpha \beta}_{kk'} \label{eq:sec-moment-entry-jac}
\end{equation}

\paragraph{In the case $i \neq j$}
In that specific case we have by independence an equation that greatly simplifies the calculations

\begin{align}
\mathbb{E}[(\rmW_L)_{ik}(\rmW_L)_{jk'} (\rmW_L)_{ip}(\rmW_L)_{jp'}] &= \mathbb{E}[(\rmW_L)_{ik}(\rmW_L)_{ip}] \,\mathbb{E}[ (\rmW_L)_{jk'} (\rmW_L)_{jp'}] \\
&= \delta_{kp} \delta_{k'p'} \sigma_w^4
\end{align}

This leads to: 

\begin{align}
 \mathbb{E}\left[\left(\mathcal{J}_L^\alpha \mathcal{J}_L^{\beta\top}\right)_{ij}^2\right] &= \frac{\sigma_w^4}{n_{L-1}^2}\sum_{k,p} \mathbb{E}[(\rmC^{\alpha \beta}_{kp})^2] \\
&= \frac{\sigma_w^4}{n_{L-1}^2} \mathbb{E}\Tr{\rmC^{\alpha \beta} \rmC^{\alpha \beta \top}}.
\end{align}

Furthermore, since $C^{\alpha \beta} = A^{\alpha} A^{\beta}$ by Cauchy-Schwarz we have directly: 

\begin{align}
\frac{1}{n_{L-1}} \mathbb{E}\Tr{\rmC^{\alpha \beta} \rmC^{\alpha \beta \top}} &\leq \sqrt{\frac{1}{n_{L-1}} \mathbb{E}\Tr{(\rmC^{\alpha \alpha})^2}}\sqrt{\frac{1}{n_{L-1}} \mathbb{E}\Tr{(\rmC^{\beta \beta})^2}}\\
&\leq \sqrt{m^{(2)}_{\rmC^{\alpha \alpha}}} \sqrt{m^{(2)}_{\rmC^{\beta \beta}}}.
\end{align}
Using that $m^{(2)}_{\rmC^{\alpha \alpha}}$ is finite, we get:
\begin{equation}
  \Var \left(\mathcal{J}_L^\alpha \mathcal{J}_L^{\beta\top}\right)_{ij} = \mathbb{E}\left[\left(\mathcal{J}_L^\alpha \mathcal{J}_L^{\beta\top}\right)_{ij}^2\right] = \mathcal{O}(\frac{1}{n_{L-1}}).
\end{equation}
\paragraph{Case $i = j$} The derivation is more involved. It is useful to note first that a weight matrix having finite fourth moment $\forall i,j, \, \mu_k = \mathbb{E}[\rmW_{ij}^k] < \infty$ satisfies: 
\begin{equation}
  \begin{split}
  \mathbb{E}[(\rmW_L)_{ik}(\rmW_L)_{ik'} (\rmW_L)_{ip}(\rmW_L)_{ip'}] = &\sigma_w^4 \left(\delta_{kk'}\delta_{p p'} + \delta_{kp}\delta_{k'p'} + \delta_{k'p}\delta_{kp'}\right) \\ &+ \kappa_4 \delta_{kk'}\delta_{kp}\delta_{kp'}
  \end{split}
\end{equation}
with $\kappa_4 = \mu_4 - 3\sigma_w^4$ the fourth cumulant. Considering \eqref{eq:sec-moment-entry-jac} we have: 
\begin{equation}
  \begin{split}
  \E \left[\left(\mathcal{J}_L^\alpha \mathcal{J}_L^{\beta\top}\right)_{ii}^2\right] &= \frac{\sigma_w^4}{n_{L-1}^2} \left( \sum_{k,p}\mathbb{E}[\rmC^{\alpha \beta}_{pp} \rmC^{\alpha \beta}_{kk}] + \sum_{k,k'}\mathbb{E}[(\rmC^{\alpha \beta}_{kk'})^2] + \sum_{k,k'}\mathbb{E}[\rmC^{\alpha \beta}_{kk'}\rmC^{\alpha \beta}_{k'k}] \right) \\
  &\phantom{\quad} + \frac{\kappa_4}{n_{L-1}^2} \sum_{k} \mathbb{E}[(\rmC_{kk}^{\alpha\beta})^2] \\
  &= \frac{\sigma_w^4}{n_{L-1}^2} \left( \mathbb{E}[(\operatorname{Tr} \rmC^{\alpha \beta})^2] + \mathbb{E}[\operatorname{Tr} \rmC^{\alpha \beta}\rmC^{\alpha \beta\top}]+ \mathbb{E}[\operatorname{Tr} (\rmC^{\alpha \beta})^2]\right) \\
  &\phantom{\quad} + \frac{\kappa_4}{n_{L-1}^2} \sum_{k} \mathbb{E}[(\rmC_{kk}^{\alpha\beta})^2]. \\
  \end{split}
  \label{eq:second-diagonal-moment-jac}
\end{equation}
Furthermore, we notice that: 
\begin{equation}
  \sum_{k} \mathbb{E}[(\rmC_{kk}^{\alpha\beta})^2] \leq \sum_{k,p} \mathbb{E}[(\rmC^{\alpha \beta}_{kp})^2].
\end{equation}
Thus, having already proved the vanishing of $\frac{\sigma_w^4}{n_{L-1}^2}\sum_{k,p} \mathbb{E}[(\rmC^{\alpha \beta}_{kp})^2] $, we deduce that the second and last terms in \eqref{eq:second-diagonal-moment-jac} vanish as well. Hence, we end up with: 
\begin{equation}
   \mathbb{E}\left[\left(\mathcal{J}_L^\alpha \mathcal{J}_L^{\beta\top}\right)_{ii}^2\right] = \frac{\sigma_w^4}{n_{L-1}^2} \left( \mathbb{E}[(\operatorname{Tr} \rmC^{\alpha \beta})^2] + \mathbb{E}[\operatorname{Tr} (\rmC^{\alpha \beta})^2]\right) .
\end{equation}
The last term in this equation vanishes because $m^{(2)}_{\rmC^{\alpha \beta}}$ is finite. Hence, the total variance can be expressed as: 
\begin{align}
   \Var \left[\left(\mathcal{J}_L^\alpha \mathcal{J}_L^{\beta\top}\right)_{ii}^2\right] &= \sigma_w^4 \left( \mathbb{E}\left[\left(\tfrac{1}{n_{L-1}}\operatorname{Tr} \rmC^{\alpha \beta}\right)^2\right] - \left(\mathbb{E}\left[\tfrac{1}{n_{L-1}}\operatorname{Tr} \rmC^{\alpha \beta}\right]\right)^2\right) \\
   &= \Var\left[\tfrac{1}{n_{L-1}}\operatorname{Tr} \rmC^{\alpha \beta}\right] \xrightarrow{n_{L-1} \to \infty} 0.
\end{align}
To conclude, since $\mathcal{J}_L^\alpha \mathcal{J}_L^{\beta\top} \in \mathbb{R}^{n_\textrm{out} \times n_\textrm{out}}$ is of finite size with $n_{\mathrm{out}} = O(1)$, having proven that each entry of the matrix vanishes as $N \to \infty$ implies \eqref{eq:appResToBeProved_S_Jac} in this limit.

\section{Neural Tangent Kernel \label{app:ntk}}

\subsection{Expression of the NTK operator}

Using the definition of the NTK (\ref{eq:NTK-def}) and the relation $\rvz_\ell^\alpha=\frac{1}{\sqrt{n_{\ell - 1}}}\rmW_\ell \phi\left(\rvz_{\ell-1}^\alpha\right)$ (there is no bias at criticality), we get:
\begin{align}
  \widehat{ \Theta}_{\ell}^{\alpha\beta} & = \frac{\partial \rvz_{\ell}^{\alpha}}{\partial \theta_{\ell}} \left(\frac{\partial \rvz_{\ell}^{\beta}}{\partial \theta_{\ell}}\right)^{\top} +  \rmJ_{\ell}^{\alpha} \widehat{ \Theta}_{\ell - 1}^{\alpha\beta}  \rmJ_{\ell}^{\beta\top} \\
   & = \widehat{\Phi}^{\alpha\beta}_{\ell} + \sum_{i = 1}^{\ell-1} {\widehat\Phi^{\alpha \beta}_{i}} \prod_{k = 0}^{\ell-i-1} \rmJ_{\ell-k}^{\alpha} \left( \prod_{k = 0}^{\ell-i-1} \rmJ_{\ell-k}^{\beta}\right)^{\top}
\end{align}
where $\rmJ_\ell(\cdot)$ is the layer-wise Jacobian matrix introduced equation (\ref{eq:ef_jac}) and \begin{equation}
 \quad \widehat\Phi^{\alpha \beta}_{\ell} = \frac{1}{N} \phi(\rvz_{\ell-1}^{\alpha})^{\top} \phi(\rvz_{\ell-1}^{\beta}) \quad \text{with} \quad \widehat\Phi^{\alpha \beta}_{1} = \frac{1}{N}\rvx^{\alpha\top}\rvx^{\beta}
\end{equation} the empirical kernel of the post-activation at layer \(\ell\). 
This empirical kernel is related to the empirical Covariance kernel through 
\begin{equation}
 \text{\textbf{Gaussian}: } \E[\widehat{K}^{\alpha \beta}_{\ell}] = \sigma_{w}^2 \E[\widehat{\Phi}^{\alpha \beta}_{\ell}]\qquad
   \text{\textbf{Orthogonal}: }\widehat{K}^{\alpha \beta}_{\ell} = \sigma_{w}^2 \widehat{\Phi}^{\alpha \beta}_{\ell}.
\end{equation}
In the large $N$ limit, owing to the concentration of $\widehat{\Phi}^{\alpha \beta}_\ell$ to its averaged value, we have
\begin{equation}
   \quad \lim_{N\rightarrow\infty}\widehat\Phi^{\alpha \beta}_{\ell} = \frac{K^{\alpha\beta}_\ell}{\sigma_w^2} .
\end{equation}

\noindent This leads directly to the expected formula (\ref{eq:ntk_compact}).

\subsection{Second Moment of the NTK}\label{sec:second-moment-ntk-derivation}

Let's first start by expanding the second moment as follows:

\begin{align}
  m^{(2)}_{ \Theta_{L}^{\alpha\beta}} & = \frac{1}{n_\textrm{out}} \Tr \E\left[( \Theta_{L}^{\alpha\beta})^2 \right] \\                             
  & = \frac{1}{(AL)^2} \frac{1}{\sigma_w^4} \sum_{\ell = 0}^L\sum_{\ell' = 0}^L \rho^{\alpha\beta}_\ell \rho^{\alpha\beta}_{\ell'} \frac{1}{n_\textrm{out}} \Tr \E \left[ {\rmP}^{\alpha \beta}_{L\to \ell} {\rmP}^{\alpha \beta}_{L\to \ell'} \right].
\end{align}

Then, using the definition of ${\rmP}^{\alpha \beta}_{L\to \ell}$: 
\begin{align}
 \frac{1}{n_\textrm{out}} \Tr \E \left[ {\rmP}^{\alpha \beta}_{L\to \ell} {\rmP}^{\alpha \beta}_{L\to \ell'} \right] 
 & = F^L_{\ell, \ell'} \frac{1}{n_\textrm{out}} \Tr \E\left[ 
 \left(\rmJ_{L \to \ell'}^{\beta}\right)^{\top} 
 \rmJ_{L \to \ell'}^{\alpha} 
 \rmJ_{\ell'\to \ell}^{\alpha} 
 \left(\rmJ_{\ell'\to \ell}^{\beta}\right)^{\top} 
 \left(\rmJ_{L \to \ell'}^{\beta}\right)^{\top} 
 \rmJ_{L \to \ell'}^{\alpha}
 \right] \\
  & = F^L_{\ell, \ell'} \frac{1}{n_\textrm{out}} \Tr \E\left[ 
  \left(\rmJ_{L \to \ell'}^{\beta}\right)^{\top} 
  \rmJ_{L \to \ell'}^{\alpha} 
  \left(\rmJ_{L \to \ell'}^{\beta}\right)^{\top} 
  \rmJ_{L \to \ell'}^{\alpha} 
  \rmJ_{\ell'\to \ell}^{\alpha} 
  \left(\rmJ_{\ell'\to \ell}^{\beta}\right)^{\top} 
  \right] \\
  & = F^L_{\ell, \ell'} \frac{1}{n_\textrm{out}} \Tr \E\left[ 
  \left(\rmG_{\ell'}^{\alpha\beta}\right)^2 
  \rmH_{\ell,\ell'}^{\alpha \beta}
  \right],
\end{align}
with
\[
 F^L_{\ell, \ell'}:= (A L)^2 
 \sqrt{ K^{\alpha \alpha}_\ell K^{\alpha \alpha}_{\ell'} 
 K^{\beta \beta}_\ell K^{\beta \beta}_{\ell'}} 
 \quad 
  \rmG_{\ell'}^{\alpha\beta}
  \;:=\;
  \left(\rmJ_{L \to \ell'}^{\beta}\right)^{\top}
  \rmJ_{L \to \ell'}^{\alpha}
  \quad \text{and} \quad
  \rmH_{\ell,\ell'}^{\alpha \beta}
  \;:=\;
  \rmJ_{\ell'\to \ell}^{\alpha} 
  \left(\rmJ_{\ell'\to \ell}^{\beta}\right)^{\top}.
\]
In the limit of infinite width 
\(\rmG_{\ell'}^{\alpha\beta}\) and 
\(\rmH_{\ell,\ell'}^{\alpha\beta} \) are freely independent, which allows us to write the first moment of their product as the product of their first moments:
\begin{align}
  \frac{1}{n_{\ell'}} \Tr \E\left[
  \left(\rmG_{\ell'}^{\alpha\beta}\right)^2
  \rmH_{\ell,\ell'}^{\alpha \beta}
  \right]
  &=
  \frac{1}{n_{\ell'}} \Tr \E\left[
  \left(\widetilde{\rmG}_{\ell'}^{\alpha\beta}\right)^2
  \right]
  \frac{1}{n_{\ell'}} \Tr \E\left[
  \rmH_{\ell,\ell'}^{\alpha \beta}
  \right] \\
  &= n_\textrm{out} 
  m^{(2)}_{\widetilde{\rmG}_{\ell'}^{\alpha\beta}} 
  m^{(1)}_{\rmH_{\ell,\ell'}^{\alpha \beta}} \label{eq:ntk_gh}.
\end{align}
In the second equality, we introduced
\begin{equation}
  \widetilde{\rmG}_{\ell'}^{\alpha\beta} 
  := 
  \rmJ_{L\to \ell'}^{\alpha}
  (\rmJ_{L \to \ell'}^{\beta})^\top,
\end{equation}
which is a matrix of size \(n_{\text{out}}\times n_{\text{out}}\). One then needs to compute the second moment of this matrix. Note that we already computed the second moment of 
\(\widetilde{\rmG}^{\alpha\beta}_0=\mathcal{J}_L^{\alpha}\mathcal{J}_L^{\beta\top}\) 
in Sec.~\ref{sec:secondmoment}. The very same strategy can directly be applied to the computation of 
\(\widetilde{\rmG}^{\alpha\beta}_{\ell'}\), and this leads to 
\begin{align} \label{eq:rel-variance-Gtilde}
    m^{(1)}_{\widetilde{\rmG}_{\ell'}^{\alpha\beta}} 
    &= \prod_{\ell = \ell'+1}^L \chi_\ell^\perp = \frac{\Xi_L^{\alpha\beta}}{\Xi_{\ell'}^{\alpha\beta}} \\
    \mathbb{V}\left[ \widetilde{\rmG}_{\ell'}^{\alpha\beta} \right] 
    &= \frac{n_{\text{out}}}{n_{\ell'}} - 1 
    + \sum_{r = \ell'+1}^L 
    \frac{n_{\text{out}}}{n_{r - 1}} 
    \left(
    \mathbb{V}\left[\frac{1}{n_{r-1}}\rmW_{r}^\top \rmW_{r}\right] 
    + \mathbb{V}\left[\rmD_{r}^{\alpha}\rmD_{r}^{\beta}\right]
    \right).
\end{align}
Then applying again the same logic used in the Jacobian study, we can distinguish the Gaussian weight case from the orthogonal one: 

\begin{itemize}
  \item \textit{Gaussian Case:} Combining \eqref{eq:rel-variance-weight} and \eqref{eq:rel-variance-Gtilde}, we obtain:
\begin{equation}
  \mathbb{V}\left[\widetilde{\rmG}_{\ell'}^{\alpha\beta}\right] 
  = 
  \sum_{r = \ell'}^{L-1} \frac{n_{\text{out}}}{n_r} 
  + 
  \sum_{r = \ell'+1}^{L} \frac{n_{\text{out}}}{n_{r - 1}} 
  \mathbb{V}\left[\rmD_{r}^{\alpha}\rmD_{r}^{\beta}\right].
  \label{eq:VtildeG_gaussian}
\end{equation}
We introduce the variance residual 
\begin{equation}
R^L_{\ell'}:=
\sum_{r = \ell'+1}^{L} 
\mathbb{V}\left[\rmD_{r}^{\alpha}\rmD_{r}^{\beta}\right].
\label{eq:R_L_ellprime}
\end{equation}
For \(1<\ell'<L\), equation \ref{eq:VtildeG_gaussian} can be simplified as 
\begin{equation} 
\label{eq:rel-var-gaussian-G}
\mathbb{V}\left[\widetilde{\rmG}_{\ell'}^{\alpha\beta}\right]
= 
\frac{n_\textrm{out}}{N}
\left( L - \ell' + R^L_{\ell'}\right).
\end{equation}
We already saw in Appendix~\ref{sec:appendix-sum-convergence} that the summation is convergent in the previous equation. The second moment is deduced from the knowledge of the relative variance and first moment:
\begin{equation}
  m^{(2)}_{\widetilde{\rmG}_{\ell'}^{\alpha\beta}} 
  = 
  \left(
  \frac{ \Xi_{\ell}^{\alpha \beta}}
  {\Xi^{\alpha \beta}_{\ell'}}
  \right)^2 
  \left(
  1 + \mathbb{V}\left[\widetilde{\rmG}_{\ell'}^{\alpha\beta}\right] 
  \right).
\end{equation}
Now we focus on the simpler case of \(\rmH_{\ell,\ell'}^{\alpha \beta}\), whose first moment is directly given by: 
\begin{equation}
  m^{(1)}_{\rmH_{\ell,\ell'}^{\alpha \beta}} 
  = 
  \frac{\Xi^{\alpha \beta}_{\ell'}}
  {\Xi^{\alpha \beta}_{\ell}}.
\end{equation}

Then combining \eqref{eq:ntk_gh} with the two previously calculated moments for \(\ell'>\ell\), we have:
\begin{equation}
  \frac{1}{n_{\ell'}} \Tr \E\left[
  \left(\rmG_{\ell'}^{\alpha\beta}\right)^2 
  \rmH_{\ell,\ell'}^{\alpha \beta}
  \right] 
  = 
  n_\textrm{out}
  \frac{(\Xi^{\alpha \beta}_{\ell})^2}
  {\Xi^{\alpha \beta}_{\ell'} \Xi^{\alpha \beta}_{\ell}} 
  \left(
  1 + \mathbb{V}\left[\widetilde{\rmG}_{\ell'}^{\alpha\beta}\right]
  \right),
\end{equation}
and now we can write the full second moment sum using \eqref{eq:second-moment-ntk}:

\begin{equation}
  \begin{split}
    m^{(2)}_{ \Theta_{L}^{\alpha\beta}} &= \left(\frac{1}{\sigma_{w}^2 AL}\right)^2\sum_{\ell = 0}^{L} \sum_{\ell'> \ell}^{L} \rho^{\alpha \beta}_\ell \rho^{\alpha \beta}_{\ell'} F^L_{\ell, \ell'}\frac{(\Xi^{\alpha \beta}_{L})^2}{\Xi^{\alpha \beta}_{\ell'} \Xi^{\alpha \beta}_{\ell}} \left(1 + \mathbb{V}\left[\widetilde{\rmG}_{\ell'}^{\alpha\beta}\right] \right)  \\
    & + \left(\frac{1}{\sigma_{w}^2 AL}\right)^2\sum_{\ell = 0}^{L} \sum_{\ell' < \ell}^{L} \rho^{\alpha \beta}_\ell \rho^{\alpha \beta}_{\ell'} F^L_{\ell, \ell'}\frac{(\Xi^{\alpha \beta}_{L})^2}{\Xi^{\alpha \beta}_{\ell'} \Xi^{\alpha \beta}_{\ell}} \left(1 + \mathbb{V}\left[\widetilde{\rmG}_{\ell'}^{\alpha\beta}\right] \right) \\
    & +  \left(\frac{1}{\sigma_{w}^2 AL}\right)^2\sum_{\ell=0}^{L}  (\rho^{\alpha \beta}_\ell)^2 F^L_{\ell, \ell}\frac{(\Xi^{\alpha \beta}_{L})^2}{(\Xi^{\alpha \beta}_{\ell})^2} \left(1 + \mathbb{V}\left[\widetilde{\rmG}_{\ell}^{\alpha\beta}\right] \right).
  \end{split}
\end{equation}

We note  that the two first sum are exactly the same under the permutation of \(\ell,\ell'\). In addition,  the third sum is  negligible with respect to the other ones: since there is only L terms of same order of magnitude than the off-diagonal term, we expect the ratio between the third and the first or second sum to behave like $O(1 / L)$. This lead to a simplified expression:
\begin{equation} \label{eq:moment-2-truncate}
    m^{(2)}_{ \Theta_{L}(x_{\alpha}, x_{\beta})} = 2\left(\frac{1}{\sigma_{w}^2 AL}\right)^2\sum_{\ell = 0}^{L} \sum_{\ell'> \ell}^{L} \rho^{\alpha \beta}_\ell \rho^{\alpha \beta}_{\ell'} F^L_{\ell, \ell'}\frac{(\Xi^{\alpha \beta}_{L})^2}{\Xi^{\alpha \beta}_{\ell'} \Xi^{\alpha \beta}_{\ell}} \left(1 + \mathbb{V}\left[\widetilde{\rmG}_{\ell'}^{\alpha\beta}\right] \right).
\end{equation}

Furthermore, as highlighted for the first moment calculation we get: 

\begin{equation}
   F^L_{\ell, \ell'}\frac{(\Xi^{\alpha \beta}_{L})^2}{\Xi^{\alpha \beta}_{\ell'} \Xi^{\alpha \beta}_{\ell}} \xrightarrow{\ell, \ell', L \to \infty} 1.
\end{equation}

Consequently, the leading part of \eqref{eq:moment-2-truncate} is: 

\begin{equation}
  m^{(2)}_{ \Theta_{L}(x_{\alpha}, x_{\beta})} = 2\left(\frac{1}{\sigma_{w}^2 AL}\right)^2\sum_{\ell = 0}^{L} \sum_{\ell'> \ell}^{L} \rho^{\alpha \beta}_\ell \rho^{\alpha \beta}_{\ell'} \left(1 + \mathbb{V}\left[\widetilde{\rmG}_{\ell'}^{\alpha\beta}\right] \right)
\end{equation}

Substituting the second moment into the definition of the spectral spread we obtain:
\begin{align} \label{eq:spectral-deviation-ntk-calculation-1}
\mathbb{S}\left[ \Theta_{L}(x_{\alpha}, x_{\beta})\right] &= \sqrt{2\left(\frac{1}{\sigma_{w}^2 AL}\right)^2\sum_{\ell = 0}^{L} \sum_{\ell'> \ell}^{L} \rho^{\alpha \beta}_\ell \rho^{\alpha \beta}_{\ell'} \mathbb{V}[\widetilde{\rmG}_{\ell'}^{\alpha\beta}} ]\\
&\propto \sqrt{\frac{1}{L^2}\sum_{\ell = 0}^{L} \sum_{\ell'> \ell}^{L} \rho^{\alpha \beta}_\ell \rho^{\alpha \beta}_{\ell'} \left(\frac{n_\textrm{out}}{N}
\left( L - \ell' + R^L_{\ell'}\right) \right)} \label{eq:spectral-deviation-ntk-calculation-2} \\
& \propto \sqrt{\frac{n_\textrm{out}}{N} L} \label{eq:spectral-deviation-ntk-calculation-3}
\end{align}

Where \eqref{eq:spectral-deviation-ntk-calculation-1} was obtained by squaring \eqref{eq:m1-NTK-derivation} and omitting the diagonal term $\ell = \ell'$. To go from \eqref{eq:spectral-deviation-ntk-calculation-1} to \eqref{eq:spectral-deviation-ntk-calculation-2} we simply injected \eqref{eq:rel-var-gaussian-G}, and the final step was using the fact that \(R^{L}_{j}\) defined in (\ref{eq:R_L_ellprime}) is bounded since it is the partial tail of a converging sum. One needs also to consider that $\rho^{\alpha \beta}_{\ell}$ is a convergent sequence. This concludes our study for the Gaussian case.

\item \textit{Orthogonal Case:} For the orthogonal case, we can have serious clues that the resulting control on the spectral standard deviation will be much tighter. Indeed, we have from \eqref{eq:relative-variance-weight-orthogonal} and \eqref{eq:rel-variance-Gtilde}:

\begin{equation} \label{eq:rel-var-ortho-G}
  \mathbb{V}\left[\widetilde{\rmG}_{\ell'
  }^{\alpha\beta}\right] = \frac{n_\textrm{out}}{N}\left(\sum_{\ell = \ell'+1}^{L} \mathbb{V}\left[\rmD_{\ell}^{\alpha}\rmD_{\ell}^{\beta}\right]\right).
\end{equation}

And using the same reasoning, we have: 

\begin{equation}
  \mathbb{S}\left[\widehat{ \Theta}^0_{L}(x_{\alpha}, x_{\beta})\right] = \sqrt{2\left(\frac{1}{\sigma_{w}^2 AL}\right)^2\sum_{\ell = 0}^{L} \sum_{\ell'> \ell}^{L} \rho^{\alpha \beta}_\ell \rho^{\alpha \beta}_{\ell'} \mathbb{V}\left[\widetilde{\rmG}_{\ell'}^{\alpha\beta}\right].}
\end{equation}
Now we can refine our previous convergence statement on $\mathbb{V}[\widetilde{\rmG}_{\ell'}^{\alpha\beta}]$, indeed since it is the tail of a $1 / \ell^2$ summand we have $\mathbb{V}[\widetilde{\rmG}_{\ell'}^{\alpha\beta}] \sim 1/ {\ell'}$. This directly gives us:

\begin{equation}
  \mathbb{S}\left[\widehat{ \Theta}^0_{L}(x_{\alpha}, x_{\beta})\right]  \propto \sqrt{\frac{n_\textrm{out}}{NL}}.
\end{equation}
\end{itemize}

\subsection{Eigenvectors and learning directions} \label{sec:app-eigenvectors}

\paragraph{Bound on $\normF{\frac{1}{L} \sum_{\ell = 1}^{L} {\rmP}^{\alpha \beta}_{L\to\ell} - \mathbf{I}}^2$}
In this section we will prove \eqref{eq:bounding-G}. First, recall that we have: 
\begin{equation}
  \normF{\frac{1}{L} \sum_{\ell = 1}^{L} \rmP^{\alpha \beta}_{L\to\ell} - \mathbf{I}}^2 \leq \frac{1}{L} \sum_{\ell = 1}^L \normF{{\rmP}^{\alpha \beta}_{L\to \ell} - \mathbf{I}}^2.
\end{equation}

Then expanding the summand leads to:
\begin{align}
  \normF{{\rmP}^{\alpha \beta}_{L\to\ell} - \mathbf{I}}^2 &= n_\textrm{out}\left( \frac{1}{n_\textrm{out}} \mathbb{E} \operatorname{Tr} \left({\rmP}^{\alpha \beta}_{L\to\ell} {\rmP}^{\alpha \beta \top}_{L\to\ell} \right) - \frac{2}{n_\textrm{out}} \mathbb{E} \operatorname{Tr} {\rmP}^{\alpha \beta}_{L\to\ell} + 1 \right)\\
  &\leq n_\textrm{out} \left(\sqrt{m^{(2)}_{{\rmP}^{\alpha \alpha}_{L\to\ell}}} \sqrt{m^{(2)}_{{\rmP}^{\beta \beta}_{L\to\ell}}} - 2 m^{(1)}_{{\rmP}^{\alpha \beta}_{L\to\ell}} + 1 \right).
\end{align}
Furthermore, we know from the definition (\ref{eq:rel-variance-def}) and the first moment calculation
(\ref{eq:m1-NTK-derivation}) that:
\begin{equation}
  \sqrt{m^{(2)}_{{\rmP}^{\alpha \alpha}_{L\to\ell}}} \sqrt{m^{(2)}_{{\rmP}^{\beta \beta}_{L\to\ell}}} = \frac{\Xi^{\alpha \alpha}_L \Xi^{\beta \beta}_L}{\Xi^{\alpha \alpha}_\ell \Xi^{\beta \beta}_\ell} \frac{L^2}{\ell^2} \sqrt{\left( 1 + \mathbb{V}\left[{\rmP}^{\alpha \alpha}_{L\to\ell}\right]\right)\left( 1 + \mathbb{V}\left[{\rmP}^{\beta \beta}_{L\to\ell}\right]\right)}.
\end{equation}
It is useful to restate previous section's results \eqref{eq:rel-var-gaussian-G} and \eqref{eq:rel-var-ortho-G}, indeed, relative variance $\mathbb{V}$ is scale independent so we have:
\begin{equation}
 \mathbb{V}\left[\rmP^{\alpha \alpha}_{L\to\ell}\right]= \mathbb{V}\left[\tilde{\rmG}_{\ell}^{\alpha\beta}\right] =
  \begin{cases}
   \frac{n_\textrm{out}}{N}\left( L - \ell + \sum_{\ell = \ell+1}^{L} \mathbb{V}[]\rmD_{\ell}^{\alpha}\rmD_{\ell}^{\beta}]\right)  & \text{(Gaussian case)}, \\
   \frac{n_\textrm{out}}{N}\left(\sum_{\ell = \ell+1}^{L} \mathbb{V}[\rmD_{\ell}^{\alpha}\rmD_{\ell}^{\beta}]\right) & \text{(Orthogonal case)}.
  \end{cases}
\end{equation}
Furthermore, we know using the same method as in (\ref{eq:m1-NTK-derivation}) that: 
\begin{equation}
  \frac{\Xi^{\alpha \alpha}_L \Xi^{\beta \beta}_L}{\Xi^{\alpha \alpha}_\ell \Xi^{\beta \beta}_\ell} \frac{L^2}{\ell^2} \xrightarrow{L, \ell \to \infty} 1.
\end{equation}
Hence, we know from previous equation that for sufficiently large $L, \ell$ we have: 
\begin{equation}
   \normF{{\rmP}^{\alpha \beta}_{L\to\ell} - \mathbf{I}}^2 \lesssim n_\textrm{out} \left(\sqrt{\left( 1 + \mathbb{V}[{\rmP}^{\alpha \alpha}_{L\to\ell}]\right)\left( 1 + \mathbb{V}[{\rmP}^{\beta \beta}_{L\to\ell}]\right)} - 1\right).
\end{equation}
Then for both orthogonal and Gaussian cases: 

\begin{equation}
 \normF{{\rmP}^{\alpha \beta}_{L\to\ell} - \mathbf{I}}^2 = 
  \begin{cases}
    O(\frac{n_{\textrm{out}}^2}{N} L)& \text{(Gaussian case)}, \\
     O(\frac{n_{\textrm{out}}^2}{NL}) = O(\frac{n_{\textrm{out}}^2}{L^2}\frac{L}{N})& \text{(orthogonal case)},
  \end{cases}
\end{equation}
Hence in the sequential limit when $L / N \ll 1$ requiring $n_\textrm{out} / L$ finite we can expect the orthogonal case to converge towards zero, which does not occur for the Gaussian initialisation. More precisely, under the following scaling with $\gamma \in (0,1)$ for the orthogonal case: 

\begin{equation}
  L \gtrsim \left(\frac{n_\textrm{out}^2}{N}\right)^{1 / (1 - \gamma)}
\end{equation}
we have directly:
\begin{equation}
   \normF{{\rmP}^{\alpha \beta}_{L\to\ell} - \mathbf{I}}^2 \lesssim L^{-\gamma} 
\end{equation}
then
\begin{equation}
   \frac{1}{L} \sum_{\ell = 1}^L \normF{{\rmP}^{\alpha \beta}_{L\to\ell} - \mathbf{I}}^2 \lesssim L^{-\gamma}
\end{equation}
and, finally, 
\begin{equation}
  \frac{1}{L} \sum_{\ell = 1}^{L} {\rmP}^{\alpha \beta}_{L\to\ell} \xrightarrow{L \to \infty} \mathbf{I}.
\end{equation}

\paragraph{Ces\`aro's theorem on the block NTK} Let's finally write $\epsilon^{\alpha\beta}_\ell$ the defect to the limiting correlation: \begin{equation}
  \epsilon_\ell^{\alpha \beta} := 
\rho^{\alpha \beta}_{\ell} - \rho^{\alpha \beta}_{\infty}  \quad \text{with}\quad \epsilon^{\alpha\beta}_\ell \to 0.
\end{equation}
Then we have: 
\begin{equation}
  \Theta^{\alpha\beta}_L = \frac{1}{A\sigma_w^2}\left( \frac{\rho_\infty^{\alpha \beta}}{L} \sum_{\ell = 0}^{L} {\rmP}^{\alpha \beta}_{L\to\ell} +\frac{1}{L} \sum_{\ell = 0}^{L} \epsilon^{\alpha\beta}_\ell {\rmP}^{\alpha \beta}_{L\to\ell}\right).
\end{equation}
Since ${\rmP}^{\alpha \beta}_{L\to\ell}$ is bounded we have by Ces\`aro's theorem the convergence in Frobenius norm:
\begin{equation}
  \Theta^{\alpha\beta}_L \xrightarrow{L \to \infty} \frac{1}{A\sigma_w^2}\rho^{\alpha \beta}_{\infty} \mathbf{I}.
\end{equation}

\subsection{From second moments of block NTK $\Theta^{\alpha\beta}_L$ to second moments of the global NTK $\Theta_L$} \label{sec:app-theta2gamma}

The full NTK matrix
 \(\Theta_{L} \in \R^{n_\textrm{out} D \times n_\textrm{out} D}\) second moment satisfies the following expansion :
  \begin{align}
    m^{(2)}_{ \Theta_{L}} & = \frac{1}{n_\textrm{out}D} \Tr \E \left[(\Theta_{L})^2\right] \\
    & = \frac{1}{D}\sum_{\alpha, \beta \in \{1,\ldots,D\}} \frac{1}{n_\textrm{out}} \Tr \E \left[ \Theta_L^{\alpha\beta} \Theta_{L}^{\alpha \beta \top} \right] 
  \end{align}
 
Using Cauchy-Schwarz, we get:
  \begin{align} \label{eq:cs-ntkApp}
    \frac{1}{n_\textrm{out}} \Tr \E\left[ { \Theta}_{L}^{\alpha\beta} { \Theta}_{L}^{\alpha \beta \top} \right] &\leq \sqrt{\frac{1}{n_\textrm{out}} \Tr \E\left( { \Theta}_{L}^{\alpha\alpha} \right)^2 } \sqrt{\frac{1}{n_\textrm{out}} \Tr \E \left( { \Theta}_{L}^{\beta \beta} \right)^2 } = \sqrt{m^{(2)}_{{ \Theta}_{L}^{\alpha\alpha} }}\sqrt{m^{(2)}_{{ \Theta}_{L}^{\beta\beta} }}.
  \end{align}
  The second useful inequality is: 
  \begin{align}
    \frac{1}{n_\textrm{out}} \Tr \E \left[ { \Theta}_{L}^{\alpha\beta} { \Theta}_{L}^{\alpha \beta \top} \right] &\geq \frac{1}{n_\textrm{out}} \Tr \E \left( {\Theta}_{L}^{\alpha\beta} \right)^2 =
    m^{(2)}_{{ \Theta}_{L}^{\alpha\beta} }
  \end{align}

\end{document}